# Simulating clinical interventions with a generative multimodal model of human physiology

Guy Lutsker[1,2,3], Gal Sapir[1,4], Jordi Merino[5], Smadar Shilo[1,2,6,7], Anastasia Godneva[1,2], Eli Meirom[3], Shie Mannor[3] , Hagai Rossman[4,8] , Gal Chechik[3] & Eran Segal[1,8,*]

## Author affiliations

[1] Department of Computer Science and Applied Mathematics, Weizmann Institute of Science, Rehovot, Israel.
[2] Department of Molecular Cell Biology, Weizmann Institute of Science, Rehovot, Israel.
[3] NVIDIA, Tel Aviv, Israel.
[4] Pheno.AI, Tel-Aviv, Israel.
[5] Novo Nordisk Foundation Center for Basic Metabolic Research, University of Copenhagen, Copenhagen, Denmark.
[6] Faculty of Medical and Health Sciences, Tel Aviv University, Tel-Aviv, Israel.
[7] The Jesse Z and Sara Lea Shafer Institute for Endocrinology and Diabetes, National Center for Childhood Diabetes, Schneider Children's Medical Center of Israel, Petah Tikva, Israel.
[8] Mohamed bin Zayed University of Artificial Intelligence, Abu Dhabi, UAE.

[*] Corresponding author:

Prof. Eran Segal, Department of Molecular Cell Biology and Department of Computer Science and Applied Mathematics, Weizmann Institute of Science, Rehovot, Israel, Tel: 972-8-934-3540, Fax: 972-8-934-4122, Email: eran.segal@weizmann.ac.il, ORCID: 0000-0002-6859-1164.

Abstract

Understanding how human health changes over time, and why responses to interventions vary between individuals, remains a central challenge in medicine. Here we present HealthFormer, a decoder-only transformer that models the human physiological trajectory generatively, by training on data from the Human Phenotype Project, a multi-visit cohort of over 15,000 deeply phenotyped individuals. We tokenise each participant's health trajectory across 667 measurements spanning seven domains: blood biomarkers, body composition, sleep physiology, continuous glucose monitoring, gut microbiome, wearable-derived physiology, and behaviour and medication exposure. We train HealthFormer to forecast individual physiological trajectories across these domains, and from this single generative objective a range of clinically relevant tasks can be expressed as queries on the model. We show that, without task-specific training, HealthFormer transfers to four independent cohorts and improves prediction for 27 of 30 incident-disease and mortality endpoints, exceeding established clinical risk scores in every comparison. We further show that the model can simulate interventions in silico: in a held-out personalised-nutrition trial, intervention-conditioned predictions recover individual six-month biomarker changes (e.g., Pearson r = 0.78 for diastolic blood pressure). Across 41 randomised intervention–outcome comparisons drawn from published trials, our results show that the predicted direction of effect agrees in every case, and the predicted mean falls within the reported 95% confidence interval in 30 cases. We position HealthFormer as an initial health world model, from which forecasting, risk stratification, and intervention-conditioned simulation arise as queries, providing a basis for clinical digital twins.

## Introduction

Understanding how human physiology evolves over time and responds to intervention is central to precision medicine. Although clinical decision-making depends on dynamic biological processes, most computational models are derived from episodic healthcare data that provide only sparse snapshots of patient state. Large-scale deep phenotyping [1], wearable sensing, and high throughput molecular profiling now enable continuous and repeated observation of human physiology in free-living conditions, capturing multi-year streams of measurements across timescales that standard clinical measures cannot resolve. Coupled with advances in self-supervised learning, these developments create an opportunity to learn physiological dynamics directly from complex data. Such measurements motivate computational representations of an individual's evolving health state, an idea captured by the clinical digital-twin concept [2] [3–5], and enable data-driven exploration of how responses to interventions vary between individuals.

Recent generative models of electronic health records have shown that longitudinal clinical histories contain sufficient structure to support prediction of disease onset, progression, and clinical outcomes at scale [6,7]. These approaches establish that unified patient representations can be learned from discrete medical histories [8–12]. However, these models are built primarily from routine-care records: diagnoses, procedures, medication orders, and infrequent laboratory measurements. They therefore represent physiology indirectly, through episodic clinical encounters, with limited resolution into the continuous and subclinical dynamics shaped by diet, behavior, environment, and treatment exposure. Furthermore, these records over-represent symptomatic individuals and under-represent healthy or pre-symptomatic populations, fundamentally limiting what can be learned about pre-clinical disease states and trajectories. By contrast, deeply phenotyped longitudinal cohorts combine protocolized clinical measurements with wearable, behavioral, imaging, and molecular data, enabling a more direct representation of physiological state and its evolution over time.

Complementary model classes address other pieces of this problem. Foundation models for individual modalities [13], such as retinal imaging [14], clinical text [15], and glucose monitoring [16], are biologically rich but confined to single data types; task-specific intervention models, such as statin effects on LDL [17] or exercise effects on blood pressure [18], are restricted to narrowly defined outcomes. Multimodal models integrate heterogeneous inputs but are generally static, ignoring the temporal dynamics that govern disease progression and treatment response [19] [20,21]. Analogous self-supervised approaches in single-cell genomics show that generative models trained on observational molecular profiles can learn perturbation-associated responses [22] [23–25,26], although the cellular setting differs from whole-person physiology in scale, timing, and degree of cross-system cross-talk. Existing approaches thus capture either longitudinal clinical events without dense physiology, or rich physiological measurements without whole-person temporal dynamics.

A central methodological challenge is how to integrate these heterogeneous data streams into a unified representation that preserves temporal structure across modalities. Tabular measurements, time series, molecular profiles, medications, and behavior are measured on different scales, at different frequencies, and with different missingness patterns. A model of human physiology must therefore represent both the content of each measurement and its timing, while allowing information from one physiological system to inform predictions in another.

To address this challenge, we present HealthFormer, a decoder-only transformer trained on the Human Phenotype Project (HPP) [1], a longitudinal, multi-visit cohort of over 15,000 deeply phenotyped individuals (Fig. 1a). Each participant's multimodal trajectory is represented as a unified, time-ordered token sequence spanning 667 parameters across seven domains (blood biomarkers, body composition, sleep physiology, glycaemic control, molecular profiling, wearable-derived physiology, and behavioral and medication data; see Supplementary Table S14). To our knowledge, this is a broader set of modalities than previously reported health-record transformers.

At inference, the model is queried by specifying a target modality and future timepoint, returning a full probability distribution over possible values in a single forward pass. This enables both longitudinal forecasting (Fig. 1b) and intervention-conditioned predictions under hypothetical trajectory modifications, including medications and lifestyle modifications, by conditioning on modified participant's medical events (Fig. 1c).Without task-specific training, HealthFormer produces intervention-conditioned predictions consistent in direction and approximate magnitude with effects reported in 41 intervention–outcome comparisons from randomised trials, including in populations distinct from the training cohort. For each comparison, we generated a synthetic population matched to the trial's reported baseline characteristics, with intervention tokens then appended at inference. These results suggest that deeply phenotyped multimodal longitudinal health data capture the necessary structure to motivate a data-driven world model of individual physiology [27,28]: a single representation from which forecasting, risk stratification, and intervention simulation emerge as downstream queries, and an initial step toward clinical digital twins.

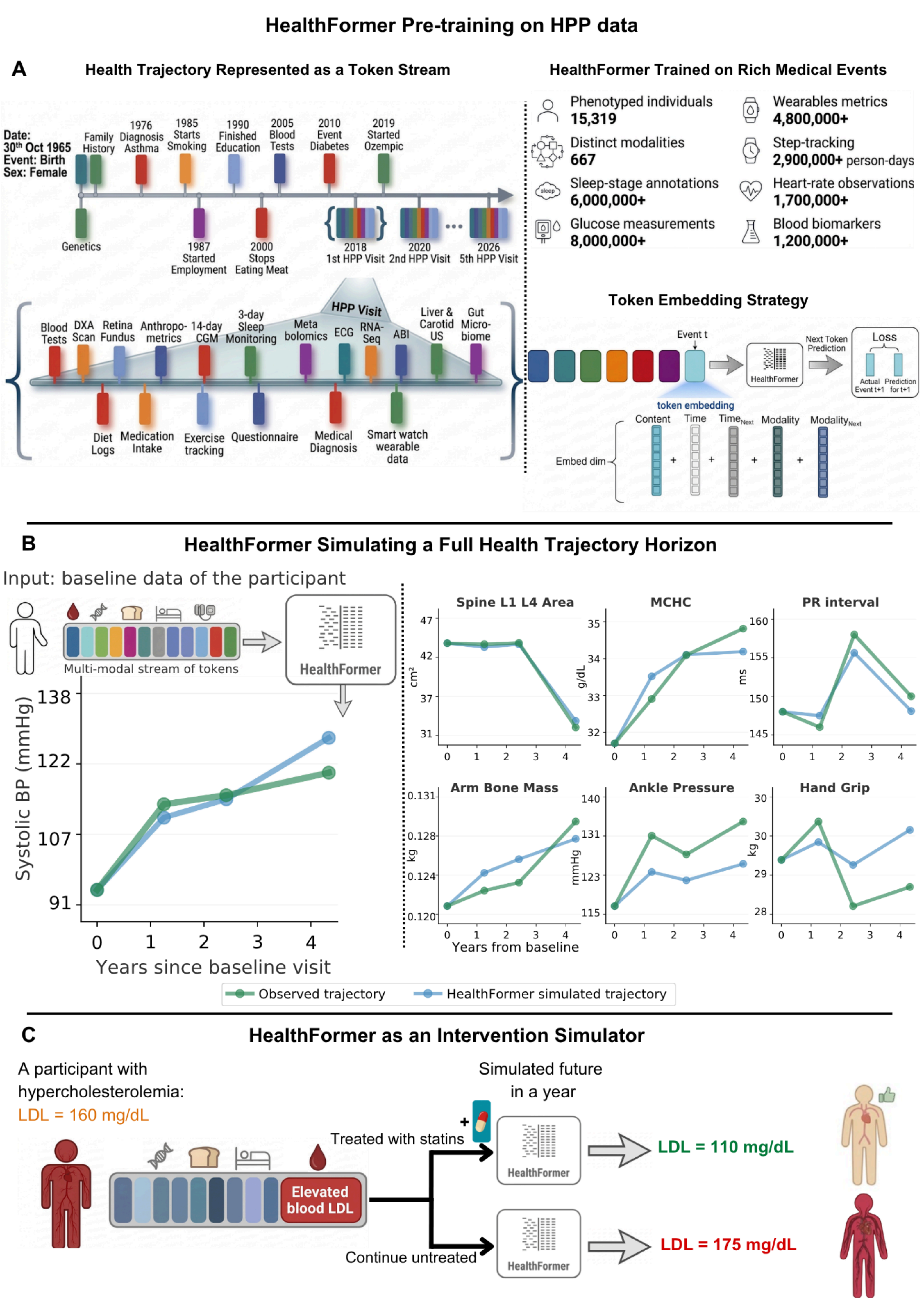
HealthFormer Pre-training on HPP data
A
Health Trajectory Represented as a Token Stream
Date: 30th Oct 1965
Event: Birth
Sex: Female
Family History
1976 Diagnosis Asthma
1985 Starts Smoking
1990 Finished Education
2005 Blood Tests
2010 Event Diabetes
2019 Started Ozempic
Genetics
1987 Started Employment
2000 Stops Eating Meat
2018 1st HPP Visit
2020 2nd HPP Visit
2026 5th HPP Visit
HPP Visit
Blood Tests
DXA Scan
Retina Fundus
Anthropo-metrics
14-day CGM
3-day Sleep Monitoring
Meta bolomics
ECG
RNA-Seq
ABI
Liver & Carotid US
Gut Micro-biome
Diet Logs
Medication Intake
Exercise tracking
Questionnaire
Medical Diagnosis
Smart watch wearable data
HealthFormer Trained on Rich Medical Events
Phenotyped individuals 15,319
Distinct modalities 667
Sleep-stage annotations 6,000,000+
Glucose measurements 8,000,000+
Wearables metrics 4,800,000+
Step-tracking 2,900,000+ person-days
Heart-rate observations 1,700,000+
Blood biomarkers 1,200,000+
Token Embedding Strategy
Event t
HealthFormer
Next Token Prediction
Loss
Actual Event t+1
Prediction for t+1
token embedding
Content
Time
Time_Next
Modality
Modality_Next
Embed dim
B
HealthFormer Simulating a Full Health Trajectory Horizon
Input: baseline data of the participant
Multi-modal stream of tokens
HealthFormer
Systolic BP (mmHg)
Years since baseline visit
Spine L1 L4 Area
MCHC
PR interval
Arm Bone Mass
Ankle Pressure
Hand Grip
Years from baseline
Observed trajectory
HealthFormer simulated trajectory
C
HealthFormer as an Intervention Simulator
A participant with hypercholesterolemia:
LDL = 160 mg/dL
Elevated blood LDL
Simulated future in a year
Treated with statins
Continue untreated
HealthFormer
LDL = 110 mg/dL
LDL = 175 mg/dL

**Figure 1 | HealthFormer model overview.**

**a,** HealthFormer pre-training on HPP. Each participant's multi-visit measurements - spanning blood biomarkers, DEXA body composition, sleep physiology, CGM, diet logs, ECG, gut microbiome sequencing, retinal imaging, and liver ultrasound are tokenized into a unified, time-ordered sequence with content, timestamp, inter-event interval, modality, and next-modality embeddings combined per token. The model is trained on 15,000+ participants with an autoregressive next-token objective; see Methods for the full modality list, the vocabulary, and the tokenization scheme.

**b,** Observed (green) and HealthFormer-simulated (blue) values for seven biomarkers over approximately four years for a single participant, conditioned only on that participant's baseline-visit measurements. Biomarkers span cardiovascular (Systolic BP, Ankle Pressure, PR interval), musculoskeletal (Spine L1 L4 Area, Hand Grip), body composition (Arm Bone Mass), and haematological (MCHC) domains.

**c,** HealthFormer as an intervention-conditioned predictor. Schematic of the framework for comparison of model-implied trajectories under alternative intervention. A sequence from a patient with hypercholesterolemia (elevated blood LDL = 160 mg/dL) is duplicated. In one path, tokens representing a statin medication are appended; the model's predicted LDL under the statin-conditioned sequence is lower (110 mg/dL) than under the untreated sequence (175 mg/dL).

## Results

### HealthFormer learns a useful unified representation of human physiology

We first asked whether a generative model trained on longitudinal multimodal health data can learn a unified representation of human physiology that captures structure across biological systems and over time. To address this, we evaluated HealthFormer across four complementary axes, ordered from local to temporal structure: reconstruction of observed biomarker values within visits, forecasting of future physiological states across a two-year horizon, scaling with model size, and recovery of cross-system relationships.

Across 667 health parameters on the held-out test set, within-visit reconstruction fidelity at the modality-category level was highest for ankle-brachial index (ABI, mean Pearson $r = 0.82$), DEXA body composition ($r = 0.81$), and sleep physiology ($r = 0.79$; Fig. 2a). All category-level mean correlations were significant ($p < 10^{-10}$ for each category). At the individual-biomarker level, within-visit predictions reached $r > 0.9$ (all $p < 10^{-10}$) for representative blood tests, body measures, and high-frequency physiological signals across 17 clinical biomarkers spanning seven domains (Fig. 2b). These within-visit correlations are supported by co-occurring measurements in the context window; the longitudinal and cross-system evaluations below provide more demanding tests.

To evaluate longitudinal predictive performance, we conditioned the model on baseline (visit 1) data for each participant and predicted measurements at a subsequent visit, which occurred approximately two years later (Methods). To execute this efficiently across all biomarkers simultaneously, we designed a specialised attention mask that permits parallel prediction of all visit 2 parameters in a single forward pass (see Methods). Longitudinal prediction accuracy varied across modalities, with mean Pearson correlations ranging from $r = 0.65$ for anthropometric measurements (19/19 modalities significant at $p < 0.05$) to $r = 0.64$ for microbiome composition (3/3 significant) to $r = 0.25$ for ultrasound-derived measures (30/30 significant) (Fig. 2c). Per-modality correlations were significant ($p_{FDR} < 0.05$) for 330 of 338 modalities evaluated; the 8 non-significant modalities were sparsely sampled sleep and blood markers (Supplementary Information). At the individual-biomarker level, within-visit and longitudinal performance can be compared directly across the same 17 biomarkers (Fig. 2b versus 2d). Within-visit prediction is systematically higher. The gaps are largest for variables that drift substantially over two years (sleep efficiency, QT interval, ankle pressure) and smallest for stable, highly autocorrelated traits (body weight, HDL, HbA1c). In longitudinal forecasting, HealthFormer outperforms independently trained per-modality regression models in 51 of 57 modalities with significant differences ($p_{FDR} < 0.05$), suggesting that a unified representation captures shared physiological structure beyond modality-specific optimization (Supplementary Fig. S2).

Prediction quality scaled with model size: increasing HealthFormer from 14 M to 139 M parameters produced monotonic improvements in longitudinal Pearson r and corresponding decreases in validation loss (Supplementary Fig. S17), consistent with scaling laws reported for language and vision. All subsequent results use the largest (139 M-parameter) model.

To test whether the model captures cross-modal relationships, we queried conditional distributions between variables spanning different physiological domains (Fig. 2e). Specifically, we constructed synthetic sequences containing a single physiological variable and queried the model for the predicted distribution of a variable in a different domain. Predicted conditional distributions closely matched empirical population-level relationships. These include both expected associations (for example, body weight from pelvic bone area, $r = 0.97$, $p = 2.4 \times 10^{-8}$; waist circumference predicted from RDI, $r = 0.91$, $p = 1.6 \times 10^{-7}$) and cross-system dependencies that require integration of information across modalities (all pairs significant at $p < 10^{-4}$). These associations emerge from self-supervised training without explicit variable pairing, indicating that the model captures population-level relationships across organ systems.

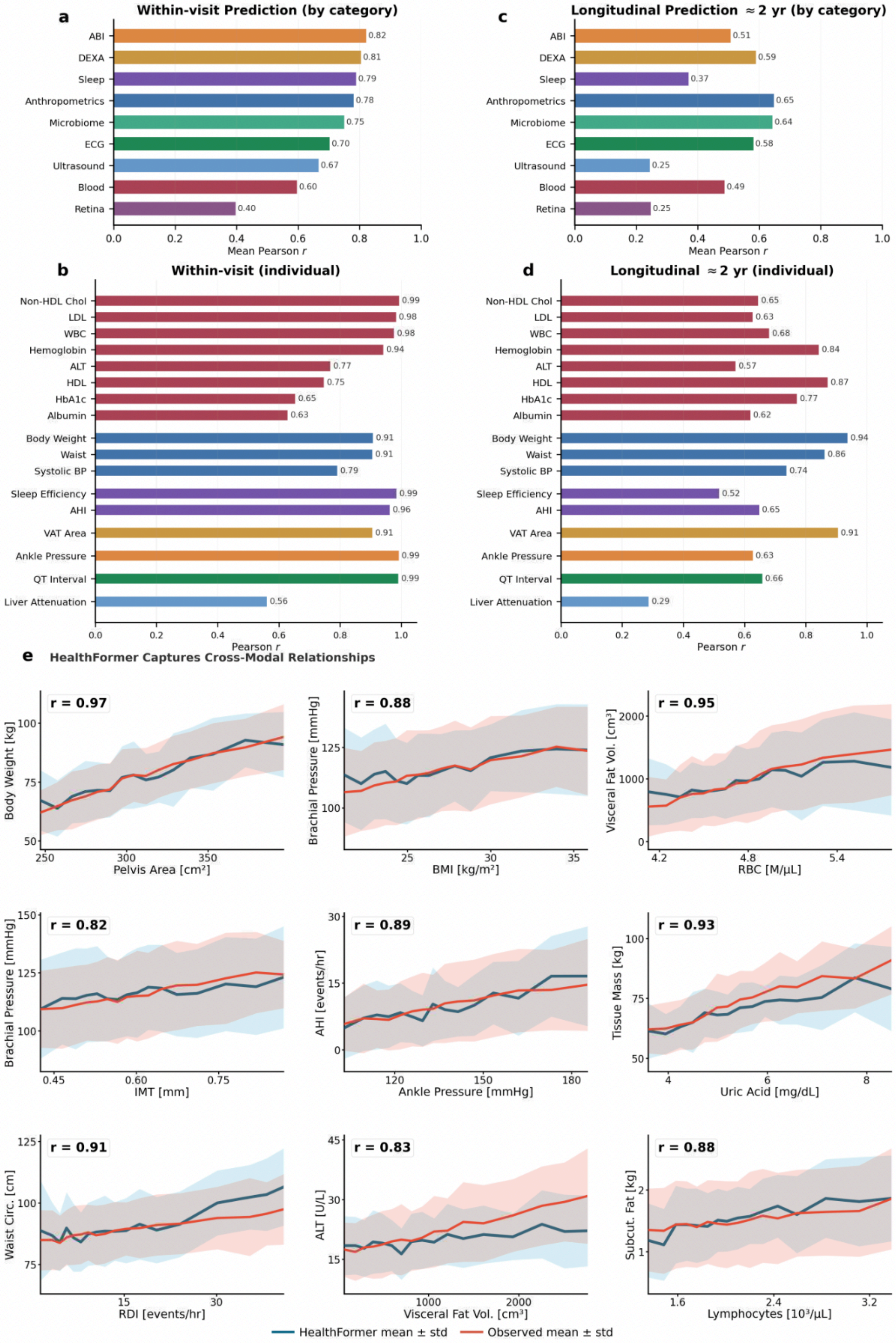

**Figure 2 | Within-visit reconstruction, longitudinal prediction, and cross-modal relationships learned by HealthFormer.**

a, Within-visit prediction performance across modality categories on the held-out test set (n= 3,217), measured by mean Pearson r. Highest values: ABI (r = 0.82), DEXA (r = 0.81), and Sleep (r = 0.79); all $p < 10^{-10}$.

b, Granular within-visit prediction for 17 individual clinical biomarkers spanning seven data domains: blood tests (Non-HDL Cholesterol, LDL, WBC, Hemoglobin, HDL, ALT, HbA1c, Albumin), body measures (Body Weight, Waist, Systolic BP), sleep (Sleep Efficiency, AHI), DEXA (VAT Area), vascular (Ankle Pressure), cardiac (QT Interval), and ultrasound (Liver Attenuation). Bars coloured by domain.

c, Longitudinal prediction accuracy over an approximately two-year horizon (predicting visit 2 states given only visit 1 data), grouped by modality category.

d, Granular longitudinal prediction for the same 17 biomarkers as panel b, enabling direct comparison.

e, HealthFormer captures cross-modal relationships. Nine plots display the model's predicted conditional distribution (blue, mean ± s.d.) versus the empirical population distribution (red, mean ± s.d.) for physiological variable pairs spanning data domains.

## Generalization to independent cohorts and disease-risk prediction

We next examined whether the learned representations generalize beyond the training cohort and encode information that can be used for clinical translation.

In zero-shot transfer, HealthFormer was applied without adaptation (model weights and tokenizer were both frozen) to four independent datasets: the UK Biobank[29] (United Kingdom; 74 shared modalities with HPP), the National Health and Nutrition Examination Survey [30] (NHANES, United States; 37 shared modalities), PNP3 [31] (Israel; 29 shared modalities), and the Framingham Heart Study [32] (United States; 8 shared modalities). Incident disease and mortality labels used for the downstream risk-prediction tasks below were derived from the respective cohorts' linked clinical records (UKB HES and death registries; NHANES linked mortality files), and were used only as evaluation targets, not as training signals. Performance varied across cohorts and modalities, with the UK Biobank achieving a median external accuracy of r = 0.52 (71/74 modalities significant at $p < 0.05$). NHANES and Framingham reached r = 0.39 (31/37 significant) and r = 0.45 (8/8 significant) respectively, and PNP3 r = 0.70 (27/29 significant) (Fig. 3a, 3b). On the UK Biobank, predictions of follow-up measurements given baseline data achieved correlations comparable to the training cohort for shared modalities (Fig. 3c).

We then evaluated whether the learned representations encode information for major health outcomes. Using HealthFormer's embeddings derived from UK Biobank baseline data alone, a downstream penalized Cox proportional-hazards model predicted incident disease and mortality across 30 endpoints (Fig. 3d; Methods). Against an age + sex + BMI baseline, HealthFormer embeddings improved concordance on 27 of 30 endpoints, with the largest gains for chronic obstructive pulmonary disease (+0.13), chronic kidney disease (+0.09), and hypertension (+0.07). Where established clinical risk scores are available, HealthFormer exceeded them in every comparison; for hypertension an increase over Framingham CVD score (+0.06), and in ischaemic stroke over PREVENT-ASCVD (+0.17). Absolute concordance indices reached, for example, C = 0.826± 0.005 for cardiovascular mortality, C = 0.808 ± 0.003 for heart failure, and C = 0.834± 0.002 for chronic kidney disease (mean ± s.d. across five cross-validation folds). Biological-age projection of the embeddings shows consistent metabolic and disease-prevalence gradients across NHANES quintiles (Fig. 3e–f), in line with the clinical interpretation of the learned representation.

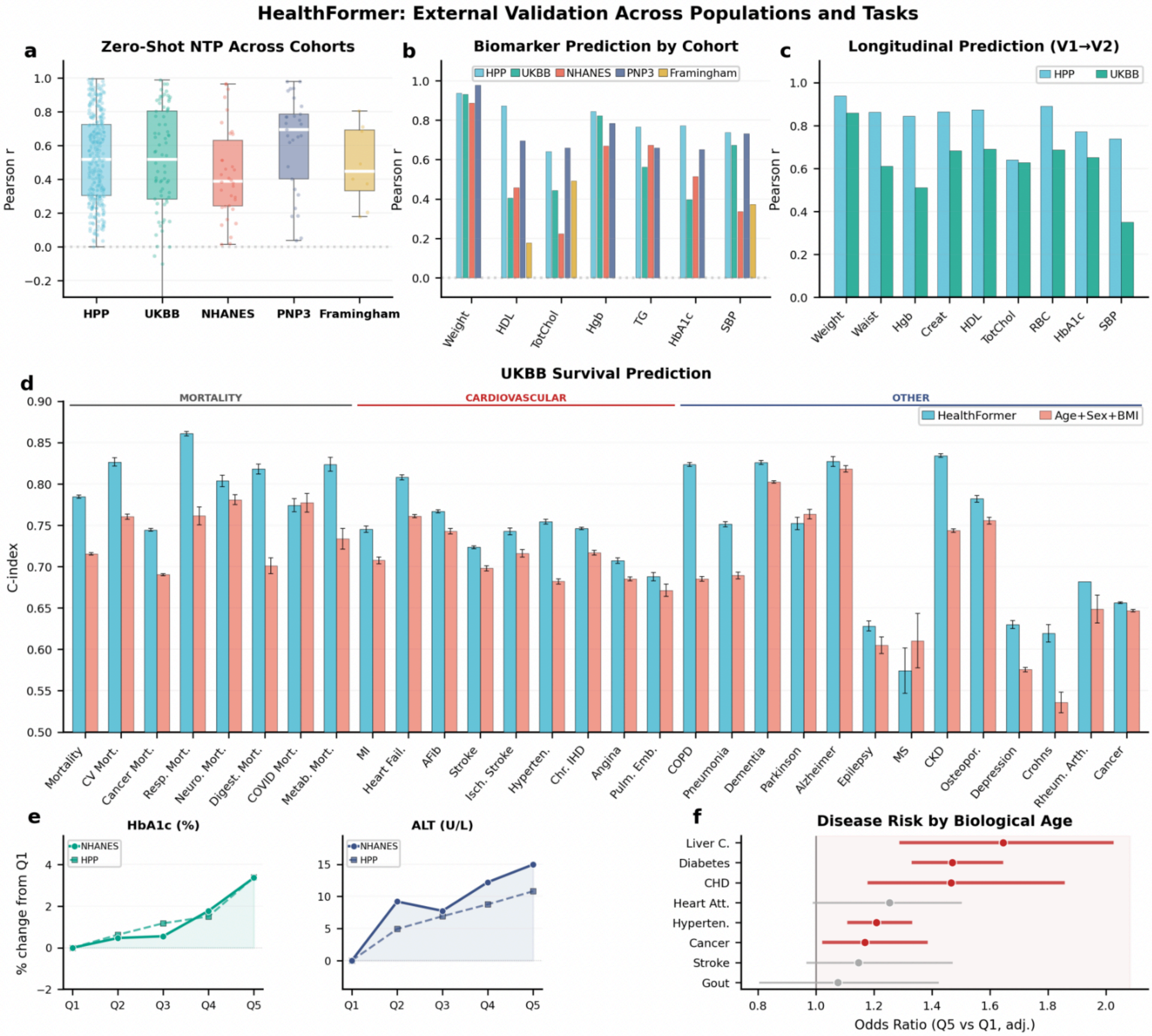

**Figure 3 | Generalization to independent populations and disease prediction.**

a, Zero-shot next-token prediction across cohorts. Per-modality Pearson r without fine-tuning for HPP (median r = 0.518, n = 338 modalities), UK Biobank (0.518, n = 74), NHANES (0.389, n = 37), PNP3 (0.695, n = 29), and Framingham (0.448, n = 8).

b, Biomarker-level zero-shot Pearson r across the five cohorts for specific clinical parameters (Weight, HDL, TotChol, Hgb, TG, HbA1c, SBP).

c, Longitudinal prediction (UK Biobank). Pearson r for predicting visit 2 measurements from visit 1 data, comparing HPP (blue) and UK Biobank (green).

d, UK Biobank survival prediction. C-index for 30 incident disease and mortality endpoints comparing HealthFormer embeddings (blue) against age + sex + BMI (red).

e, NHANES biological age quintile analysis. HealthFormer-derived biological age quintiles (Q1 = youngest, Q5 = oldest; adjusted for chronological age and sex) plotted against HbA1c and liver function (ALT) across age groups.
f, Disease risk by biological age (NHANES). Odds ratios (Q5 vs Q1, adjusted for age and sex) for eight conditions. Red bars indicate $P < 0.05$.

## Intervention-conditioned predictions in a personalised-nutrition trial

To test whether intervention-conditioned predictions track individual outcomes, we examined the Personalised Nutrition Project 3 (PNP3) [31], an independent randomised trial of 200 participants who received CGM-guided personalised nutrition over six months. PNP3 was not used during model training. For each participant, we tokenised baseline measurements (blood tests, body measures, demographics), six months of continuous glucose monitoring with concurrent diet logging, and any interim clinical measurements, and provided them to HealthFormer as a single token stream. The model was then used zero-shot to predict the six-month change on each biomarker (Fig. 4a).

At the individual level, predicted and observed six-month changes correlated across biomarkers. Three representative outcomes are shown in Fig. 4b: BMI ($r = 0.63$, $p = 3.7 \times 10^{-23}$), fasting glucose ($r = 0.61$, $p = 6.1 \times 10^{-22}$,), and diastolic blood pressure ($r = 0.78$, $p = 4.0 \times 10^{-40}$). Across the ten biomarkers examined, Pearson correlations between predicted and observed change ranged from $r = 0.49$ for Hba1c ($p = 1.4 \times 10^{-13}$) to $r = 0.89$ for triglycerides ($p = 4.6 \times 10^{-5}$) (Fig. 4c). All correlations were significant ($p < 0.05$) except HDL cholesterol (n=15). Per-biomarker p-values and sample sizes are reported in Supplementary Table S9. Providing the model with progressively more of each participant's pre-outcome trajectory improved prediction monotonically: the mean correlation across the biomarkers rose from $r = 0.39$ when only baseline measurements were available to $r = 0.64$ when CGM, diet, and interim clinical measurements were all included (Fig. 4d).

At the group level, predicted average treatment effects overlapped with the trial's reported estimates for most endpoints (Fig. 4e). Against a no-change baseline, HealthFormer reduced mean absolute error on 9 of 10 biomarkers, with the largest reductions on triglycerides, total cholesterol, and LDL cholesterol; the model did not improve on the no-change baseline for HDL cholesterol (Fig. 4f).

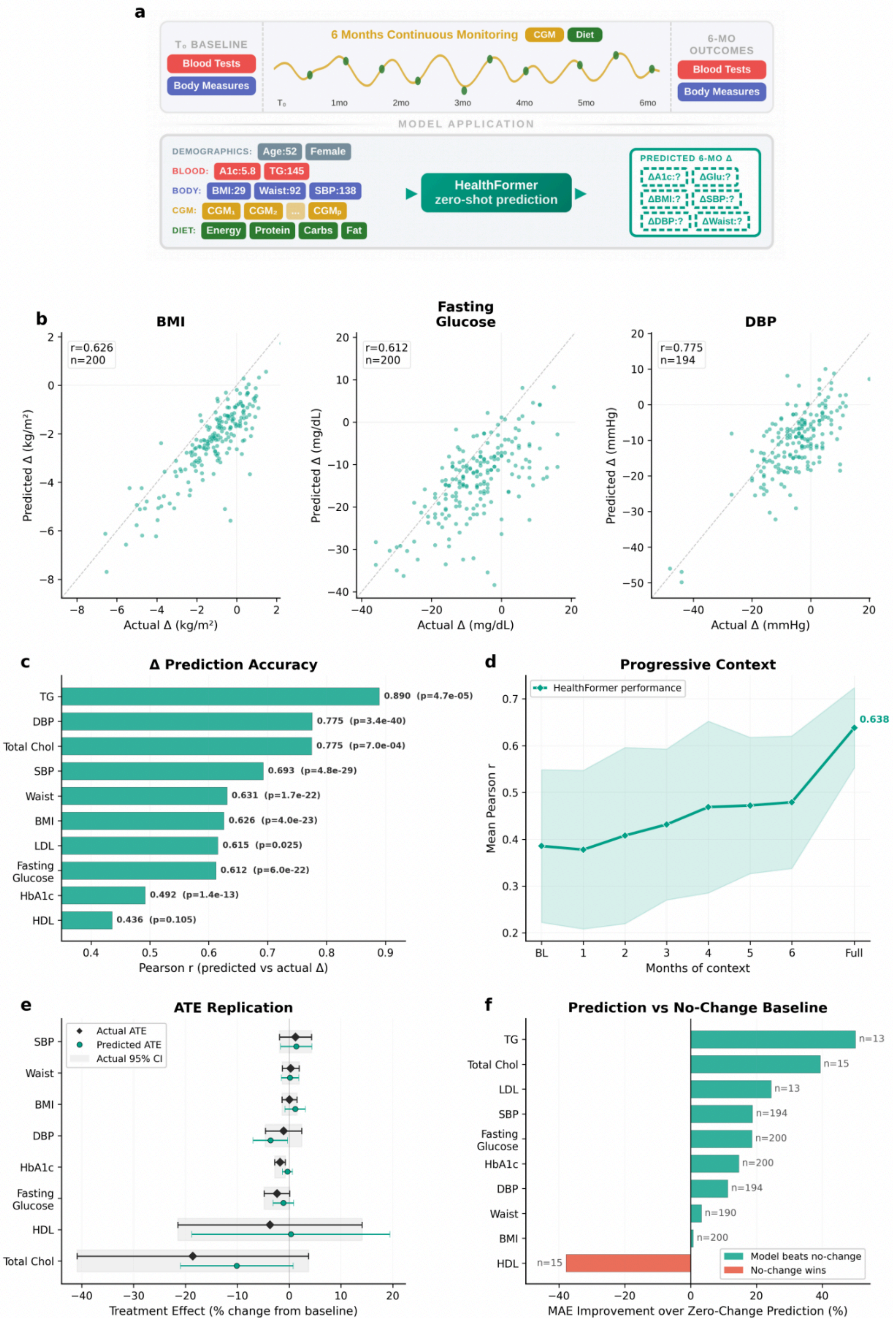

a
T₀ BASELINE
Blood Tests
Body Measures
6 Months Continuous Monitoring
CGM
Diet
1mo
2mo
3mo
4mo
5mo
6mo
6-MO OUTCOMES
MODEL APPLICATION
DEMOGRAPHICS: Age:52 Female
BLOOD: A1c:5.8 TG:145
BODY: BMI:29 Waist:92 SBP:138
CGM: CGM₁ CGM₂ ... CGMₚ
DIET: Energy Protein Carbs Fat
HealthFormer zero-shot prediction
PREDICTED 6-MO Δ
ΔA1c:? ΔGlu:? ΔBMI:? ΔSBP:? ΔDBP:? ΔWaist:?
b
BMI
r=0.626 n=200
Predicted Δ (kg/m²)
Actual Δ (kg/m²)
Fasting Glucose
r=0.612 n=200
Predicted Δ (mg/dL)
Actual Δ (mg/dL)
DBP
r=0.775 n=194
Predicted Δ (mmHg)
Actual Δ (mmHg)
c
Δ Prediction Accuracy
TG 0.890 (p=4.7e-05)
DBP 0.775 (p=3.4e-40)
Total Chol 0.775 (p=7.0e-04)
SBP 0.693 (p=4.8e-29)
Waist 0.631 (p=1.7e-22)
BMI 0.626 (p=4.0e-23)
LDL 0.615 (p=0.025)
Fasting Glucose 0.612 (p=6.0e-22)
HbA1c 0.492 (p=1.4e-13)
HDL 0.436 (p=0.105)
Pearson r (predicted vs actual Δ)
d
Progressive Context
HealthFormer performance
0.638
Mean Pearson r
BL 1 2 3 4 5 6 Full
Months of context
e
ATE Replication
Actual ATE
Predicted ATE
Actual 95% CI
SBP
Waist
BMI
DBP
HbA1c
Fasting Glucose
HDL
Total Chol
Treatment Effect (% change from baseline)
f
Prediction vs No-Change Baseline
TG n=13
Total Chol n=15
LDL n=13
SBP n=194
Fasting Glucose n=200
HbA1c n=200
DBP n=194
Waist n=190
BMI n=200
HDL n=15
Model beats no-change
No-change wins
MAE Improvement over Zero-Change Prediction (%)

**Fig. 4 | Intervention-conditioned predictions at the individual level in a personalised-nutrition trial (PNP3).**

**a, Study design and model application.** Top: PNP3 trial structure, with baseline and six-month clinical measurements bracketing six months of continuous glucose monitoring (CGM) and diet logging; green markers on the CGM trace indicate diet-logging events. Bottom: for each participant, demographics, blood, body, CGM stream, and macronutrient diet are tokenized and provided to HealthFormer, which predicts six-month biomarker changes zero-shot.

**b, Predicted versus observed six-month change** for BMI ($r = 0.626$, $n = 200$), fasting glucose ($r = 0.612$, $n = 200$), and diastolic blood pressure ($r = 0.775$, $n = 194$). Each point is one participant; the dashed line indicates $y = x$.

**c, Pearson correlation between predicted and observed six-month change for each of ten biomarkers,** using the full available context (CGM, diet, and interim clinical measurements). P-values are shown adjacent to each bar.

**d, Mean Pearson r across the six well-powered biomarkers ($n \geq 190$) as a function of the amount of pre-outcome context supplied to the model,** from baseline only through successive months of CGM and diet logging to the full context including interim clinic visits. Shaded band, ±1 s.d. across biomarkers.

**e, Trial-reported average treatment effects (black diamonds) and HealthFormer predicted average treatment effects (teal circles),** expressed as percentage change from baseline, with 95% confidence intervals.

**f, Percentage reduction in mean absolute error achieved by HealthFormer relative to a no-change (zero-delta) baseline prediction, per biomarker.**

$n = 200$ trial participants held out from model training; per-biomarker n is indicated in each panel where it varies.

## Comparison with randomized controlled trial estimates

To assess whether HealthFormer's intervention-conditioned predictions are consistent with effects reported in randomized controlled trials, we compared the model's predicted effect sizes against published estimates drawn from 41 intervention–outcome pairs spanning lipid-lowering, antihypertensive, antidiabetic, body-weight, exercise, and combination interventions (Fig. 5). For each comparison, a synthetic trial-matched population was generated by following the demographics and baseline biomarkers reported in the source trial's published Table 1 (Methods). Interventions were simulated by appending the appropriate medication or behaviour, tokens to each participant's sequence, and predicted changes were expressed in the same units as the source publication (see Methods).

Across 27 primary-outcome comparisons (Fig. 5a), the predicted direction of change agreed with the published trial estimate in every case, and the predicted effect size fell within the published 95% confidence interval in 21 of 27 comparisons. For eight of nine primary glycaemic endpoints, the predicted effect size fell within the published 95% confidence interval. These included empagliflozin on fasting glucose (EMPA-REG, Zinman 2015), empagliflozin on HbA1c (EMPA-REG, Zinman 2015), metformin on HbA1c (UKPDS 34), metformin on glucose (DPP, Knowler 2002), aerobic exercise on HbA1c (Boule 2001), yoga on fasting glucose (Cui 2017), dapagliflozin on HbA1c (DECLARE-TIMI 58, Wiviott 2019), and walking on HbA1c (Qiu 2014). The exception was semaglutide on HbA1c (SUSTAIN-1, Sorli 2017), with a predicted effect size of 11.4% (95% CI 11.2–11.6) compared to the published 19.3% (16.9–21.6). In the blood-pressure domain, the ALLHAT effects of amlodipine and chlorthalidone (the trial's thiazide-type diuretic) on systolic blood pressure were closely reproduced (7.9% [95% CI 7.4–8.4] and 7.7% [95% CI 7.2–8.3] against published values of 7.5% [6.5–8.5] and 8.2% [7.2–9.2] respectively), as were the aerobic-exercise (Cornelissen 2013), swimming (Nualnim 2012), and beta-blocker (Wiysonge 2017) endpoints. Losartan on SBP (LIFE, Dahlöf 2002) was predicted at 15.0% (95% CI 14.5–15.6) against a published 17.4% (16.7–17.9), falling just below its published interval. Ramipril on SBP (HOPE, Yusuf 2000) was over-predicted at 10.1% (95% CI 9.6–10.5) versus a published 2.4% (1.5–3.3%). Of the three primary body-weight endpoints, liraglutide (SCALE, Pi-Sunyer 2015) and low-dose semaglutide (SUSTAIN-6, Marso 2016) were recovered within their intervals (7.5% [95% CI 6.9–8.0] and 5.1% [95% CI 4.5–5.7]), whereas high-dose semaglutide in STEP-1 (Wilding 2021) was under-predicted (8.3% [95% CI 7.8–8.8] versus a published 14.9% [13.8–16.0]).

For lipid lowering medications, the model reproduced the lower-magnitude endpoints within their published intervals and under-predicted the high-potency statin endpoints from STELLAR (Jones 2003). Ezetimibe on LDL cholesterol was predicted at 18.1% (95% CI 16.6–19.6), against 18.6% (17.5–19.7) published by Pandor 2009. Aerobic exercise on HDL cholesterol was predicted at +3.8% (95% CI +2.7 to +4.8), against +5.6% (3.0–8.0) reported by Kodama 2007. The two highest-potency statin endpoints from STELLAR were under-predicted: rosuvastatin on LDL at 24.7% (95% CI 23.1–26.3) against a published 46% (43–49), and atorvastatin on LDL at 25.1% (95% CI 23.9–26.3) against 37% (35.5–38.5), mirroring the under-prediction seen for high-dose semaglutide on HbA1c. Among the four combination-therapy endpoints, all four were recovered within the published interval: ACE-inhibitor + amlodipine (benazepril/amlodipine, ACCOMPLISH, Jamerson 2008); amlodipine + aerobic exercise (additive estimate, AHA 2017 + Cornelissen 2013); ramipril + HCTZ (additive estimate, Law 2009); and empagliflozin + losartan (Scholtes 2023). The empagliflozin + losartan endpoint sat at the upper limit of its interval, predicted at 15.0% (95% CI 14.5–15.5) against a published 10.8% (7.2–15.1).

For the 14 secondary-outcome comparisons (Fig. 5b), the predicted direction matched the published trial estimate in every case, and the predicted effect size fell within the published 95% confidence interval in 9 of 14 comparisons. Recovered within their intervals were semaglutide on SBP (STEP-1), walking on SBP (Murtagh 2015), metformin on body weight (DPP, Knowler 2002), semaglutide on waist circumference (STEP-1), the SGLT2-inhibitor class effect on triglycerides (Bechmann 2024), and the STELLAR secondary-lipid endpoints (rosuvastatin and atorvastatin on triglycerides and HDL). The remaining five rows followed a consistent pattern: the small off-target SBP and weight effects attributable to SGLT2 inhibitors and GLP-1 agonists (empagliflozin and dapagliflozin on SBP and weight, and liraglutide on SBP) were over-predicted by 2–9 percentage points. For example, empagliflozin on SBP was predicted at 9.8% (95% CI 9.4–10.2) against a published 3.1% (2.0–4.5). This pattern may reflect a tendency of observational training data to attribute a larger share of co-occurring cardiometabolic change to these agents than randomised trials report.

Across all 41 endpoints, the predicted direction agreed with the published estimate in every comparison, and the predicted mean fell within the published 95% confidence interval in 30 of 41 (73.2%) comparisons. The mis-specified endpoints clustered into three interpretable groups. First, the highest-dose statin and GLP-1 endpoints, where a fixed intervention-token budget may not resolve dose-intensive regimens. Second, the LIFE losartan and HOPE ramipril BP-drug primaries, where the predicted SBP response either fell just below the published interval (LIFE) or substantially exceeded it (HOPE). Third, secondary SGLT2/GLP-1 off-target SBP and weight endpoints, to which the model assigned a larger off-target effect than the source trials reported. The remaining endpoints were recovered within the published intervals without task-specific fine-tuning, using only intervention tokens appended to synthetic populations matched to each trial's published Table 1. These comparisons are consistent with broad agreement in direction and approximate magnitude between intervention-conditioned predictions and published randomized trial estimates, rather than with demonstrated causal treatment effects, which would require prospective evaluation.

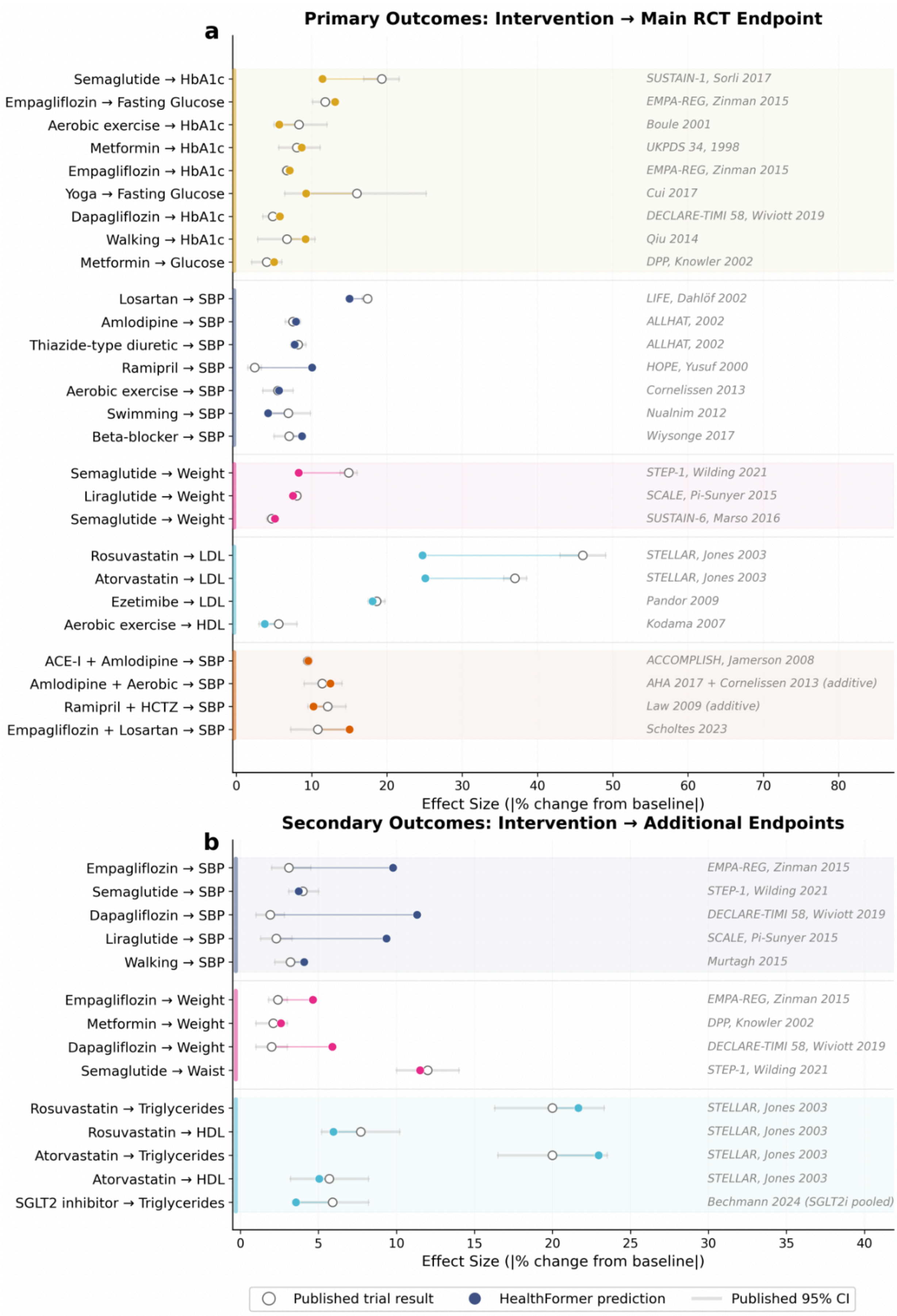

**Figure 5 | Comparison of intervention-conditioned predictions with published randomised trial estimates.**
Forest plots of HealthFormer-predicted effect sizes (filled coloured circles) against published trial point estimates (open circles) and 95% confidence intervals (grey bars), with a thin coloured line connecting each model–trial pair to indicate the absolute gap, for 41 intervention–outcome pairs. Values are absolute percent change from baseline. Trials are grouped by outcome domain (coloured left margin): lipids (teal), blood pressure (dark blue), glycemic control (gold), body weight (pink), and combination therapies (orange). Source trial and first-author year are shown in the right margin.
a, Primary outcomes (n = 27 comparisons): predicted direction agreed with the published estimate in 27 of 27 cases, and predicted effect size fell within the published 95% CI in 21 of 27 cases.
b, Secondary outcomes (n = 14 comparisons): predicted direction agreed in 14 of 14 cases, and predicted effect size fell within the published 95% CI in 9 of 14 cases. Across both panels (n = 41), 30 of 41 predictions fell within the published 95% CI and all 41 agreed in direction. Predictions were generated zero-shot by appending intervention tokens to synthetic populations matched to each trial's published Table 1 (see Methods).

## Predicted twelve-month trajectories across pharmacological, exercise, and dietary interventions

To examine the time course of intervention-conditioned predictions, we generated model-implied twelve-month trajectories at monthly resolution across pharmacological, exercise, dietary, and behaviour interventions (Fig. 6). For each analysis, the predicted trajectory of a treated synthetic population is compared with an untreated control population matched on baseline covariates. Pharmaceutical predictions are restricted to treatment-naive participants with the relevant indication.

Among lipid-lowering drugs at daily dosing, predicted LDL cholesterol fell progressively under five treatment regimens (rosuvastatin, ezetimibe + simvastatin, atorvastatin, simvastatin, and ezetimibe alone), in an order consistent with published potency rankings [33] (Fig. 6a). Among antihypertensive medications simulated at their respective clinical dosing frequencies in treatment-naive hypertensive participants, five treatment regimens (thiazide diuretics, amlodipine + valsartan, losartan, enalapril, and bisoprolol) each produced sustained reductions in systolic blood pressure over twelve months, with the thiazide class giving the largest predicted change (Fig. 6b). Among glucose-lowering drugs in treatment-naive diabetic participants, predicted fasting glucose fell under metformin at twice-daily dosing, an aggregated antidiabetic class, empagliflozin, and semaglutide (Fig. 6c). Across exercise modalities, running, weight lifting, swimming, and basketball were each associated with progressively lower predicted visceral adipose tissue over twelve months (Fig. 6d). Walking, swimming, basketball, and weight lifting each lowered predicted systolic blood pressure over the same period, with walking producing the largest sustained reduction (Fig. 6e). Walking frequency was itself dose-ordered. Daily sessions produced a larger predicted reduction in systolic blood pressure than three sessions per week, which in turn exceeded weekly sessions (Fig. 6f).

The remaining analyses examined interventions for which no directly comparable randomised trial is available. Caloric restriction simulated at 10%, 20%, 30%, and 40% produced monotonically larger predicted reductions in systolic blood pressure over twelve months (Fig. 6g). Simulated changes in dietary fibre intake, ranging from −30% to +30% of baseline, produced dose-ordered changes in predicted gut-microbiome Shannon diversity. Positive shifts were associated with increased diversity and negative shifts with decreased diversity, and the predicted effect scaled with the magnitude of the dietary change (Fig. 6h). Simulated CPAP therapy was modelled as a 75% reduction in the apnoea-hypopnoea index in participants with sleep-disordered breathing, within the range reported for adherent use [34]. It produced concurrent reductions in predicted systolic blood pressure, fasting glucose, HbA1c, triglycerides, and ALT over twelve months (Fig. 6i). These panels illustrate the range of interventions HealthFormer can be conditioned on and the time course of the resulting predictions. They do not establish causal treatment effects: the trajectories shown should be read as candidates for prospective evaluation rather than as evidence of clinical benefit.

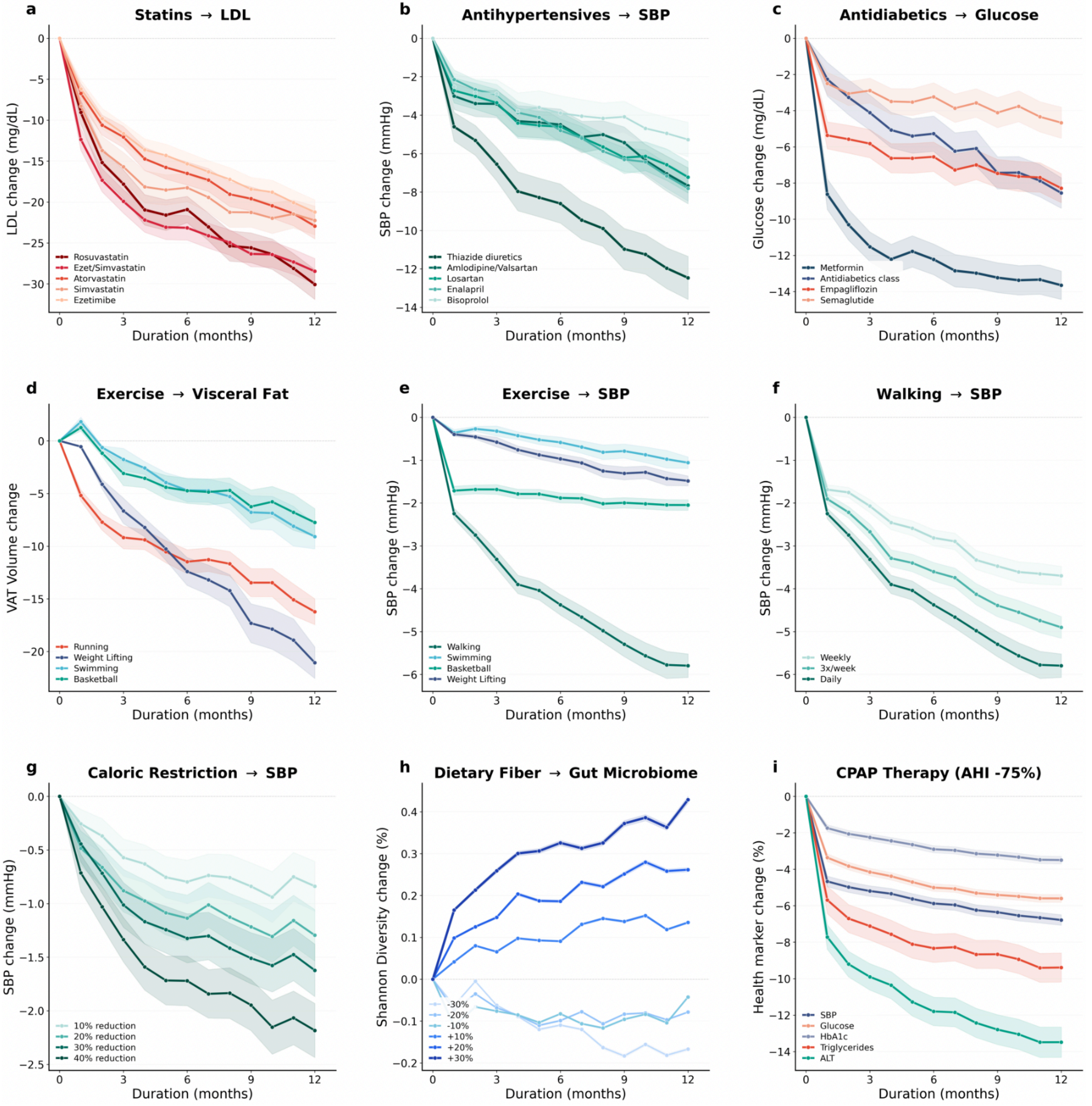

**Figure 6 | Predicted twelve-month trajectories of intervention-conditioned outcomes across pharmacological, exercise, and dietary interventions.**

All panels show the mean predicted change in a treated synthetic population relative to an untreated control population matched on baseline covariates, at monthly resolution over twelve months. Shaded bands indicate ±1 s.e.m. Pharmaceutical predictions are restricted to treatment-naive participants with the relevant indication.

**a,** LDL cholesterol under five lipid-lowering agents at daily dosing: rosuvastatin, ezetimibe with simvastatin, atorvastatin, simvastatin, and ezetimibe.

**b,** Systolic blood pressure under five antihypertensive agents at their respective clinical dosing frequencies in treatment-naive hypertensive participants: thiazide diuretics, amlodipine with valsartan, losartan, enalapril, and bisoprolol.
**c,** Fasting glucose under four glucose-lowering agents in treatment-naive diabetic participants: metformin (twice daily), an aggregated antidiabetic class, empagliflozin, and semaglutide.
**d,** Visceral adipose tissue volume under four exercise modalities: running, weight lifting, swimming, and basketball.
**e,** Systolic blood pressure under four exercise modalities: walking, swimming, basketball, and weight lifting.
**f,** Systolic blood pressure as a function of walking frequency: weekly, three times weekly, and daily.
**g,** Systolic blood pressure under simulated caloric restriction at four levels: 10%, 20%, 30%, and 40% of baseline intake.
**h,** Gut-microbiome Shannon diversity under simulated changes in dietary fibre intake at six levels: −30%, −20%, −10%, +10%, +20%, and +30% of usual intake.
**i,** Systolic blood pressure, fasting glucose, HbA1c, triglycerides, and alanine aminotransferase (ALT) under simulated CPAP therapy, modelled as a 75% reduction in apnoea-hypopnoea index (AHI) in participants with sleep-disordered breathing.

## Discussion

We asked whether a unified, generative representation of human physiology could be learned from multimodal longitudinal data, and whether such a representation could recapitulate the population-level effects of known medical interventions reported in randomised trials. HealthFormer, trained without task-specific objectives, reconstructed within-visit health state, forecast biomarker changes over a multi-year horizon, transferred to four independent cohorts, encoded incident disease and mortality risk, and produced intervention-conditioned trajectories that were broadly aligned with external trial evidence.

Much of clinical AI has treated the patient either as a set of discrete medical events or as a collection of separate modalities. Recent large-scale models [35] [7] show the power of modelling clinical records as longitudinal sequences of diagnoses, procedures, medications, notes, images, and other events. Modality-specific foundation models show that individual physiological signals can support broad downstream prediction. HealthFormer expands the scope of these recent studies by addressing a different object: quantitative physiological trajectories across interacting biological systems. Its tokens include quantitative measurements of blood chemistry, body composition, sleep, glycaemic control, molecular profiles, wearable-derived physiological parameters, behaviour, and medication exposure. The change is therefore from modelling the medical record, a log of clinical encounters, to modelling the physiological trajectory that unfolds between them.

This distinction matters because intervention simulation is not naturally a property of static risk models. A conventional risk model can estimate whether a person is likely to develop disease; it does not readily answer how that person's LDL cholesterol, HbA1c, blood pressure, liver enzymes, adiposity, sleep physiology, and microbiome diversity might evolve under alternative health trajectories. In HealthFormer, these questions are posed within the same generative representation in which a patient's history is modified, an intervention token is introduced, and the model returns a distribution over future physiological states. The relevant claim is not that this procedure generates evidence of individual causal effects, but that intervention-conditioned prediction becomes a native query on a learned model of physiological dynamics.

The intervention results support this interpretation. In the PNP3 nutrition trial, predicted six-month biomarker changes tracked observed individual-level changes, and prediction quality improved monotonically as more of each participant's pre-outcome trajectory was supplied. Across 41 published randomised-trial comparisons, intervention-conditioned predictions agreed with the reported direction of effect in every case, and the predicted mean fell within the published 95% confidence interval in 30 of 41. The model recovers target-specific effects across lipid-lowering, antihypertensive, glycaemic, weight-loss, exercise, dietary, and behaviour-related interventions, indicating that it has learnt more than a generic association between treatment and subsequent improvement.

The pattern of disagreement is itself informative. The largest under-predictions cluster at the high end of the published effect-size distribution: rosuvastatin and atorvastatin on LDL, semaglutide on HbA1c, and semaglutide on weight are among the largest reported effects in our comparison set, and each is under-predicted. We interpret this primarily as a training-distribution effect rather than a set of drug-specific failures. HPP is a deeply phenotyped general-population cohort in which most participants do not undergo biomarker changes of the magnitude observed in randomised trials over the training window, and pharmacological exposures are observational rather than protocolised at fixed dose. The model therefore fits a conditional response distribution that under-represents the large-effect tail seen in RCT-eligible populations, and its predictions regress toward the smaller, more typical changes that dominate the training data. Drug-specific factors compound this. Statin exposure tokens do not currently distinguish moderate-intensity from high-intensity regimens, which compresses the LDL response distribution at the high-potency end. The most recent generation of high-potency GLP-1 receptor agonists, including semaglutide, is sparsely represented in HPP relative to earlier and lower-dose exposures. The cardiometabolic secondary effects of SGLT2 inhibitors and GLP-1 receptor agonists, on blood pressure, lipids, and weight, depend on pathways the current vocabulary captures only indirectly through their primary glycaemic tokens. Both the structural and the drug-specific contributions are addressable: through richer dose, intensity, titration, and adherence tokens; through targeted enrichment of recent-generation drug exposures in future training cohorts; and through trial-aligned baseline sampling at inference time.

HealthFormer reproduces the direction and approximate magnitude of population-level trial effects across many drug classes, and tracks individual biomarker changes in a held-out prospective trial, but we do not present it as a counterfactual or causal engine. It learns associations between intervention exposure and subsequent physiological dynamics with enough specificity to be tested against external evidence: a stronger claim than ordinary forecasting, and a weaker one than causal identification. This makes intervention simulation a testable downstream use of a generative physiological model, rather than a separate supervised task trained one endpoint at a time. It also provides a practical route for generating hypotheses about which physiological systems may change under a candidate intervention, over what timescale, and in which baseline phenotypic contexts.

A fitted generative model of physiology can be queried on input configurations that were not directly observed during training, and its predictions tested against external evidence — analogous to the way a fitted physical model is interrogated under hypothetical conditions. The randomised-trial comparisons exercise exactly this property. Synthetic cohorts were generated based on each trial's published Table 1, corresponding to populations whose specific covariate combinations were absent from, or under-represented in, the training cohort, and interventions were represented by the same tokens used during pre-training. The PNP3 analysis (Fig. 4) provides the complementary empirical anchor: held-out real participants, observed six-month biomarker changes, and individual- as well as group-level evaluation. The two analyses answer different questions: whether the model's internal dynamics agree with external trial evidence (Fig. 5), and whether they track real individual outcomes (Fig. 4). Together they constrain interpretation more than either alone. In this sense, the trial comparisons treat HealthFormer as a model of the world to be probed.

Several limitations are important to mention. First, HealthFormer is trained on observational trajectories and learns associational rather than causal predictions. As evaluated here, intervention-conditioned outputs support prognostic stratification across patients exposed to a given intervention; they do not constitute counterfactual efficacy estimation across alternative interventions for the same individual, which would require outcome data on multiple comparable interventions per participant — a far harder data regime to obtain. Longitudinal prediction remains imperfect, particularly for less stable or sparsely sampled modalities. The randomised-trial comparisons use synthetic trial-matched populations rather than original trial participants, and cover continuous biomarker outcomes rather than hard clinical events. Intervention representation is still coarse: dose, adherence, titration, discontinuation, indication, and timing are only partially captured, and intention-to-treat versus per protocol estimates can not be generated. The model also operates on discretised summaries of physiology, whereas many clinically important processes are continuous, high-frequency, spatial, or mechanistic.

Each of these limitations corresponds to a tractable next step. A model such as HealthFormer could serve as the generative substrate on which more explicitly causal and mechanistic systems are built. Prospective randomised calibration, expanded prospective deployments analogous to PNP3, and external validation against future trial readouts will define when HealthFormer's intervention-conditioned predictions can support individual decision-making. Complementary causal-inference designs, including target-trial emulation and balancing methods, can be layered on the learned representation for specific endpoints when the data and design support them, but the generative model is the substrate, not a substitute for that machinery. Richer tokenisation of dose and adherence, together with raw wearable streams, ECG, imaging, molecular time series, and dense longitudinal sampling, should improve both temporal resolution and biological specificity. In this sense, HealthFormer is not yet a clinical digital twin. It is an initial health world model: a single generative representation of human physiology from which forecasting, risk stratification, and intervention-conditioned simulation emerge as downstream queries. Causally calibrated clinical digital twins may eventually be built on top of it.

## Methods

Study Cohort and Data

The Human Phenotype Project

HealthFormer was developed using data from the Human Phenotype Project (HPP) (Reicher et al., Nat. Med. 2025), a longitudinal deep-phenotyping study that collected multimodal health measurements from 15,319 participants. Each participant underwent comprehensive clinical assessments spanning laboratory tests, imaging, wearable sensor recordings, genomic profiling, microbiome sequencing, dietary logging, and structured questionnaires. Participants attended one or more clinical visits, with repeat visits separated by approximately two years, providing longitudinal trajectories for a subset of the cohort.

The resulting dataset comprised 64,223,972 tokens in total, of which 51,817,346 (80%) represented continuous-valued measurements and 12,406,626 (20%) represented categorical variables. The continuous tokens encoded numeric measurements (e.g., blood glucose concentration in mg/dL, bone mineral density in g/cm2, heart rate in bpm), while categorical tokens encoded discrete outcomes (e.g., medication class, exercise type, questionnaire response categories).

Modality Categories

The 667 modalities were organized into 19 categories, spanning clinical laboratory tests, imaging, physiological monitoring, genomics, metabolomics, and behavioural assessments.

The categories are: (1) CGM: 1 modality, approximately 8.1 million tokens (continuous), capturing interstitial glucose concentration via subcutaneous sensor at approximately 5-minute intervals. (2) Blood: 69 modalities, approximately 1.2 million tokens (continuous), comprising standard clinical test panels including complete blood count, comprehensive metabolic panel, lipid panel, liver function, renal function, inflammatory markers, hormones, and glycaemic markers. (3) Body Measures: 26 modalities, approximately 409,000 tokens (continuous), including height, weight, BMI, waist and hip circumference, body fat percentage, and limb-specific measurements. (4) DEXA: 95 modalities, approximately 1.8 million tokens (continuous), providing regional body composition from dual-energy X-ray absorptiometry. (5) ECG: 61 modalities, approximately 1.1 million tokens (continuous), capturing resting 12-lead electrocardiography features. (6) Ultrasound: 37 modalities, approximately 733,000 tokens (continuous), including carotid intima-media thickness, plaque measurements, liver steatosis, and vascular flow parameters. (7) Sleep: 74 modalities, approximately 3.0 million tokens (continuous), from polysomnography-derived metrics including total sleep time, sleep efficiency, AHI, ODI, and SpO2 measures. (8) Wearable: 12 modalities, approximately 4.8 million tokens (continuous), from actigraphy and wearable sensors. (9) Diet: 35 modalities, approximately 13.1 million tokens (continuous), from nutritional intake logging. (10) Exercise: 17 modalities (categorical), recording exercise type as discrete events. (11) Medications: 234 modalities, approximately 62,000 tokens (categorical), organized by pharmacological class. (12) Microbiome: species-level relative abundances for 1,259 microbial species across approximately 6.1 million tokens (continuous), derived from shotgun metagenomic sequencing. (13) RNA: 500 modalities, approximately 1.9 million tokens (continuous), capturing the 500 most variable genes by inter-individual expression variance. (14) PRS: 16 modalities, approximately 63,000 tokens (continuous), representing 16 principal components from genome-wide SNP data. (15) Nightingale: 20 modalities, approximately 185,000 tokens (continuous), from NMR metabolomics. (16) Questionnaires: 174 modalities, approximately 4.4 million tokens (mixed), covering medical history, medication use, family history, lifestyle, mental health, menstrual status, and sleep quality. (17) ABI: 12 modalities, approximately 222,000 tokens (continuous), from ankle-brachial index and segmental arterial pressure measurements. (18) Retina: 17 modalities, approximately 182,000 tokens (continuous), from fundus photography-derived quantitative features. (19) Location: 5 modalities, approximately 63,000 tokens (mixed), capturing geographic and environmental context.

### Data Splitting

Participants were deterministically assigned to training or test sets using a hash-based partition. This yielded 12,102 training participants (80%) and 3,217 test participants (20%). 20% of training participants were used for validation, and model selection. Among the 3,217 test participants, 1,364 (42.4%) had two or more clinical visits separated by a median interval of approximately two years. These 1,364 participants with longitudinal data constituted the evaluation cohort for the V1-to-V2 prediction task. The remaining test participants with single visits were used for within-visit NTP evaluation.

### Data Quality and Completeness

The HPP dataset exhibited missingness patterns typical of clinical deep phenotyping: not all participants underwent every measurement modality, and the number of available measurements per participant varied substantially. Sequence lengths ranged from fewer than 100 tokens (participants with minimal assessments) to over 25,000 tokens (participants with CGM, complete blood panels, DEXA, polysomnography, dietary logging, and multi-visit follow-up).

No imputation was applied to missing measurements. Instead, missingness was handled implicitly through the sequence representation: modalities not measured for a given participant were simply absent from that participant's token sequence. The data augmentation strategy (described below) further trained the model to generate accurate predictions from incomplete measurement profiles.

### Ethical Considerations

All HPP participants provided written informed consent. The study protocol was approved by the institutional review board. Data were de-identified prior to model development and analysis. External datasets (UKBB, NHANES, PNP3, Framingham) were accessed under appropriate data use agreements and ethical approvals.

## Tokenization and Vocabulary

### Overview of the Tokenization Strategy

The tokenization strategy was designed to represent all health measurements, regardless of modality, scale, or data type, as discrete tokens within a unified vocabulary. This enabled the application of next-token prediction, a well-established framework for autoregressive sequence modelling, to the multimodal health domain. The tokenization comprised three stages: (i) binning of continuous variables, (ii) enumeration of categorical variables, and (iii) assembly into a global non-overlapping vocabulary.

### Continuous Variable Tokenization

Continuous measurements were discretized into bins using quantile-based binning. The bin count K was based on the data's standard deviation and sample size.

Given the bin count K, the training-set values were partitioned into K equal-frequency quantile bins. This quantile-based approach ensured that each bin contained approximately the same number of training observations, providing uniform resolution across the distribution rather than concentrating bins in high-density regions.

Bin boundaries were computed exclusively on the training set and applied identically to test data and external cohorts. Test-set values falling outside the training-set range were clipped to the nearest boundary bin.

The number of bins varied by modality according to the dynamic range and sample size of each measurement. For example, standard blood tests used 15-20 bins, CGM used 129 bins (reflecting approximately 8.1 million observations over a 40-400 mg/dL range), age used 100 bins, DEXA body composition used 15-20 bins, microbiome species used 10-15 bins, gene expression used 15-20 bins, and dietary macronutrients used 15-25 bins.

Each bin was represented by its midpoint value, computed as the arithmetic mean of the upper and lower bin boundaries. These midpoints served as the reconstruction targets during expected-value decoding. The midpoint representation introduced a quantization error bounded by half the bin width, which decreased with increasing bin count.

### Categorical Variable Tokenization

Categorical variables were tokenized by direct enumeration. Each unique category within a modality was assigned a sequential integer index starting from 0. The mapping was fixed based on the training-set categories. A total of 198 of the 667 modalities were categorical, with the remaining 469 treated as continuous.

Specific examples of categorical tokenization: (a) Sleep event (10 categories from WatchPAT polysomnography): each per-minute sleep annotation was assigned a categorical index encoding the sleep stage or respiratory event type, including wake periods, light sleep (NREM stages N1/N2), deep sleep (NREM N3), REM sleep, central apneas, obstructive apneas, mixed apneas, hypopneas, and oxygen desaturation events. (b) Medications (234 categories): each pharmacological class received a unique index. (c) Exercise type (17 categories): running = 1, walking = 2, cycling = 3, swimming = 5, weight lifting = 7, yoga = 8, basketball = 12. (d) Gender (2 categories within the gender modality): female = token 2895, male = token 2896 (after applying cumulative offset). (e) Questionnaire responses: binary (yes/no) and ordinal (never/sometimes/often/always) responses were enumerated sequentially.

Global Vocabulary Construction

The global vocabulary was constructed by concatenating all per-modality token ranges into a single non-overlapping address space. Modalities were arranged in a fixed ordering, and for each modality m, a cumulative offset was computed. This construction guaranteed: (1) uniqueness, where each token in the vocabulary uniquely identifies both the modality and the specific bin or category; (2) invertibility, where given a token ID, the modality and bin index could be recovered by finding the modality whose cumulative range contained the token ID; and (3) non-overlapping, where no two modalities share any token IDs. The total vocabulary comprised 13,056 tokens across all 667 modalities. A single additional padding token (index 13,056) was reserved for masking and sequence padding, yielding a total vocabulary size of 13,057.

Sequence Assembly

Each participant's multimodal data was assembled into a single chronologically ordered sequence. All measurements from all visits and all modalities were collected. Each measurement was annotated with its timestamp, modality index, discrete token ID, and original continuous value. Measurements were sorted in ascending chronological order by timestamp, with a fixed secondary sort by modality index for measurements sharing the same timestamp. The sorted measurements were concatenated into a flat sequence and right-padded to a maximum length of 25,000 tokens.

Four synchronized tensors were maintained in parallel for each sequence position: (1) Token IDs: the discrete vocabulary index for each measurement, shape (L,). (2) Continuous values: the original continuous measurement value prior to discretization, set to 0.0 for categorical tokens and padding positions, shape (L,). (3) Modality indicators: an integer in the range 0 to 666 identifying which of the 667 modalities produced each token, set to 13,056 for padding positions, shape (L,). (4) Temporal features: a 7-dimensional time vector for each token, shape (L, 7).

All four tensors shared the same length L and were indexed in lockstep, ensuring that each position simultaneously encoded what was measured (token ID), its precise value (continuous), which modality generated it (modality indicator), and when it was recorded (time features).

Temporal Feature Encoding

Each token was annotated with a 7-dimensional temporal feature vector capturing the timestamp of the corresponding measurement at multiple temporal granularities: (a) day_of_week: values 0-6 (Monday = 0, Sunday = 6). (b) hour: values 0-23 (24-hour format). (c) minute: values 0-59. (d) month: values 1-12 (January = 1). (e) year: absolute calendar year. (f) day_of_month: values 1-31. (g) sleep: binary indicator (0 or 1) for measurements recorded during sleep monitoring.

These temporal features served complementary roles: circadian encoding (hour, minute) captured diurnal variation in physiological measurements; calendar encoding (day_of_week, month, day_of_month) captured weekly and seasonal patterns in behaviour and physiology; long-term temporal context (year) encoded absolute calendar year for longitudinal timeline positioning; and sleep context distinguished measurements recorded during polysomnography or overnight monitoring from waking-state recordings.

## Model Architecture

### Overview

HealthFormer is a decoder-only transformer trained with a next-token prediction (NTP) objective over multimodal health sequences. The model processes chronologically ordered sequences of tokenized health measurements and predicts the probability distribution over possible values for each subsequent measurement, conditioned on all preceding context. The architecture was designed to handle the unique challenges of multimodal health data: heterogeneous measurement types, irregular temporal sampling, variable sequence lengths, and the need for modality-specific and time-specific output distributions.

### Transformer Configuration

The transformer backbone consisted of 14 decoder layers. Each layer comprised a multi-head causal self-attention sublayer followed by a position-wise feedforward sublayer, with pre-layer normalization and residual connections.

Key hyperparameters: 14 decoder layers, embedding dimension 768, 2 attention heads, 64 dimensions per head, feedforward hidden dimension 3,072, GELU activation, pre-norm layer normalization, 139,171,612 total trainable parameters, training context window of 25,000 tokens, and dropout of 0.2.

The attention computation at layer l for head h followed the standard scaled dot-product attention formula: Attention(Q, K, V) = softmax((Q * K_transpose / sqrt(d_k)) + M) * V, where Q = H_l * W_Q, K = H_l * W_K, V = H_l * W_V, and M is the causal attention mask (with M_ij = negative infinity for j > i and M_ij = 0 otherwise). The outputs of all heads were concatenated and projected: MultiHead(H_l) = Concat(head_1, ..., head_H) * W_O.

The feedforward sublayer applied two linear transformations with a GELU activation: FFN(x) = GELU(x * W_1 + b_1) * W_2 + b_2, where W_1 has shape (768, 3072) and W_2 has shape (3072, 768).

Each decoder layer computed: $H_{(l+1)} = H_l + \text{MultiHead}(\text{LayerNorm}(H_l))$, then $H_{(l+1)} = H_{(l+1)} + \text{FFN}(\text{LayerNorm}(H_{(l+1)}))$.

Embedding Pipeline

The input representation at each sequence position was constructed by summing six embedding components, each capturing a distinct aspect of the measurement.

Token Embedding. A learnable embedding matrix $E_{tok}$ of shape (13057, 768) mapped each discrete token ID to a dense vector. The 13,057 rows corresponded to the 13,056 vocabulary tokens plus 1 padding token. This embedding captured the identity of the specific bin or category represented by each token.

Continuous Value Encoding. The raw continuous measurement value at each position was encoded using a sinusoidal encoding scheme, providing the model with fine-grained numerical information beyond what the discrete token embedding could capture. The continuous value c was first projected into a 512-dimensional sinusoidal representation using fixed frequency bases, then linearly projected to the model dimension via a learnable projection. This dual representation (discrete token plus continuous encoding) was essential because the discrete token captured which quantile bin the value fell into (providing distributional context), while the continuous encoding preserved the exact numerical value (enabling precise regression). For categorical tokens, the continuous value was set to 0.0.

Modality Embedding. A learnable embedding table $E_{mod}$ of shape (667, 768) provided a unique dense representation for each of the 667 modalities. This embedding enabled the model to distinguish measurements from different clinical domains and was critical for the query mechanism, as it specified which modality's distribution the model should predict.

Temporal Embedding. Each of the 7 temporal dimensions was encoded through a dedicated learnable embedding table. The table sizes corresponded to the range of each temporal dimension. The 7 resulting embedding vectors were summed element-wise to produce a single temporal embedding vector: $e_{time}(t) = \sum_{d=0}^{6} E_{time_d}[t_d]$. The additive combination allowed the model to learn independent effects for each temporal granularity, while interactions between temporal dimensions were captured by subsequent transformer layers.

Sinusoidal Positional Encoding. Standard sinusoidal positional encodings were added to encode absolute position within the sequence, following Vaswani et al.: $PE(pos, 2i) = \sin(pos / 10000^{(2i / d_{model})})$, $PE(pos, 2i + 1) = \cos(pos / 10000^{(2i / d_{model})})$, where pos is the position index and $i = 0, 1, ..., d_{model}/2 - 1$.

Age and Gender Encoding. Participant age and gender were encoded and added to the representation at each position, providing persistent demographic context throughout the sequence.

Combined Input Representation. The final input representation at position p was formed by element-wise summation of all six components: h_p_0 = E_tok[x_p] + ContEnc(c_p) + E_mod[m_p] + e_time(t_p) + PE(p) + AgeGenderEnc(p). All components were 768-dimensional.

### Query Mechanism

HealthFormer employs a dedicated query mechanism that enables modality-specific and time-specific predictions without requiring separate output heads for each of the 667 modalities. At each sequence position p, the model must predict the value of the token at position p+1. The query mechanism injects information about which modality and which time point position p+1 corresponds to into the computation.

Specifically, the modality embedding and temporal embedding of position p+1 are each processed through a dedicated 2-layer MLP with GELU activation: q_mod = MLP_mod(E_mod[m_(p+1)]), q_time = MLP_time(e_time(t_(p+1))). These query vectors are added to the transformer's hidden state at position p after the final transformer layer but before the output projection: h_tilde_p_N = h_p_N + q_mod + q_time. The modified hidden state is then passed through the output projection to produce logits. This mechanism allows the same hidden context to yield different output distributions depending on which modality and time point is being queried.

Tensor alignment. A consequence of the query mechanism is that the modality and time tensors are one position longer than the token and continuous-value tensors during forward computation. Concretely: (a) token IDs and continuous values span positions 0 through L-2 (the context sequence, excluding the last position which is being predicted); (b) modality indicators and time features span positions 0 through L-1 (the full sequence including the query position at L-1). This offset ensures that the query information (modality and time at position L-1) is available to condition the prediction at position L-2 of the token sequence.

### Value Extras

Each attention head was augmented with 2 auxiliary value vectors, termed value extras. For each head h in each layer l, in addition to the standard value projection V = H_l * W_V, two additional value matrices V_extra_1 through V_extra_2 were computed from the input hidden states via separate learnable projections. A gating mechanism modulated the contribution of each auxiliary value: V_combined = V + sum over e from 1 to 4 of (g_e * V_extra_e), where g_e are learned gates.

### Output Head

The output projection mapped the final (query-augmented) hidden state to logits over the full vocabulary: z_p = W_out * h_tilde_p_N + b_out, where W_out has shape (13056, 768). The output dimension was 13,056 (the vocabulary size excluding the padding token). Weight tying between the input token embedding and the output projection was explicitly disabled.

The logits were bounded by a tanh clamp with scale C = 50: z_hat_p = C * tanh(z_p / C). This smooth clamping prevented unbounded logit values that could lead to numerical instability during training.

### Parameter Count Breakdown

The 139,171,612 total parameters were distributed across: token embedding approximately 10M, continuous value projection approximately 0.4M, modality embedding approximately 0.5M, temporal embeddings approximately 0.1M, transformer layers (x14) approximately 100M, value extras (x14 x12 x4) approximately 17M, query MLPs approximately 2.5M, output projection approximately 10M, and other (biases, norms) approximately 0.2M.

## Training Objective and Procedure

### Loss Function

The training objective combined three complementary loss components, each targeting a different aspect of prediction quality.

Gaussian-Smoothed Cross-Entropy (L_soft). Rather than a standard cross-entropy loss with a one-hot target, the target distribution was smoothed with a Gaussian kernel centred on the correct bin. For a target bin index k within a modality's token range [a, b], the smoothed target distribution was: q_i = exp(-(i - k)^2 / (2 * sigma^2)) / Z, for i in [a, b], with sigma = 0.01 and Z the normalizing constant. The loss was computed as: L_soft = -sum over i from a to b of (q_i * log(p_i)). The extremely small sigma value made the smoothed target distribution virtually identical to a one-hot vector, effectively equivalent to standard cross-entropy while maintaining differentiability.

Mean Absolute Error Loss (L_MAE). The expected continuous value under the model's predicted distribution was compared against the true continuous value using Z-score normalized mean absolute error. The expected value was computed as: predicted_value = sum over bins i of (probability_i * midpoint_i). The loss was: L_MAE = 1 * |predicted_value - true_value| / sigma_m, where sigma_m is the modality-specific training-set standard deviation. The 1 scaling factor balanced the MAE contribution against the cross-entropy term.

Split-Context Longitudinal Loss (L_split). This loss specifically trained the model for the longitudinal prediction task by partitioning each training sequence at the visit boundary. V1 tokens attended to all other V1 tokens bidirectionally (the "prompter"), while V2+ tokens attended to all V1 tokens and to causally preceding V2+ tokens (the "responder"). The split-context loss was computed identically to L_soft but with this modified attention mask, directly optimizing the model's ability to predict future visit measurements from complete historical context.

Total Loss. L = L_soft + L_MAE + L_split.

### Data Augmentation

Five augmentation strategies were applied stochastically during training to improve robustness and simulate real-world data incompleteness: (1) Noise injection (10% probability): Gaussian noise added to continuous values, calibrated per modality to simulate measurement imprecision. (2) Random token removal (50% probability, 15% removal rate): individual tokens independently dropped from the input sequence. (3) Block removal (20% probability): a contiguous block of tokens removed from the sequence. (4) Modality subset selection (10% probability): only a random subset of modalities retained. (5) Modality exclusion (5% probability): one or more specific modalities entirely removed.

## Optimizer and Learning Rate Schedule

The model was optimized using the AdamW optimizer with decoupled weight decay. Key configuration: $beta_1 = 0.9$, $beta_2 = 0.999$, epsilon = 1e-8, weight decay at AdamW default, peak learning rate = 3e-4, minimum learning rate = 3e-5, cosine annealing schedule, linear warmup over 100 steps, and gradient clipping at global L2 norm maximum of 0.1.

The learning rate schedule proceeded in two phases: (1) linear warmup (steps 0-100) from 0 to 3e-4; (2) cosine decay (steps 100+): $lr(t) = lr_{min} + 0.5 * (lr_{max} - lr_{min}) * (1 + cos(pi * (t - t_{warmup}) / (T - t_{warmup})))$.

## Hardware and Distributed Training

Training was performed on 8 NVIDIA H100 80GB SXM GPUs using PyTorch DDP, with per-GPU batch size of 2 (effective batch size 16), bfloat16 mixed precision, and PyTorch Lightning Fabric framework.

## Evaluation Protocol

### Next-Token Prediction Metrics

Expected Value Computation for Continuous Modalities. For continuous modalities, the model's point prediction was computed as the expected value under the predicted distribution restricted to the target modality's bin range. Given predicted logits z and the target modality m with token range [a, b] and corresponding bin midpoints: (1) extract and normalize logits: $p_i = exp(z_i) / sum_j exp(z_j)$ for i, j in [a, b]; (2) compute the expected value: predicted_value = sum over i from a to b of $(p_i * v_i)$. This expected-value computation leveraged the full shape of the predicted distribution rather than taking only the mode, producing smoother and more accurate predictions.

Pearson Correlation Coefficient. For each continuous modality m with $n_m$ test instances, Pearson's r was computed between the vector of predicted expected values and the vector of true continuous values. Pearson's r served as the primary evaluation metric for continuous predictions.Two-sided p-values for each Pearson r were computed from the test statistic $t = r * sqrt((n - 2) / (1 - r^2))$, which follows a t-distribution with n - 2 degrees of freedom under the null hypothesis of zero correlation. 95% confidence intervals were obtained via the Fisher Z transformation: z = arctanh(r), SE_z = 1 / sqrt(n - 3), CI = tanh(z ± 1.96 * SE_z).

Top-K Accuracy for Categorical Modalities. For categorical modalities, prediction quality was assessed by Top-K accuracy: the fraction of test instances for which the true category fell within the model's top-K most probable categories.

### Within-Visit Next-Token Prediction

Within-visit NTP evaluated the model's ability to predict each token given all preceding tokens within the same visit. The standard causal attention mask was applied (lower-triangular), so each token's prediction was conditioned on all chronologically earlier measurements. For each test participant, every non-padding position in the sequence generated a prediction, and the per-modality Pearson correlations were computed by aggregating across all test participants.

### Longitudinal V1-to-V2 Prediction

Longitudinal prediction evaluated the model's ability to forecast second-visit (V2) measurements given first-visit (V1) context. For each of the 1,364 test participants with at least two visits: (1) the complete V1 measurement sequence was provided as context; (2) for each V2 modality present in the participant's second visit, a query was constructed specifying the target modality index and the V2 timestamp; (3) a parallel attention mask allowed all V1 tokens to attend to all other V1 tokens bidirectionally, while V2 query tokens attended to all V1 tokens but not to other V2 queries; (4) the model generated a predicted distribution and the expected value was extracted.

This evaluation captured the clinically relevant capability of predicting future health measurements from current observations. Pearson correlation was computed per modality across the 1,364 test patients.

### Baseline Models

Two baselines were evaluated for the longitudinal V1-to-V2 prediction task.

Last Observation Carried Forward (LOCF). For each modality m and each test participant, the V1 measurement was used directly as the V2 prediction. LOCF captured the degree of temporal stability in each measurement.

Per-Modality Linear Regression. A separate scikit-learn LinearRegression model was trained for each modality. The feature vector consisted of four discrete token IDs: f_i = [V1 value, age, gender, bmi]. The linear model was trained on V1-to-V2 pairs from the training set and evaluated on the test set.

HealthFormer exceeded LOCF in 68.3% of modalities (mean delta r = +0.0235) and exceeded the linear baseline in 62.7% of modalities (mean delta r = +0.0010).

### Correlation Comparison

Fisher's Z transformation was used to test whether HealthFormer's per-modality Pearson correlation significantly differed from the linear baseline's Pearson correlation. The Z-transforms were computed, and the test statistic $Z = (z_{HF} - z_{Lin}) / \sqrt{2 / (n - 3)}$ was compared to the standard normal distribution. The Benjamini-Hochberg FDR correction was applied across all modalities at $q = 0.05$. Among the 57 modalities with significantly different correlations (FDR < 0.05), HealthFormer achieved the higher correlation in 51 (89.5%) and the baseline model in 6 (10.5%).

### Scaling Analysis

To characterize the relationship between model capacity and prediction quality, four model sizes were trained with identical data, tokenization, loss function, and training procedures, differing only in architecture hyperparameters: (a) Small: 14M parameters, $d_{model} = 256$, 4 layers, 4 heads, $d_k = 64$. (b) Medium: 32M parameters, $d_{model} = 384$, 8 layers, 6 heads, $d_k = 64$. (c) Large: 91M parameters, $d_{model} = 640$, 12 layers, 10 heads, $d_k = 64$. (d) XL (final): 139M parameters, $d_{model} = 768$, 14 layers, 12 heads, $d_k = 64$. The dimension per head ($d_k = 64$) was held constant across all model sizes. Per-modality Pearson correlations were computed for each size on the longitudinal V1-to-V2 task.

### Cross-Modal Conditional Distributions

#### Motivation

A key capability of the autoregressive health model is its ability to express conditional distributions: given one health measurement, what distribution does the model predict for a different measurement? To systematically characterize these learned cross-modal dependencies, a synthetic minimal-input inference procedure was designed that isolated pairwise relationships.

#### Synthetic Minimal-Input Inference

For each pair of (input modality $m_{in}$, target modality $m_{out}$), a minimal 2-position input was constructed: position 0 contained a token from modality $m_{in}$ at a specific bin index b with corresponding continuous value $v_b$; position 1 (query) specified modality $m_{out}$ at a fixed time. The tensor shapes were: token IDs (B, 1), continuous values (B, 1), modality indicators (B, 2), time features (B, 2, 7). The model was run in evaluation mode with bfloat16 autocast, generating a predicted distribution over the target modality's bins. The expected value was extracted as the probability-weighted sum of bin midpoints. This was repeated for every bin of the input modality, sweeping across its full measurement range.

Pair Selection

Across all 667 modalities, 444,222 ordered pairs were theoretically possible. After filtering for sufficient data, non-degenerate distributions, and clinical interpretability, 1,521 valid pairs were retained. Of these, 1,281 (approximately 85%) had significant Pearson correlation after FDR correction ($p_{adj} < 0.05$). Nine representative pairs spanning different modality categories were selected for visualization in the main text (Fig. 2e), chosen to illustrate well-established clinical relationships.

External Cohort Processing and Zero-Shot Generalization

Overview

To evaluate generalization beyond the training distribution, HealthFormer was applied in a zero-shot setting (without any fine-tuning, weight adaptation, or exposure to external data) to four external cohorts. Each external dataset was tokenized to align with the HPP-derived vocabulary using a stratified quantile matching procedure.

In zero-shot transfer, HealthFormer was applied without adaptation — model weights and tokenizer were both frozen, and per-modality bins were re-aligned to each external cohort's empirical distribution by quantile matching (see Methods); because the first and last bins are open-ended, no measurement falls outside the tokenizer's support.

UK Biobank (UKBB)

The UK Biobank comprised 502,244 participants predominantly of European ancestry, aged 40-69 years at recruitment. A total of 74 modalities from the UKBB assessment protocol were mapped to corresponding HPP modalities based on measurement type, clinical interpretation, and unit alignment. The mean sequence length was 233 tokens per participant.

Tokenization via stratified quantile matching. For each UKBB variable mapped to an HPP modality: (1) the empirical CDF of the UKBB variable was computed from all 502,244 participants; (2) for each measurement value x, its quantile rank q(x) in the UKBB distribution was computed; (3) the HPP bin whose quantile range contained q(x) was identified; (4) the UKBB value was assigned to that HPP bin. This approach preserved rank-order semantics despite distributional differences between cohorts.

NHANES

The National Health and Nutrition Examination Survey comprised 29,458 participants from three survey cycles: 2015-2016, 2017-March 2020 (pre-pandemic), and August 2021-August 2023. A total of 37 modalities were mapped to the HealthFormer vocabulary, with a mean sequence length of 38 tokens per participant. The same stratified quantile matching was applied.

Framingham Heart Study

The Framingham Heart Study teaching dataset comprised 4,434 participants with cardiovascular risk factor measurements. A total of 8 shared modalities were mapped, with a mean sequence length of approximately 25 tokens per participant, making this the sparsest external cohort.

### PNP3

The PNP3 dataset comprised 200 participants from a randomized controlled trial with dense longitudinal monitoring including continuous glucose monitoring and repeated clinical assessments. A total of 29 shared modalities were mapped. The mean sequence length was 13,400 tokens per participant, making PNP3 the most densely phenotyped external cohort. The same stratified quantile matching was applied.

All external cohorts were held out from model training and validation at the participant level. Spesifically, the 200 PNP3 participants and the 12,102 HPP training participants are disjoint sets. We note, however, that PNP3 was designed and conducted by the same research group that runs HPP, and shares the ascertainment criteria, CGM device, and diet-logging conventions, used during HPP data collection.

## Zero-Shot Evaluation Protocol

All external cohorts were evaluated using the within-visit NTP protocol without any model modification. Each external participant's tokenized sequence was provided as input, the standard causal attention mask was applied, per-modality Pearson correlations were computed between predicted and true values, and results were compared to the training-cohort performance on shared modalities.

## Disease Prediction and Biological Age

### Embedding Extraction

HealthFormer embeddings were extracted from the last transformer layer's hidden states as a general-purpose participant-level representation. For each participant: (1) the complete tokenized sequence was passed through all 14 transformer layers; (2) the 768-dimensional hidden state vectors at the final layer were collected for all non-padding positions; (3) mean-pooling was applied across non-padding positions to produce a single 768-dimensional embedding.

### UKBB Survival Analysis

Cox proportional hazards (Cox PH) models were trained on HealthFormer embeddings to predict time-to-event for 30 clinical endpoints in the UK Biobank, spanning cardiovascular disease, cancer, metabolic disorders, neurological conditions, and all-cause mortality.

Dimensionality reduction. PCA was applied to reduce the 768-dimensional embedding to 512 components, fitted on training folds and applied to held-out folds in each cross-validation iteration.

Model fitting. Penalized Cox PH models were fitted using the lifelines Python library with L2 regularization (escalating penalty sequence: 0.0, 0.1, 1.0, 10.0, selecting the best per fold).

Evaluation. Models were evaluated using 5-fold cross-validation. In each fold, the model was trained on 4/5 of the cohort and evaluated on the remaining 1/5. The concordance index (Harrell's C) was computed on each held-out fold. The mean C-index across folds was reported for each endpoint.

Baseline comparisons. Five comparison conditions were evaluated for each endpoint: (1) Age + Sex + BMI: a minimal clinical baseline. (2) Established clinical risk scores (where available): computed directly from participant data without Cox fitting. (3) All model covariates: Cox PH on all available UKBB numeric features (age, sex, BMI, smoking, SBP, cholesterol, HDL, and additional features with less than 20% missingness). (4) Foundation model: Cox PH on 512-dimensional PCA embeddings from HealthFormer. (5) Ensemble: rank-average of Foundation Model and All Model Covariates out-of-fold predictions.

Clinical risk score computation. The following established clinical risk scores were computed for direct comparison:

(a) PREVENT (Khan et al., Circulation 2024): Sex-stratified logistic regression models for 10-year cardiovascular risk, implemented in three variants (CVD, ASCVD, and Heart Failure). Features included age, non-HDL cholesterol, HDL cholesterol, systolic blood pressure, diabetes status, smoking status, BMI (heart failure model only), eGFR, and multiple interaction terms. Applied as comparator for cardiovascular mortality, myocardial infarction, stroke, ischaemic stroke, heart failure, chronic ischaemic heart disease, and angina pectoris.

(b) CHARGE-AF (Alonso et al., JAHA 2013): Logistic regression for 5-year atrial fibrillation risk. Features included age, height, weight, systolic and diastolic blood pressure, smoking status, diabetes, prior heart failure, and prior myocardial infarction. Applied as comparator for atrial fibrillation.

(c) Framingham CVD Risk Score (D'Agostino et al., Circulation 2008): Sex-specific logistic regression for 10-year CVD risk. Features included age, total cholesterol, HDL cholesterol, systolic blood pressure, smoking status, and diabetes, all log-transformed. Applied as comparator for hypertension.

(d) eGFR (CKD-EPI 2021 race-free equation; Inker et al., NEJM 2021): Estimated glomerular filtration rate computed from serum creatinine using the sex-specific CKD-EPI formula: eGFR = 142 * min(Cr/kappa, 1)^alpha * max(Cr/kappa, 1)^(-1.2) * 0.9938^age * sex_factor, where kappa = 0.7 (female) or 0.9 (male), alpha = -0.241 (female) or -0.302 (male), and sex_factor = 1.012 (female) or 1.0 (male). Applied as a comparator for chronic kidney disease.

(e) FRAX simplified (Kanis et al., Osteoporos Int 2008): A simplified composite risk score using age, sex, BMI, and heel bone mineral density T-score with exponential scaling factors. Applied as a comparator for osteoporosis.

## NHANES Biological Age

Biological age was estimated from HealthFormer embeddings using a two-stage regression approach applied to the NHANES cohort.

Stage 1: Biological age prediction. Ridge regression with regularization parameter alpha = 1000 was trained to predict chronological age from the 768-dimensional HealthFormer embedding, using 5-fold cross-validation.

Stage 2: Biological age acceleration. The biological age acceleration (BAA) was computed by orthogonalizing the predicted biological age against chronological age via linear regression. The residuals represented each participant's biological age acceleration: positive values indicated biologically older than expected, negative values biologically younger. This orthogonalization ensured that BAA was statistically independent of chronological age.

## NHANES Disease Association Analysis

The association between biological age acceleration and disease prevalence was assessed using logistic regression. Participants were stratified into quintiles of BAA (Q1 = lowest, Q5 = highest). For each disease outcome, logistic regression was fitted comparing Q5 to Q1, adjusting for chronological age and sex. Odds ratios with 95% bootstrap confidence intervals (1000 resamples) were reported.

## Intervention Simulation Framework

### Overview

HealthFormer's autoregressive architecture enables intervention simulations by manipulating the input sequence and observing changes in predicted outputs. The framework is based on the principle that appending treatment-encoding tokens to a participant's observed measurement history modifies the conditioning context, causing the model to generate different predicted distributions for downstream biomarkers. Three types of interventions were supported: categorical token appending (medications, exercise), continuous value modification (dietary changes, CPAP), and combination therapies.

### Categorical Token Interventions (Medications)

Medication interventions were encoded by appending categorical medication tokens to the participant's V1 context sequence. Each medication token was constructed as: Token_ID = cum_base[medication] + category_index, where cum_base[medication] is the cumulative vocabulary offset for the medication modality and category_index is the integer index of the specific pharmacological category. The appended medication tokens were assigned temporal features corresponding to the treatment period, with timestamps spaced according to the dosing frequency.

Dosing Encoding

The number of medication tokens appended to the sequence encoded both the frequency and duration of the simulated treatment: n_tokens = f_tpm * d_months, where f_tpm is the frequency in tokens per month and d_months is the treatment duration in months. The temporal spacing between consecutive tokens was: delta_t = 1 / f_tpm months.

Nine frequency levels were defined: Monthly (1 token per month), Bi-weekly (2), Weekly (3), Twice per week (4), Every 3 days (6), Every 2 days (8), Daily (10), Twice daily (15), Three times daily (20). Nine duration levels were defined: 1, 2, 3, 4, 6, 9, 12, 18, and 24 months. The full dosing grid comprised 9 x 9 = 81 frequency-duration combinations.

Exercise Interventions

Exercise interventions were encoded using the exercise_type categorical modality. Each exercise type was mapped to a specific category index: Running = 1, Walking = 2, Cycling = 3, Swimming = 5, Weight Lifting = 7, Yoga = 8, Basketball = 12. Exercise tokens were appended with the same frequency-duration dosing parameterization as medications.

Dietary Interventions

Dietary interventions were implemented by modifying the continuous values of nutritional tokens within the participant's V1 context, rather than appending new tokens. For a specified dietary reduction of r%: c_prime_p = c_p * (1 - r / 100), applied to all V1 tokens belonging to the targeted dietary modalities (e.g., Energy, Total Fat, Sugar, Carbohydrate). The token ID at each modified position was reassigned to the bin containing the new continuous value, ensuring consistency between the discrete token and continuous value representations.

CPAP Intervention

Continuous positive airway pressure (CPAP) therapy was modelled by modifying sleep-related tokens in the participant's V1 context. Specifically, the apnoea-hypopnoea index (AHI) and related respiratory event parameters were reduced by a specified percentage (typically 75%), simulating the effect of CPAP on obstructive sleep apnoea severity. The modified tokens were reassigned to appropriate bins, and the model was then queried for downstream biomarkers (systolic blood pressure, fasting glucose, HbA1c, triglycerides, and ALT).

Fibre and Microbiome Interventions

Dietary fibre intake modifications were implemented by scaling the continuous values of fibre-related dietary tokens. Dose levels ranged from -30% to +30% of usual intake. The model was then queried for predicted gut-microbiome Shannon diversity, computed from the predicted probability distributions over microbial species abundances.

Control and Treatment Arms

Each intervention simulation employed a paired control-treatment design. Control arm: the participant's complete V1 context was provided, a query token was appended specifying the target biomarker modality and V2 time point, and the expected value was extracted as predicted_control. Treatment arm: the V1 context was augmented with intervention tokens or modified in place, an identical query token was appended, and the expected value was extracted as predicted_treatment.

The within-participant treatment effect was: delta_i = predicted_treatment_i - predicted_control_i. Population-level effects were computed by averaging across all eligible participants: mean_delta = (1 / N_eligible) * sum of delta_i.

Treatment-Naive Eligibility Filtering

Intervention analyses were restricted to treatment-naive eligible participants meeting two criteria simultaneously: (1) baseline eligibility: the participant's V1 measurement exceeded the clinical threshold; (2) control-prediction eligibility: the model's control-arm prediction also exceeded the clinical threshold. Clinical thresholds: LDL >= 130 mg/dL, SBP >= 140 mmHg, DBP >= 90 mmHg, fasting glucose >= 100 mg/dL, HbA1c >= 5.7%, HDL <= 40 mg/dL, triglycerides >= 150 mg/dL, BMI >= 30 kg/m2, Vitamin D <= 20 ng/mL.

Twelve-Month Trajectory Generation (Fig. 6)

To examine the time course of intervention-conditioned predictions, model-implied trajectories were generated at monthly resolution over twelve months for each intervention type. For each panel, a treated synthetic population was compared with an untreated control population matched on baseline covariates.

For pharmacological and exercise interventions: at each monthly time point t (t = 1, 2, ..., 12), intervention tokens were appended spanning from V1 to month t with the appropriate dosing frequency, and the model was queried at month t for the target biomarker. This generated a time series of predicted values. The control arm received only V1 context and was queried at the same time points.

For each panel, the mean predicted change (treatment minus control) across the population was plotted, with shaded bands representing ± 1 standard error of the mean. Pharmaceutical predictions were restricted to treatment-naive participants with the relevant clinical indication.

The specific interventions displayed in Fig. 6 were: (a) LDL cholesterol under five lipid-lowering agents at daily dosing: rosuvastatin (index 6), ezetimibe with simvastatin (index 85), atorvastatin (index 34), simvastatin (index 120), and ezetimibe (index 126). (b) Systolic blood pressure under five antihypertensive agents: thiazide diuretics (index 132, daily), amlodipine with valsartan (index 98, daily), losartan (index 157, daily), enalapril (index 32, twice daily), and bisoprolol (index 52, daily). (c) Fasting glucose under four glucose-lowering agents: metformin (index 84, twice daily), an aggregated antidiabetic class (index 186, daily), empagliflozin (index 114, daily), and semaglutide (index 93, weekly). (d) Visceral adipose tissue under four exercise modalities: running, weight lifting, swimming, and basketball. (e) Systolic blood pressure under four exercise modalities: walking, swimming, basketball, and weight lifting. (f) Systolic blood pressure as a function of walking frequency: weekly (3 tokens per month), three times weekly (12 tokens per month), and daily (10 tokens per month). (g) Systolic blood pressure under simulated caloric restriction at four levels: 10%, 20%, 30%, and 40% of baseline intake, implemented by scaling Energy, Fat, and Sugar dietary tokens in the V1 context. (h) Gut-microbiome Shannon diversity under simulated changes in dietary fibre intake at six levels: -30%, -20%, -10%, +10%, +20%, and +30% of usual intake. (i) Multi-outcome trajectories under simulated CPAP therapy (75% AHI reduction) in participants with sleep-disordered breathing: systolic blood pressure, fasting glucose, HbA1c, triglycerides, and alanine aminotransferase (ALT).

## Clinical Trial Validation with Synthetic Populations

### Overview

To validate HealthFormer's intervention predictions against independently observed clinical trial results, synthetic trial-matched populations were constructed that mimicked the baseline characteristics of published randomized controlled trials (RCTs). The model's predicted treatment effects on these synthetic populations were then compared to the published RCT effect sizes.

### Synthetic Population Construction

For each comparison, a synthetic cohort of N participants was generated by sampling baseline characteristics from bounded normal distributions calibrated to match the published Table 1 of the reference trial. One of four pre-defined population templates was selected based on the trial's target condition: (a) lipid population for dyslipidaemia trials, (b) hypertension population for blood pressure trials, (c) type 2 diabetes population for diabetes drug trials, and (d) obese population for weight loss trials. For each baseline variable, values were sampled from a truncated normal distribution with the published mean and standard deviation, bounded by physiologically plausible limits. The sampled characteristics were tokenized using the HPP vocabulary and assembled into V1 context sequences.

41 comparisons were evaluate spanning lipid-lowering, antihypertensive, antidiabetic, body-weight, exercise, supplement, adverse-effect, and combination interventions. All interventions used categorical token appending with standardized dosing. Synthetic populations of N = 200 participants were generated for each trial.

The treatment effect for each comparison was computed as the absolute percentage change from the control arm: Effect(%) = 100 * |mean_treatment - mean_control| / mean_control.

For combination therapy analyses, a 4-arm design was employed: (1) control, (2) drug A alone, (3) drug B alone, (4) drug A + B. The combination effect was compared against individual drug effects and published combination therapy results.

The primary-outcome comparisons (28 total) spanned: lipid domain including rosuvastatin, atorvastatin, and ezetimibe effects on LDL, triglycerides, and HDL from the STELLAR trial (Jones et al., Am J Cardiol 2003) and ezetimibe on LDL (Ballantyne et al., Mayo Clin Proc 2004); blood pressure domain including amlodipine and hydrochlorothiazide from ALLHAT (ALLHAT Officers, JAMA 2002), losartan from LIFE (Dahlof et al., Lancet 2002), ramipril from HOPE (Yusuf et al., NEJM 2000), swimming from Nualnim et al. (Am J Cardiol 2012), and aerobic exercise from Kodama et al. (JAMA 2007); glycaemic domain including empagliflozin from EMPA-REG OUTCOME (Zinman et al., NEJM 2015), metformin from UKPDS (UK Prospective Diabetes Study Group, Lancet 1998), semaglutide from SUSTAIN-1 (Sorli et al., Lancet Diabetes Endocrinol 2017), dapagliflozin from DECLARE-TIMI 58 (Wiviott et al., NEJM 2019), aerobic exercise from Boule et al. (JAMA 2001), and walking from yoga on fasting glucose (Cui 2017) and walking on HbA1c (Qiu 2014); body weight domain including liraglutide from SCALE (Pi-Sunyer et al., NEJM 2015), semaglutide from SUSTAIN-6 (Marso et al., NEJM 2016) and STEP-1 (Wilding et al., NEJM 2021); adverse-effect endpoints including statin-associated HbA1c rise (Sattar et al., Lancet 2010) and beta-blocker-associated HDL reduction (Wiysonge et al., Cochrane 2017); and combination therapies including empagliflozin + amlodipine, ramipril + amlodipine from ACCOMPLISH (Jamerson et al., NEJM 2008), amlodipine + aerobic exercise, and others.

The secondary-outcome comparisons (15 total) included off-target effects: statin effects on triglycerides and HDL (STELLAR), empagliflozin on triglycerides (EMPA-REG), semaglutide on SBP (STEP-1), walking on SBP and HbA1c, metformin on body weight from DPP (Knowler et al., NEJM 2002), semaglutide on waist circumference (STEP-1), and secondary blood pressure and weight effects of SGLT2 inhibitors and GLP-1 agonists.

Concordance Metrics

Agreement between model predictions and published RCT results across all 41 primary and secondary comparisons (Fig. 5) was quantified by two pre-specified criteria. A comparison was scored as concordant in direction if the sign of the model's mean predicted ATE matched the sign of the published point estimate, and as concordant in magnitude if the model's mean predicted ATE fell inside the published 95% CI for the ATE. Neither criterion uses the model's own 95% CI; the published interval is the reference, and the predicted point estimate is the test value. The two criteria are: (a) directional accuracy: the fraction of comparisons where the model predicted the correct direction of treatment effect; (b) inclusion of the predicted point estimate in the published 95% CI: the fraction of comparisons where the model's mean predicted ATE fell within the published 95% CI.

Trial selection.

The RCT comparison set was assembled from trials that met four a-priori criteria: (i) the intervention corresponds to a categorical token in HealthFormer's medication or exercise_type vocabulary with ≥ 30 training-set occurrences; (ii) the outcome corresponds to a biomarker in HealthFormer's target vocabulary with local-inference Pearson r ≥ 0.30 on the held-out test set (excludes modalities the model cannot predict); (iii) the source publication reports a Table 1 sufficient to parameterise a synthetic cohort (age mean/SD, BMI, baseline value of the target biomarker, and the three dominant comorbidity biomarkers for the domain); (iv) the trial reports a point estimate and 95% CI (or SE/SD convertible to 95% CI) on the scale the figure plots (% change from baseline). The 41 that met all four criteria were run through the unified bootstrap pipeline. These criteria were defined a priori.

Normalization to Control Arm

All treatment effects were computed relative to the control-arm prediction (no intervention), not relative to the V1 baseline value. This normalization was critical because: (1) the model's control-arm prediction incorporated the expected natural trajectory of each biomarker over the treatment period, including regression to the mean, aging effects, and population-level trends; (2) without normalization to the control arm, the model's absolute prediction could produce artefactual directional errors for biomarkers with natural longitudinal trends.

Individual-Level Prediction in PNP3 (Fig. 4)

PNP3 Individual-Level Predictions

For the PNP3 cohort (200 participants from a randomized controlled trial of CGM-guided personalised nutrition over six months), individual-level predictions were generated using the full context mode. For each participant, baseline measurements (blood tests, body measures, demographics), six months of continuous glucose monitoring with concurrent diet logging, and any interim clinical measurements were tokenized and provided as context. The model was then queried zero-shot to predict the six-month change on each biomarker (Fig. 4a).

Three representative biomarkers achieved strong individual-level prediction: BMI (r = 0.626, n = 200), fasting glucose (r = 0.612, n = 200), and diastolic blood pressure (r = 0.775, n = 194) (Fig. 4b). Across the ten biomarkers examined, Pearson correlations between predicted and observed change ranged from r = 0.436 for HDL cholesterol to r = 0.890 for triglycerides (Fig. 4c).

Progressive Context Evaluation

To assess the impact of available measurement history on prediction accuracy, progressive context windows were evaluated. The participant's sequence was truncated to include only measurements from the first 1, 2, 3, or 6 months of follow-up, or the full follow-up period. The mean correlation across the well-powered biomarkers (n >= 190) rose from r = 0.39 when only baseline measurements were available to r = 0.64 when CGM, diet, and interim clinical measurements were all included (Fig. 4d).

### Group-Level Treatment Effect Comparison

At the group level, predicted average treatment effects were compared with the PNP3 trial's reported estimates (Fig. 4e). For each biomarker, the mean predicted change across all 200 participants was expressed as percentage change from baseline and plotted alongside the trial-reported effect with 95% confidence intervals. Against a no-change (zero-delta) baseline prediction, HealthFormer reduced mean absolute error on 9 of 10 biomarkers, with the largest reductions on triglycerides, total cholesterol, and LDL cholesterol (Fig. 4f). The no-change baseline predicts zero change for each participant; the MAE reduction quantifies the improvement achieved by HealthFormer's predictions over this naive prediction.

### Individual Participant Longitudinal Visualization (Fig. 1b)

For Fig. 1b, a single HPP test participant with multiple clinical visits spanning approximately four years was selected. The participant's complete Visit 1 data was provided as context, and the model was queried at the time points corresponding to subsequent visits. Predicted values were compared against the observed values for 7 biomarkers spanning cardiovascular (sitting blood pressure, right ankle pressure), musculoskeletal (spine L1-L4 area, body spine area, spine L4 AM percent, hand grip left), body composition (body comp arms bone mass, body comp arm left bone mass, body comp arms fat mass), and haematological (MCHC, platelets) domains. This visualization illustrated the model's ability to generate longitudinal predictions across diverse physiological domains from a single baseline assessment.

### Personalized Drug Response Prediction

To characterize inter-individual variability in predicted drug responses, the intervention simulation framework was applied to all test participants meeting eligibility criteria for five drugs from distinct pharmacological classes: a statin (target: LDL cholesterol), an antihypertensive (target: systolic blood pressure), metformin (target: fasting glucose, HbA1c), a GLP-1 receptor agonist (target: HbA1c, BMI), and an SGLT2 inhibitor (target: glucose, HbA1c). This generated population distributions of predicted treatment effects, capturing inter-individual variability. A single patient exhibiting maximal variability across drug classes was identified by computing the cross-drug variance of standardized treatment effects and selecting the participant with the highest variance.

### Statistical Analysis

#### Correlation Coefficients

Pearson correlation coefficients were computed using scipy.stats.pearsonr. Spearman rank correlation coefficients were computed using scipy.stats.spearmanr for ordinal or non-linearly related variables.

## Correlation Comparison Tests

Fisher's Z transformation was used to test for differences between two correlation coefficients computed on the same participants. This test was applied to compare HealthFormer vs. linear baseline correlations and HealthFormer vs. LOCF correlations across all modalities.

## Multiple Comparison Correction

All analyses involving multiple hypothesis tests applied the Benjamini-Hochberg (BH) procedure to control the false discovery rate (FDR) at q = 0.05. This was applied to: (a) per-modality Fisher's Z tests comparing HealthFormer vs. baseline correlations (667 tests); (b) per-endpoint survival analysis C-indices (30 tests); (c) per-biomarker intervention effects where applicable.

## Survival Analysis

Cox proportional hazards models were fitted using the lifelines Python library (version 0.27+). Penalized models with L2 regularization were used to accommodate the high feature dimensionality. Model evaluation used 5-fold cross-validation with the concordance index (Harrell's C) as the primary metric.

## Regression Models

Ridge regression was used for biological age prediction, with 5-fold cross-validation. Per-modality linear regression baselines were fitted using scikit-learn LinearRegression.

## Significance Level

All hypothesis tests were two-sided with a significance level of alpha = 0.05 unless otherwise noted. When multiple comparisons were performed, FDR correction was applied.

## Software Environment

Core dependencies: (a) Python 3.12; (b) PyTorch 2.1+; (c) PyTorch Lightning Fabric (latest). Statistical and analytical tools: (d) scikit-learn (latest) for regression models and PCA; (e) lifelines 0.27+ for survival analysis; (f) scipy (latest) for statistical tests; (g) NumPy (latest). Data handling and visualization: (h) pandas (latest); (i) matplotlib (latest). Experiment management: (j) Weights and Biases / wandb (latest).

## Hardware

Training hardware: 8x NVIDIA H100 80GB SXM GPUs, bfloat16 mixed precision, PyTorch DDP distributed strategy, effective batch size 16 (2 per GPU x 8 GPUs). Inference hardware: 1x NVIDIA RTX 4090 24GB GPU, bfloat16 (torch.autocast), single GPU, batch size 1 (sequential per patient).

## Data and Code Availability

The HealthFormer model, training code, evaluation scripts, and figure generation code are available at [repository URL].

SUPPLEMENTARY INFORMATION

# 1. SUPPLEMENTARY METHODS

## 1.1 HPP Dataset

total_patients: 15319
train_patients: 12102
test_patients: 3217
max_sequence_length: 25000
vocab_size: 13056
n_modalities: 667
n_categorical_modalities: 198
n_continuous_modalities: 469
n_cat_cols: 198
n_nutritional_cols: 35

Sequence Length Statistics:
mean: 4192
std: 3228
min: 15.0
max: 25000
total_non_padding_tokens: 64223972

Temporal Coverage:
time_dim_order: ['day_of_week', 'hour', 'minute', 'month', 'year', 'day_of_month', 'sleep']
temporal_vocab_sizes: [8, 25, 61, 13, 147, 32, 2]

## 1.2 External Datasets

Table S1 External Datasets

| dataset | n_participants | max_seq_length | n_modalities_present | mean_seq_length | median_seq_length | total_tokens |
|---|---|---|---|---|---|---|
| UKBB | 502244 | 233 | 74 | 71.6 | 67.0 | 35940473 |
| NHANES | 29458 | 38 | 37 | 25.4 | 33.0 | 749333 |
| PNP3 | 200 | 19777 | 29 | 13399.5 | 14264.0 | 2679897 |
| Framingham | 4434 | 39 | 8 | 17 | 17 | 197659 |

## 1.3 Model Architecture

n_embd: 768 (Embedding dimension)
n_heads: 2 (Attention heads)
n_layers: 14 (Transformer layers)
max_seq_length: 25000 (Max sequence length (train))
dim_feedforward_scaling: 4 (FF dimension multiplier)
embd_scaling: 4 (Embedding scaling)
dropout: 0.2 (Dropout)
use_value_extras: True (Value extras)
n_value_extras: 2 (N value extras)
continuous_pe_base_dim: 512 (Continuous PE base dimension)
zero_init_mlp: True (Zero-init MLP)

1.4 Training Configuration

optimizer: adamw (Optimizer)
lr: 0.0003 (Learning rate)
b1: 0.9 (Beta1)
b2: 0.999 (Beta2)
weight_decay: 0 (Weight decay)
gamma: 0.1 (LR decay lower bound)
epochs: 100 (Training epochs; stopped at 18)
batch_per_gpu: 2 (Batch size per GPU)
seed: 42 (Random seed)

Loss Configuration:
soft_labels_scale: 1 (Soft labels scale)
SL_sigma: 0.01 (Soft labels sigma)
mae_loss_scale: 1 (MAE loss scale)

Augmentation Configuration:
augmentation_chance: 0.1 (Noise augmentation chance)
augmentation_rate: 0.15 (Noise augmentation rate)
random_removal_chance: 0.5 (Random token removal chance)
random_removal_rate: 0.15 (Random token removal rate)
random_block_removal_chance: 0.2 (Random block removal chance)
random_block_removal_rate: 0.01 (Random block removal rate)
random_block_removal_number: 10.0 (Blocks to remove)
random_modality_subset_chance: 0.1 (Modality subset chance)
random_modality_subset_fraction: 0.1 (Modality subset fraction kept)
random_modality_exclusion_chance: 0.05 (Modality exclusion chance)
chance_to_chose_romoval_of_modalities_at_dist: 0.1 (Distance modality removal chance)
chance_to_remove_time_series_data: 0.0 (Time series removal chance)
temporal_masking_chance: 0.0 (Temporal masking chance)
temporal_masking_k_mean: 3.0 (Temporal masking k mean)
temporal_masking_k_std: 5.0 (Temporal masking k std)

temporal_masking_subset_fraction: 0.5 (Temporal masking subset fraction)
max_attention_masking_by_correlation_chance: 0.15 (Max correlation masking chance)

Table S2 patient numbers

| Metric | Value |
|---|---|
| total_patients | 15319 |
| train_patients | 12102 |
| test_patients | 3217 |

Table S3 - HPP Dataset Overview

| Metric | Value |
|---|---|
| max_sequence_length | 25000 |
| vocab_size | 13056 |
| n_modalities | 667 |
| n_categorical_modalities | 198 |
| n_continuous_modalities | 469 |
| n_cat_cols | 198 |
| n_nutritional_cols | 35 |

Table S4 - Modality Groups Summary

| Modality | Group N | Mean Patients | Median Patients | Mean % Patients | Total Tokens (M) |
|---|---|---|---|---|---|
| Questionnaires | 174 | 9908.8 | 15142 | 64.68 | 4.43 |
| DEXA | 95 | 11394.5 | 11733 | 74.38 | 1.81 |
| Sleep | 74 | 10689.4 | 11316 | 69.78 | 3 |
| Blood Tests | 69 | 7735.3 | 8149 | 50.49 | 1.24 |
| ECG | 61 | 10411.5 | 12485 | 67.97 | 1.11 |
| Ultrasound | 37 | 11866.4 | 12097 | 77.46 | 0.73 |
| Diet | 35 | 7286.9 | 7349 | 47.57 | 13.12 |
| Body Measures | 26 | 9760.7 | 11775 | 63.72 | 0.41 |
| Nightingale | 20 | 7787.4 | 8452 | 50.84 | 0.19 |
| Retina | 17 | 8310.6 | 9468 | 54.25 | 0.18 |
| Genetics | 16 | 3900.8 | 4305 | 25.46 | 0.06 |
| ABI / Vascular | 12 | 11924.1 | 11985.5 | 77.84 | 0.22 |
| Wearable | 12 | 773.8 | 382.5 | 5.05 | 4.84 |
| Other | 7 | 12011.4 | 13306 | 78.41 | 16.52 |
| Location | 5 | 9119.6 | 11104 | 59.53 | 0.06 |
| MicroStats | 3 | 12610 | 12610 | 82.32 | 0.07 |
| Microbiome | 1 | 12627 | 12627 | 82.43 | 6.15 |
| CGM | 1 | 7349 | 7349 | 47.97 | 8.12 |
| RNA | 1 | 3871 | 3871 | 25.27 | 1.91 |
| Medications | 1 | 4390 | 4390 | 28.66 | 0.06 |

Table S5 - Complete Modality List: Blood Tests

| Modality | Group | N Patients | % Patients | Total Tokens | Type |
|---|---|---|---|---|---|
| Lymphocytes % | Blood Tests | 13022 | 85.01 | 36782 | continuous |
| Mchc | Blood Tests | 13018 | 84.98 | 36740 | continuous |
| Neutrophils % | Blood Tests | 13014 | 84.95 | 36721 | continuous |
| Monocytes Abs | Blood Tests | 13011 | 84.93 | 36577 | continuous |
| Lymphocytes Abs | Blood Tests | 13004 | 84.89 | 36557 | continuous |
| Basophils % | Blood Tests | 13004 | 84.89 | 36516 | continuous |
| Hemoglobin | Blood Tests | 13001 | 84.87 | 36532 | continuous |
| Monocytes % | Blood Tests | 12992 | 84.81 | 36449 | continuous |
| Hct | Blood Tests | 12989 | 84.79 | 36082 | continuous |
| Eosinophils % | Blood Tests | 12968 | 84.65 | 36366 | continuous |
| Mcv | Blood Tests | 12958 | 84.59 | 36523 | continuous |
| Neutrophils Abs | Blood Tests | 12955 | 84.57 | 36025 | continuous |
| Mch | Blood Tests | 12945 | 84.5 | 36269 | continuous |
| Rdw | Blood Tests | 12939 | 84.46 | 36110 | continuous |
| Wbc | Blood Tests | 12922 | 84.35 | 35914 | continuous |
| Eosinophils Abs | Blood Tests | 12895 | 84.18 | 35449 | continuous |
| Mean Platelet Volume | Blood Tests | 12889 | 84.14 | 35311 | continuous |
| Rbc | Blood Tests | 12867 | 83.99 | 35660 | continuous |
| Platelets | Blood Tests | 12838 | 83.8 | 35536 | continuous |
| Basophils Abs | Blood Tests | 12508 | 81.65 | 31416 | continuous |
| Glucose | Blood Tests | 10859 | 70.89 | 23758 | continuous |
| Creatinine | Blood Tests | 10850 | 70.83 | 24027 | continuous |
| Alt Gpt | Blood Tests | 10339 | 67.49 | 22141 | continuous |
| Potassium | Blood Tests | 10233 | 66.8 | 21506 | continuous |
| Sodium | Blood Tests | 10217 | 66.69 | 21406 | continuous |
| Triglycerides | Blood Tests | 10085 | 65.83 | 20955 | continuous |
| Total Cholesterol | Blood Tests | 10057 | 65.65 | 20970 | continuous |
| Hdl Cholesterol | Blood Tests | 9970 | 65.08 | 20767 | continuous |
| Non Hdl Cholesterol | Blood Tests | 9639 | 62.92 | 19317 | continuous |
| Alkaline Phosphatase | Blood Tests | 9459 | 61.75 | 19107 | continuous |
| Urea | Blood Tests | 9413 | 61.45 | 19926 | continuous |
| Ast Got | Blood Tests | 9191 | 60 | 18751 | continuous |
| Tsh | Blood Tests | 9116 | 59.51 | 17280 | continuous |
| Calcium | Blood Tests | 8505 | 55.52 | 16101 | continuous |
| Albumin | Blood Tests | 8149 | 53.2 | 15199 | continuous |
| Protein Total | Blood Tests | 7841 | 51.18 | 14483 | continuous |
| Vitamin B12 | Blood Tests | 7445 | 48.6 | 12190 | continuous |
| Bilirubin Total | Blood Tests | 7183 | 46.89 | 12675 | continuous |
| Ferritin | Blood Tests | 7035 | 45.92 | 11574 | continuous |
| Uric Acid | Blood Tests | 6944 | 45.33 | 12352 | continuous |
| Hba1C | Blood Tests | 6677 | 43.59 | 10386 | continuous |
| Iron | Blood Tests | 5962 | 38.92 | 9815 | continuous |
| Vitamin D | Blood Tests | 5721 | 37.35 | 8339 | continuous |
| Ldl Cholesterol | Blood Tests | 5714 | 37.3 | 10571 | continuous |
| Urine Ph | Blood Tests | 5631 | 36.76 | 8976 | continuous |
| Folic Acid | Blood Tests | 5032 | 32.85 | 7011 | continuous |
| Phosphorus | Blood Tests | 4820 | 31.46 | 8242 | continuous |
| Egfr | Blood Tests | 4433 | 28.94 | 8553 | continuous |
| Ldh | Blood Tests | 4388 | 28.64 | 7510 | continuous |
| Crp Wide Range | Blood Tests | 3927 | 25.63 | 5226 | continuous |
| Luc % | Blood Tests | 3546 | 23.15 | 7285 | continuous |
| Luc Abs | Blood Tests | 3541 | 23.12 | 7261 | continuous |
| Rbc Hypo % | Blood Tests | 3434 | 22.42 | 7095 | continuous |
| Rbc Micro% | Blood Tests | 3407 | 22.24 | 7063 | continuous |
| Rbc Macro% | Blood Tests | 3275 | 21.38 | 6695 | continuous |
| Transferrin | Blood Tests | 3264 | 21.31 | 4680 | continuous |
| Rbc Hyper % | Blood Tests | 3145 | 20.53 | 6320 | continuous |
| Rbc Micro Hypo Ratio | Blood Tests | 3020 | 19.71 | 6436 | continuous |
| Bilirubin Direct | Blood Tests | 2619 | 17.1 | 4725 | continuous |
| T4 Free | Blood Tests | 2126 | 13.88 | 3384 | continuous |
| Pt | Blood Tests | 1816 | 11.85 | 2164 | continuous |
| Urine Sg | Blood Tests | 1771 | 11.56 | 2615 | continuous |
| Magnesium | Blood Tests | 1735 | 11.33 | 2367 | continuous |
| Immature Granulocytes % | Blood Tests | 883 | 5.76 | 1521 | continuous |
| Immature Granulocytes | Blood Tests | 878 | 5.73 | 1513 | continuous |
| Pt % | Blood Tests | 761 | 4.97 | 921 | continuous |
| Estradiol E2 | Blood Tests | 742 | 4.84 | 896 | continuous |
| Creatine Kinase | Blood Tests | 100 | 0.65 | 173 | continuous |
| Ebv Igg | Blood Tests | 100 | 0.65 | 101 | continuous |

Table S6- Complete Modality List: Sleep + ECG

| Modality | Group | N Patients | % Patients | Total Tokens | Type |
|---|---|---|---|---|---|
| ECG | R R Ms | 13085 | 85.42 | 23929 | continuous |
| ECG | Qt Ms | 13079 | 85.38 | 23896 | continuous |
| ECG | R Ms Ii | 13077 | 85.36 | 23912 | continuous |
| ECG | Qtc Ms | 13074 | 85.34 | 23897 | continuous |
| ECG | R Mv V4 | 13073 | 85.34 | 23889 | continuous |
| ECG | R Ms I | 13060 | 85.25 | 23860 | continuous |
| ECG | R Mv V3 | 13055 | 85.22 | 23844 | continuous |
| ECG | R Ms V6 | 13047 | 85.17 | 23798 | continuous |
| ECG | R Ms V5 | 13035 | 85.09 | 23728 | continuous |
| ECG | Qrs Ms | 13028 | 85.04 | 23757 | continuous |
| ECG | R Ms V3 | 13028 | 85.04 | 23758 | continuous |
| ECG | R Ms V4 | 13021 | 85 | 23659 | continuous |
| ECG | Pr Ms | 12999 | 84.86 | 23603 | continuous |
| ECG | P Ms | 12993 | 84.82 | 23541 | continuous |
| ECG | R Mv V2 | 12992 | 84.81 | 23617 | continuous |
| ECG | R Ms Avf | 12989 | 84.79 | 23631 | continuous |
| ECG | T Mv I | 12974 | 84.69 | 23586 | continuous |
| ECG | R Ms V2 | 12962 | 84.61 | 23540 | continuous |
| ECG | T Mv V3 | 12915 | 84.31 | 23366 | continuous |
| ECG | T Mv V4 | 12871 | 84.02 | 23225 | continuous |
| ECG | T Mv V2 | 12789 | 83.48 | 23027 | continuous |
| ECG | R Ms Iii | 12777 | 83.41 | 22963 | continuous |
| ECG | R Mv Iii | 12747 | 83.21 | 22888 | continuous |
| ECG | P Axis | 12693 | 82.86 | 22753 | continuous |
| ECG | R Ms V1 | 12665 | 82.68 | 22679 | continuous |
| ECG | R Mv V1 | 12636 | 82.49 | 22604 | continuous |
| ECG | R Ms Avl | 12604 | 82.28 | 22740 | continuous |
| ECG | P Mv Ii | 12595 | 82.22 | 22438 | continuous |
| ECG | St Mv V3 | 12530 | 81.79 | 22118 | continuous |
| ECG | St Mv V1 | 12518 | 81.72 | 21947 | continuous |
| ECG | T Axis | 12485 | 81.5 | 22127 | continuous |
| ECG | P Mv V3 | 12342 | 80.57 | 21480 | continuous |
| ECG | P Mv V4 | 12299 | 80.29 | 21199 | continuous |
| ECG | P Mv I | 12011 | 78.41 | 20657 | continuous |
| ECG | P Mv V2 | 11937 | 77.92 | 20304 | continuous |
| ECG | P Mv V6 | 11888 | 77.6 | 20206 | continuous |
| ECG | P Mv Avf | 11697 | 76.36 | 20129 | continuous |
| Sleep | Total Deep Sleep Time | 11517 | 75.18 | 49601 | continuous |
| Sleep | Heart Rate Min During Wake | 11517 | 75.18 | 49601 | continuous |
| Sleep | Total Light Sleep Time | 11517 | 75.18 | 49601 | continuous |
| Sleep | Heart Rate Max During Wake | 11516 | 75.17 | 49490 | continuous |
| Sleep | Sleep Latency | 11509 | 75.13 | 49452 | continuous |
| Sleep | Total Arousal Sleep Time | 11499 | 75.06 | 49508 | continuous |
| Sleep | Sleep Efficiency | 11470 | 74.87 | 49118 | continuous |
| Sleep | Quality Score Sbp | 11456 | 74.78 | 47810 | continuous |
| Sleep | Number Of All Respiratory Events | 11448 | 74.73 | 48525 | continuous |
| Sleep | Quality Score Pat | 11444 | 74.7 | 47996 | continuous |
| Sleep | Desaturations Of 2 3 | 11437 | 74.66 | 48131 | continuous |
| Sleep | Heart Rate Min During Sleep | 11430 | 74.61 | 48953 | continuous |
| Sleep | Total Wake Time After Sleep Onset | 11429 | 74.61 | 46934 | continuous |
| Sleep | Neurokit Hrv Time Rmssd During Wake | 11428 | 74.6 | 46860 | continuous |
| Sleep | Heart Rate Max During Sleep | 11428 | 74.6 | 48768 | continuous |
| Sleep | Neurokit Hrv Time Rmssd During N... | 11397 | 74.4 | 46246 | continuous |
| Sleep | Heart Rate Max During Rem | 11382 | 74.3 | 47868 | continuous |
| Sleep | Heart Rate Min During Rem | 11382 | 74.3 | 47951 | continuous |
| Sleep | Heart Rate Min During Nrem | 11382 | 74.3 | 48024 | continuous |
| Sleep | Percent Of Deep Sleep Time | 11382 | 74.3 | 48024 | continuous |
| Sleep | Total Rem Sleep Time | 11382 | 74.3 | 47957 | continuous |
| Sleep | Percent Of Rem Sleep Time | 11382 | 74.3 | 47957 | continuous |
| Sleep | Percent Of Light Sleep Time | 11382 | 74.3 | 48026 | continuous |
| Sleep | Heart Rate Max During Nrem | 11381 | 74.29 | 47888 | continuous |
| Sleep | Quality Score Spo2 | 11380 | 74.29 | 47027 | continuous |
| Sleep | Number Of Transitions Nrem To Rem | 11379 | 74.28 | 47766 | continuous |
| Sleep | Time To Transition Nrem To Rem | 11378 | 74.27 | 47644 | continuous |
| Sleep | Rem Latency | 11372 | 74.23 | 47332 | continuous |
| Sleep | Time To Transition Rem To Nrem | 11358 | 74.14 | 46979 | continuous |
| Sleep | Number Of Transitions Rem To Nrem | 11358 | 74.14 | 47046 | continuous |
| Sleep | Time To Transition Nrem To Wake | 11356 | 74.13 | 46303 | continuous |
| Sleep | Saturation Min Value | 11353 | 74.11 | 47514 | continuous |
| Sleep | Total Valid Apnea Central Sleep ... | 11334 | 73.99 | 46889 | continuous |
| Sleep | Snore Above 40Db | 11330 | 73.96 | 46705 | continuous |
| Sleep | Time To Transition Wake To Nrem | 11317 | 73.88 | 45348 | continuous |
| Sleep | Ahi | 11316 | 73.87 | 46999 | continuous |
| Sleep | Number Of Transitions Wake To Nrem | 11316 | 73.87 | 45338 | continuous |
| Sleep | Desaturations Mean Nadir | 11305 | 73.8 | 46250 | continuous |
| Sleep | Total Supine Sleep Time | 11301 | 73.77 | 46087 | continuous |
| ECG | Qrs Axis | 11299 | 73.76 | 19967 | continuous |
| Sleep | Total Non Supine Sleep Time | 11294 | 73.73 | 46185 | continuous |
| Sleep | Number Of 4 Percent Desaturation... | 11293 | 73.72 | 45194 | continuous |
| Sleep | Ahi During Rem | 11225 | 73.28 | 44456 | continuous |
| Sleep | Odi | 11186 | 73.02 | 44916 | continuous |
| Sleep | Snore Above 60Db | 11180 | 72.98 | 44420 | continuous |
| Sleep | Snore Above 45Db | 11109 | 72.52 | 43977 | continuous |
| Sleep | Rdi During Non Supine | 11060 | 72.2 | 43145 | continuous |
| Sleep | Ahi During Supine | 10980 | 71.68 | 41780 | continuous |
| Sleep | Total Right Sleep Time | 10972 | 71.62 | 40022 | continuous |
| Sleep | Ahi During Non Supine | 10966 | 71.58 | 41328 | continuous |
| Sleep | Total Left Sleep Time | 10959 | 71.54 | 39626 | continuous |
| Sleep | Odi During Rem | 10931 | 71.36 | 39389 | continuous |
| ECG | St Mv V5 | 10931 | 71.36 | 17854 | continuous |
| Sleep | Number Of Transitions Rem To Wake | 10926 | 71.32 | 35232 | continuous |
| Sleep | Time To Transition Rem To Wake | 10925 | 71.32 | 35207 | continuous |
| Sleep | Premature Beats Per Minute | 10905 | 71.19 | 41238 | continuous |
| ECG | T Mv Avl | 10831 | 70.7 | 18657 | continuous |
| ECG | R Ms Avr | 10765 | 70.27 | 18422 | continuous |
| ECG | J Mv V2 | 10716 | 69.95 | 17370 | continuous |
| Sleep | Odi During Supine | 10710 | 69.91 | 37963 | continuous |
| ECG | P Mv V1 | 10608 | 69.25 | 16649 | continuous |
| Sleep | Odi During Non Supine | 10521 | 68.68 | 35322 | continuous |
| Sleep | Rdi During Right | 10380 | 67.76 | 33827 | continuous |
| Sleep | Rdi During Left | 10312 | 67.32 | 33278 | continuous |
| Sleep | Ahi During Right | 9989 | 65.21 | 30208 | continuous |
| Sleep | Ahi During Left | 9969 | 65.08 | 30016 | continuous |
| ECG | P Mv Iii | 9966 | 65.06 | 16266 | continuous |
| Sleep | Total Prone Sleep Time | 9624 | 62.82 | 28238 | continuous |
| Sleep | Saturation Below 90 | 9539 | 62.27 | 25645 | continuous |
| Sleep | Number Of Central Apnea Hypopnea... | 9442 | 61.64 | 25156 | continuous |
| ECG | J Mv V3 | 9391 | 61.3 | 14622 | continuous |
| Sleep | Odi During Right | 8902 | 58.11 | 22349 | continuous |
| Sleep | Odi During Left | 8865 | 57.87 | 22376 | continuous |
| Sleep | Number Of Transitions Wake To Rem | 8772 | 57.26 | 18961 | continuous |
| ECG | Q Ms Avr | 8640 | 56.4 | 14223 | continuous |
| Sleep | Time To Transition Wake To Rem | 8622 | 56.28 | 18424 | continuous |
| Sleep | Desaturations Of 10 20 | 8497 | 55.47 | 18643 | continuous |
| ECG | J Mv V1 | 8166 | 53.31 | 11998 | continuous |
| Sleep | Rdi During Prone | 8056 | 52.59 | 21737 | continuous |
| Sleep | Ahi Central During Rem | 7628 | 49.79 | 15989 | continuous |
| Sleep | Ahi During Prone | 7525 | 49.12 | 19187 | continuous |
| ECG | J Mv Avr | 7199 | 46.99 | 11081 | continuous |
| ECG | J Mv Ii | 7059 | 46.08 | 10951 | continuous |
| ECG | J Mv I | 6648 | 43.4 | 10192 | continuous |
| ECG | Q Ms Avl | 6581 | 42.96 | 10647 | continuous |
| Sleep | Odi During Prone | 6304 | 41.15 | 14039 | continuous |
| ECG | J Mv V6 | 6056 | 39.53 | 8398 | continuous |
| ECG | T Mv V1 | 5668 | 37 | 8416 | continuous |
| Sleep | Saturation Below 85 | 5235 | 34.17 | 7822 | continuous |
| ECG | Q Ms I | 5192 | 33.89 | 8212 | continuous |
| ECG | Q Ms Iii | 4662 | 30.43 | 7125 | continuous |
| ECG | Q Ms Ii | 3754 | 24.51 | 5591 | continuous |
| ECG | St Mv Avr | 3608 | 23.55 | 4841 | continuous |
| ECG | Q Ms Avf | 3371 | 22.01 | 4955 | continuous |
| ECG | Q Ms V4 | 1319 | 8.61 | 1702 | continuous |
| ECG | Q Ms V1 | 1025 | 6.69 | 1329 | continuous |
| ECG | T Mv Avr | 78 | 0.51 | 93 | continuous |

Table S7 - Complete Modality List: DEXA + Body Measures

| Modality | Group | N Patients | % Patients | Total Tokens | Type |
|---|---|---|---|---|---|
| Height | Body Measures | 13293 | 86.77 | 25067 | continuous |
| Body Weight | Body Measures | 13292 | 86.77 | 25062 | continuous |
| Waist | Body Measures | 13291 | 86.76 | 25098 | continuous |
| Sitting Blood Pressure Pulse Rate | Body Measures | 13290 | 86.76 | 25082 | continuous |
| Bmi | Body Measures | 13289 | 86.75 | 25054 | continuous |
| Sitting Blood Pressure Systolic | Body Measures | 13287 | 86.74 | 25070 | continuous |
| Sitting Blood Pressure Diastolic | Body Measures | 13286 | 86.73 | 25078 | continuous |
| Hips | Body Measures | 13277 | 86.67 | 25059 | continuous |
| Hand Grip Right | Body Measures | 13273 | 86.64 | 25022 | continuous |
| Hand Grip Left | Body Measures | 13270 | 86.62 | 25013 | continuous |
| Body Total Bmc | DEXA | 11902 | 77.69 | 20364 | continuous |
| Body Trunk Area | DEXA | 11902 | 77.69 | 20363 | continuous |
| Body Total Area | DEXA | 11902 | 77.69 | 20363 | continuous |
| Body Pelvis Area | DEXA | 11902 | 77.69 | 20362 | continuous |
| Body Arm Right Bmd | DEXA | 11901 | 77.69 | 20360 | continuous |
| Body Head Bmd | DEXA | 11901 | 77.69 | 20359 | continuous |
| Body Arms Area | DEXA | 11900 | 77.68 | 20358 | continuous |
| Body Head Bmc | DEXA | 11900 | 77.68 | 20357 | continuous |
| Body Ribs Bmc | DEXA | 11900 | 77.68 | 20355 | continuous |
| Body Spine Area | DEXA | 11900 | 77.68 | 20354 | continuous |
| Body Head Area | DEXA | 11899 | 77.67 | 20344 | continuous |
| Body Comp Arm Left Tissue Percentage | DEXA | 11876 | 77.52 | 20275 | continuous |
| Body Comp Android Bone Mass | DEXA | 11875 | 77.52 | 20275 | continuous |
| Body Comp Trunk Right Bone Mass | DEXA | 11875 | 77.52 | 20273 | continuous |
| Body Comp Leg Right Bone Mass | DEXA | 11874 | 77.51 | 20274 | continuous |
| Body Comp Total Tissue Mass | DEXA | 11874 | 77.51 | 20273 | continuous |
| Body Comp Trunk Left Bone Mass | DEXA | 11874 | 77.51 | 20271 | continuous |
| Body Comp Arms Total Mass | DEXA | 11874 | 77.51 | 20271 | continuous |
| Body Comp Arms Bone Mass | DEXA | 11874 | 77.51 | 20266 | continuous |
| Body Comp Gynoid Bone Mass | DEXA | 11873 | 77.51 | 20267 | continuous |
| Body Comp Trunk Diff Bone Mass | DEXA | 11872 | 77.5 | 20201 | continuous |
| Body Comp Arms Fat Mass | DEXA | 11855 | 77.39 | 20224 | continuous |
| Body Comp Legs Fat Mass | DEXA | 11854 | 77.38 | 20229 | continuous |
| Body Comp Arm Left Fat Mass | DEXA | 11851 | 77.36 | 20218 | continuous |
| Body Comp Arm Right Fat Mass | DEXA | 11849 | 77.35 | 20221 | continuous |
| Body Comp Trunk Diff Fat Free Mass | DEXA | 11820 | 77.16 | 20051 | continuous |
| Body Comp Trunk Diff Lean Mass | DEXA | 11815 | 77.13 | 20070 | continuous |
| Body Comp Total Diff Fat Free Mass | DEXA | 11815 | 77.13 | 20039 | continuous |
| Body Comp Total Diff Tissue Mass | DEXA | 11815 | 77.13 | 20065 | continuous |
| Body Comp Trunk Diff Tissue Mass | DEXA | 11811 | 77.1 | 20063 | continuous |
| Total Scan Sat Volume | DEXA | 11808 | 77.08 | 20089 | continuous |
| Body Comp Total Diff Lean Mass | DEXA | 11808 | 77.08 | 20017 | continuous |
| Body Comp Total Diff Total Mass | DEXA | 11807 | 77.07 | 19900 | continuous |
| Body Comp Trunk Diff Region Percentage | DEXA | 11806 | 77.07 | 19611 | continuous |
| Body Comp Trunk Diff Total Mass | DEXA | 11799 | 77.02 | 19869 | continuous |
| Lying Blood Pressure Pulse Rate | Body Measures | 11794 | 76.99 | 15924 | continuous |
| Lying Blood Pressure Diastolic | Body Measures | 11791 | 76.97 | 15922 | continuous |
| Lying Blood Pressure Systolic | Body Measures | 11789 | 76.96 | 15916 | continuous |
| Spine L4 Average Height | DEXA | 11777 | 76.88 | 19880 | continuous |
| Spine L4 Am Percent | DEXA | 11775 | 76.87 | 19864 | continuous |
| Body Comp Legs Diff Bone Mass | DEXA | 11763 | 76.79 | 19796 | continuous |
| Standing One Min Blood Pressure Systolic | Body Measures | 11761 | 76.77 | 15879 | continuous |
| Body Comp Total Diff Region Percentage | DEXA | 11759 | 76.76 | 19342 | continuous |
| Femur Right Troch Area | DEXA | 11759 | 76.76 | 19775 | continuous |
| Standing One Min Blood Pressure Diastolic | Body Measures | 11757 | 76.75 | 15873 | continuous |
| Femur Right Shaft Area | DEXA | 11757 | 76.75 | 19774 | continuous |
| Total Scan Sat Mass | DEXA | 11755 | 76.73 | 19924 | continuous |
| Standing Three Min Blood Pressure | Body Measures | 11755 | 76.73 | 15856 | continuous |
| Spine L3 L4 Area | DEXA | 11752 | 76.72 | 19810 | continuous |
| Spine L3 L4 Bmc | DEXA | 11751 | 76.71 | 19800 | continuous |
| Standing Three Min Blood Pressure Systolic | Body Measures | 11749 | 76.7 | 15843 | continuous |
| Standing Three Min Blood Pressure Diastolic | Body Measures | 11749 | 76.7 | 15848 | continuous |
| Total Scan Vat Volume | DEXA | 11742 | 76.65 | 19954 | continuous |
| Femur Left Wards Area | DEXA | 11736 | 76.61 | 19728 | continuous |
| Total Scan Vat Area | DEXA | 11735 | 76.6 | 19923 | continuous |
| Body Comp Total Diff Fat Mass | DEXA | 11733 | 76.59 | 19813 | continuous |
| Body Comp Legs Diff Region Percentage | DEXA | 11707 | 76.42 | 19330 | continuous |
| Body Comp Legs Diff Tissue Percentage | DEXA | 11707 | 76.42 | 19355 | continuous |
| Body Comp Trunk Diff Fat Mass | DEXA | 11699 | 76.37 | 19663 | continuous |
| Spine L1 L4 Average Height | DEXA | 11697 | 76.36 | 19645 | continuous |
| Body Comp Arm Right Bone Mass | DEXA | 11668 | 76.17 | 19735 | continuous |
| Femur Upper Neck Diff Bmd | DEXA | 11668 | 76.17 | 19515 | continuous |
| Femur Shaft Diff Bmd | DEXA | 11668 | 76.17 | 19559 | continuous |
| Femur Troch Diff Bmd | DEXA | 11665 | 76.15 | 19528 | continuous |
| Femur Shaft Diff Area | DEXA | 11661 | 76.12 | 19509 | continuous |
| Femur Neck Diff Bmd | DEXA | 11658 | 76.1 | 19503 | continuous |
| Femur Total Diff Bmd | DEXA | 11658 | 76.1 | 19519 | continuous |
| Femur Shaft Diff Bmc | DEXA | 11658 | 76.1 | 19537 | continuous |
| Femur Lower Neck Diff Bmd | DEXA | 11656 | 76.09 | 19499 | continuous |
| Body Comp Legs Diff Tissue Mass | DEXA | 11651 | 76.06 | 19538 | continuous |
| Femur Wards Diff Bmd | DEXA | 11650 | 76.05 | 19505 | continuous |
| Femur Troch Diff Bmc | DEXA | 11643 | 76 | 19458 | continuous |
| Femur Troch Diff Area | DEXA | 11641 | 75.99 | 19451 | continuous |
| Femur Total Diff Bmc | DEXA | 11638 | 75.97 | 19491 | continuous |
| Body Comp Legs Diff Total Mass | DEXA | 11628 | 75.91 | 19132 | continuous |
| Femur Total Diff Area | DEXA | 11617 | 75.83 | 19408 | continuous |
| Body Comp Legs Diff Lean Mass | DEXA | 11615 | 75.82 | 19454 | continuous |
| Body Comp Legs Diff Fat Free Mass | DEXA | 11613 | 75.81 | 19484 | continuous |
| Femur Upper Neck Diff Bmc | DEXA | 11592 | 75.67 | 19253 | continuous |
| Femur Neck Diff Bmc | DEXA | 11588 | 75.64 | 19315 | continuous |
| Femur Lower Neck Diff Bmc | DEXA | 11565 | 75.49 | 19190 | continuous |
| Femur Wards Diff Bmc | DEXA | 11558 | 75.45 | 19249 | continuous |
| Body Comp Arm Left Bone Mass | DEXA | 11557 | 75.44 | 19473 | continuous |
| Femur Wards Diff Area | DEXA | 11538 | 75.32 | 19198 | continuous |
| Body Comp Legs Diff Fat Mass | DEXA | 11534 | 75.29 | 19137 | continuous |
| Body Comp Total Diff Bone Mass | DEXA | 11512 | 75.15 | 18949 | continuous |
| Femur Lower Neck Diff Area | DEXA | 11494 | 75.03 | 18967 | continuous |
| Femur Upper Neck Diff Area | DEXA | 11491 | 75.01 | 18952 | continuous |
| Spine L1 L4 Area | DEXA | 11430 | 74.61 | 17231 | continuous |
| Spine L1 L4 Bmc | DEXA | 11426 | 74.59 | 17226 | continuous |
| Total Scan Vat Mass | DEXA | 11356 | 74.13 | 19185 | continuous |
| Body Comp Arms Diff Bone Mass | DEXA | 10635 | 69.42 | 16767 | continuous |
| Body Comp Arms Diff Tissue Percentage | DEXA | 10581 | 69.07 | 16643 | continuous |
| Body Comp Arms Diff Region Percentage | DEXA | 10572 | 69.01 | 16614 | continuous |
| Body Comp Arms Diff Tissue Mass | DEXA | 10375 | 67.73 | 16328 | continuous |
| Body Comp Arms Diff Lean Mass | DEXA | 10369 | 67.69 | 16285 | continuous |
| Body Comp Arms Diff Fat Free Mass | DEXA | 10341 | 67.5 | 16288 | continuous |
| Body Comp Arms Diff Total Mass | DEXA | 10270 | 67.04 | 15795 | continuous |
| Body Comp Arms Diff Fat Mass | DEXA | 10115 | 66.03 | 15558 | continuous |
| Femur Neck Mean Z Score | DEXA | 6535 | 42.66 | 9665 | continuous |
| Femur Left Total T Score | DEXA | 6248 | 40.79 | 9077 | continuous |
| Spine L2 T Score | DEXA | 6189 | 40.4 | 10138 | continuous |
| Femur Left Upper Neck T Score | DEXA | 6184 | 40.37 | 8618 | continuous |
| Trunk Fat | Body Measures | 3746 | 24.45 | 3908 | continuous |
| Body Fat | Body Measures | 3745 | 24.45 | 3907 | continuous |
| Bmr | Body Measures | 3744 | 24.44 | 3905 | continuous |
| Fingers Grip Right | Body Measures | 1312 | 8.56 | 1312 | continuous |
| Fingers Grip Left | Body Measures | 1309 | 8.54 | 1309 | continuous |
| Number Of Days From Last Period | Body Measures | 630 | 4.11 | 656 | continuous |
| Number Of Days In Cycle | Body Measures | 546 | 3.56 | 570 | continuous |

Table S8- Complete Modality List: Other Categories

| Modality | Group | N Patients | % Patients | Total Tokens | Type |
|---|---|---|---|---|---|
| Eat Chicken Poultry | Questionnaires | 15142 | 98.84 | 44387 | categorical |
| Worry Long After Embarrassment | Questionnaires | 15142 | 98.84 | 44387 | categorical |
| Years Using Mobile Phone | Questionnaires | 15142 | 98.84 | 44387 | categorical |
| Army Position | Questionnaires | 15142 | 98.84 | 44387 | categorical |
| Risks Taker | Questionnaires | 15142 | 98.84 | 44387 | categorical |
| Living Place Today | Questionnaires | 15142 | 98.84 | 44387 | categorical |
| Work Satisfaction | Questionnaires | 15142 | 98.84 | 44387 | categorical |
| Total Income | Questionnaires | 15142 | 98.84 | 44387 | categorical |
| Easy Getting Up | Questionnaires | 15142 | 98.84 | 44387 | categorical |
| Eatother Fish | Questionnaires | 15142 | 98.84 | 44387 | categorical |
| Eat Kosher | Questionnaires | 15142 | 98.84 | 44387 | categorical |
| Army Service | Questionnaires | 15142 | 98.84 | 44387 | categorical |
| Vegetarian Yes No | Questionnaires | 15142 | 98.84 | 44387 | categorical |
| Pet Past | Questionnaires | 15142 | 98.84 | 44387 | categorical |
| Why Reduce Drinking | Questionnaires | 15142 | 98.84 | 44387 | categorical |
| Suffer Nerves | Questionnaires | 15142 | 98.84 | 44387 | categorical |
| Manual Physical Work | Questionnaires | 15142 | 98.84 | 44387 | categorical |
| Never Eat | Questionnaires | 15142 | 98.84 | 44387 | categorical |
| Friendships Satisfaction | Questionnaires | 15142 | 98.84 | 44387 | categorical |
| Is Getting Period Ques | Questionnaires | 15142 | 98.84 | 44387 | categorical |
| Worrier | Questionnaires | 15142 | 98.84 | 44387 | categorical |
| Mood Swings | Questionnaires | 15142 | 98.84 | 44387 | categorical |
| Visit Friends Family Times | Questionnaires | 15142 | 98.84 | 44387 | categorical |
| High Exercise Duration | Questionnaires | 15142 | 98.84 | 44387 | categorical |
| Overall Health | Questionnaires | 15142 | 98.84 | 44387 | categorical |
| Irritable Person | Questionnaires | 15142 | 98.84 | 44387 | categorical |
| Leg Pain On Walking | Questionnaires | 15142 | 98.84 | 44387 | categorical |
| Smoking Compared 10Years | Questionnaires | 15142 | 98.84 | 44387 | categorical |
| Drink Compared 10Years | Questionnaires | 15142 | 98.84 | 44387 | categorical |
| Confide Frequency | Questionnaires | 15142 | 98.84 | 44387 | categorical |
| Nervous Person | Questionnaires | 15142 | 98.84 | 44387 | categorical |
| Miserable No Reason | Questionnaires | 15142 | 98.84 | 44387 | categorical |
| Contraception Type | Questionnaires | 15142 | 98.84 | 44387 | categorical |
| Women Symptoms | Questionnaires | 15142 | 98.84 | 44387 | categorical |
| Shortness Of Breath Same Age Group | Questionnaires | 15142 | 98.84 | 44387 | categorical |
| Tense Or Highly Strung | Questionnaires | 15142 | 98.84 | 44387 | categorical |
| Eat Beef | Questionnaires | 15142 | 98.84 | 44387 | categorical |
| Education | Questionnaires | 15142 | 98.84 | 44387 | categorical |
| Alcohol Drink Past | Questionnaires | 15142 | 98.84 | 44387 | categorical |
| Why Stop Drinking | Questionnaires | 15142 | 98.84 | 44387 | categorical |
| Diet Vary Week To Week | Questionnaires | 15142 | 98.84 | 44387 | categorical |
| Pet Present | Questionnaires | 15142 | 98.84 | 44387 | categorical |
| Add Salt To Food | Questionnaires | 15142 | 98.84 | 44387 | categorical |
| Symptoms Of Menstruation | Questionnaires | 15142 | 98.84 | 44387 | categorical |
| Grandmother Country Of Birth Mother | Questionnaires | 15142 | 98.84 | 44387 | categorical |
| Troubled By Guilt | Questionnaires | 15142 | 98.84 | 44387 | categorical |
| Grandfather Country Of Birth Mother | Questionnaires | 15142 | 98.84 | 44387 | categorical |
| Eat Cheese | Questionnaires | 15142 | 98.84 | 44387 | categorical |
| Transportaion To Work | Questionnaires | 15142 | 98.84 | 44387 | categorical |
| Cigaretts Past Per Day1 | Questionnaires | 15142 | 98.84 | 44387 | categorical |
| Smoke Tobacco Now | Questionnaires | 15142 | 98.84 | 44387 | categorical |
| Father Country Of Birth | Questionnaires | 15142 | 98.84 | 44387 | categorical |
| Smoke Houshold | Questionnaires | 15142 | 98.84 | 44387 | categorical |
| Diet Major Changes 5Years | Questionnaires | 15142 | 98.84 | 44387 | categorical |
| Intermenstrual Bleeding | Questionnaires | 15142 | 98.84 | 44387 | categorical |
| Nerves Anxiety Tension Depression | Questionnaires | 15142 | 98.84 | 44387 | categorical |
| Eat Processed Meat | Questionnaires | 15142 | 98.84 | 44387 | categorical |
| Country Of Birth | Questionnaires | 15142 | 98.84 | 44387 | categorical |
| Regular Periods Past | Questionnaires | 15142 | 98.84 | 44387 | categorical |
| High Irritable Longest Period Duration | Questionnaires | 15142 | 98.84 | 44387 | categorical |
| Health Satisfaction | Questionnaires | 15142 | 98.84 | 44387 | categorical |
| Financial Situation Satisfaction | Questionnaires | 15142 | 98.84 | 44387 | categorical |
| Period Pain | Questionnaires | 15142 | 98.84 | 44387 | categorical |
| Number Of Cars | Questionnaires | 15142 | 98.84 | 44387 | categorical |
| Chest Pain | Questionnaires | 15142 | 98.84 | 44387 | categorical |
| Employment | Questionnaires | 15142 | 98.84 | 44387 | categorical |
| Regular Periods Present | Questionnaires | 15142 | 98.84 | 44387 | categorical |
| Accommodation Type | Questionnaires | 15142 | 98.84 | 44387 | categorical |
| Feeling Lonley Often | Questionnaires | 15142 | 98.84 | 44387 | categorical |
| Grandfather Country Of Birth Father | Questionnaires | 15142 | 98.84 | 44387 | categorical |
| Fed-Up Often | Questionnaires | 15142 | 98.84 | 44387 | categorical |
| Grandmother Country Of Birth Father | Questionnaires | 15142 | 98.84 | 44387 | categorical |
| Family Relationship Satisfaction | Questionnaires | 15142 | 98.84 | 44387 | categorical |
| Chest Pain Walking Uphill Hurry | Questionnaires | 15142 | 98.84 | 44387 | categorical |
| Nap During Day | Questionnaires | 15142 | 98.84 | 44387 | categorical |
| High Exercise Times A Month | Questionnaires | 15142 | 98.84 | 44387 | categorical |
| Smoking Stop More Than 6 Months | Questionnaires | 15142 | 98.84 | 44387 | categorical |
| Disability Illness | Questionnaires | 15142 | 98.84 | 44387 | categorical |
| Sperm Count Test Results | Questionnaires | 15142 | 98.84 | 44387 | categorical |

Per-dataset significance summary:
- HPP: 330/338 significant at p < 0.05
- UKBB: 71/74 significant at p < 0.05
- NHANES: 31/37 significant at p < 0.05
- PNP3: 27/29 significant at p < 0.05
- Framingham: 8/8 significant at p < 0.05

Table S9: Individual-Level Prediction Performance in PNP3

| Biomarker | Pearson r | p-value | 95% CI lower | 95% CI upper | n |
|---|---|---|---|---|---|
| Triglycerides | 0.8896 | 4.68E-05 | 0.6641 | 0.9667 | 13 |
| Diastolic BP | 0.7754 | 3.42E-40 | 0.7124 | 0.8261 | 194 |
| Total Cholesterol | 0.7746 | 6.98E-04 | 0.4348 | 0.9213 | 15 |
| Systolic BP | 0.6925 | 4.83E-29 | 0.6113 | 0.7593 | 194 |
| Waist Circumference | 0.631 | 1.71E-22 | 0.5369 | 0.7096 | 190 |
| BMI | 0.6256 | 4.0E-23 | 0.5331 | 0.7033 | 200 |
| LDL Cholesterol | 0.6154 | 0.0251 | 0.0975 | 0.8711 | 13 |
| Fasting Glucose | 0.6121 | 5.97E-22 | 0.5173 | 0.6921 | 200 |
| HbA1c | 0.4919 | 1.39E-13 | 0.379 | 0.5904 | 200 |
| HDL Cholesterol | 0.4356 | 0.1046 | -0.0987 | 0.7749 | 15 |

Pearson correlation between HealthFormer-predicted and observed six-month biomarker changes for 10 biomarkers in PNP3 (n = 200 trial participants held out from model training). p-values are two-sided; 95% CIs via Fisher Z transformation.
HDL cholesterol did not reach significance (p = 0.105, n = 15).
Lipid endpoints have small sample sizes (n = 13-15) and wide CIs.

Table S10 Correlation between predicted and observed values for nine pairs of cross-modal biomarkers

| Pair | Pearson $r$ | n_bins | p |
|---|---|---|---|
| Pelvis Area to Body Weight | 0.97 | 17 | $2.4 \times 10^{-8}$ |
| BMI to Brachial Pressure | 0.92 | 1.70E+01 | $1.4 \times 10^{-7}$ |
| RBC to Visceral Fat | 0.95 | 1.80E+01 | $8.6 \times 10^{-10}$ |
| IMT to Brachial Pressure | 0.83 | 1.70E+01 | $3.2 \times 10^{-5}$ |
| Ankle Pressure to AHI | 0.87 | 1.70E+01 | $7.5 \times 10^{-6}$ |
| Uric Acid to Tissue Mass | 0.94 | 1.70E+01 | $2.4 \times 10^{-8}$ |
| RDI to Waist | 0.91 | 1.80E+01 | $1.6 \times 10^{-7}$ |
| Visceral Fat to ALT | 0.83 | 17 | $3.7 \times 10^{-5}$ |
| Lymphocytes to Subcut Fat | 0.93 | 1.80E+01 | $1.5 \times 10^{-8}$ |

**Summary**: 330/338 modalities significant at p < 0.05.

8 non-significant modalities (p >= 0.05):

- bt__crp_wide_range_BloodTestsLoader: r=0.1756, n=45, p=0.2485
- bt__potassium_BloodTestsLoader: r=0.0404, n=296, p=0.4888
- bt__transferrin_BloodTestsLoader: r=0.1078, n=43, p=0.4913
- number_of_transitions_wake_to_rem_: r=0.0296, n=344, p=0.5847
- premature_beats_per_minute_ r=0.0000, n=663, p=1.0000
- saturation_below_85_ r=0.0344, n=105, p=0.7277

- st_mv_aVR_ECGTextLoader: r=0.1015, n=163, p=0.1974
- time_to_transition_wake_to_rem_ r=0.0688, n=335, p=0.2089

Table S11: Leave-One-In (LOI) Top 50 Cross-Modal Pairs

| Input Modality | Target Modality | r (empirical) | r (model) | p (model) | CI_lower | CI_upper |
|---|---|---|---|---|---|---|
| bt__uric_acid | body_weight | 0.9828 | 0.978 | 1.29E-11 | 0.9384 | 0.9922 |
| bt__hemoglobin | bt__rbc | 0.9874 | 0.9773 | 3.34E-12 | 0.9387 | 0.9917 |
| body_comp_gynoid_bone_mass | body_comp_trunk_right_bone_mass | 0.9906 | 0.9728 | 6.14E-11 | 0.9244 | 0.9904 |
| height | body_weight | 0.9885 | 0.9726 | 6.42E-11 | 0.9239 | 0.9903 |
| body_comp_android_bone_mass | spine_l1_l4_bmc | 0.9915 | 0.9726 | 6.55E-11 | 0.9237 | 0.9903 |
| bt__uric_acid | body_comp_total_tissue_mass | 0.9791 | 0.97 | 1.27E-10 | 0.9168 | 0.9894 |
| total_scan_sat_mass | body_weight | 0.9989 | 0.9668 | 2.68E-10 | 9.08E-01 | 0.9882 |
| bt__lymphocytes_% | bt__neutrophils_abs | -0.994 | 0.9652 | 9.72E-11 | 0.9071 | 0.9872 |
| body_total_area | body_weight | 0.9945 | 0.9638 | 5.11E-10 | 0.9002 | 0.9872 |
| body_comp_trunk_left_bone_mass | body_comp_trunk_right_bone_mass | 0.9935 | 0.9633 | 5.71E-10 | 0.8987 | 0.987 |
| number_of_transitions_rem_to_nrem | total_arousal_sleep_time | 0.9941 | 0.963 | 1.23E-04 | 0.804 | 0.9935 |
| standing_three_min_blood_pressure_s | total_scan_vat_volume | 0.9752 | 0.9626 | 6.51E-10 | 8.97E-01 | 0.9867 |
| height | spine_l3_l4_bmc | 0.992 | 0.9602 | 1.03E-09 | 8.91E-01 | 0.9859 |
| body_comp_trunk_right_bone_mass | body_comp_android_bone_mass | 0.9898 | 0.9598 | 1.11E-09 | 0.8894 | 0.9857 |
| bt__platelets | hand_grip_left | -0.9813 | 0.9587 | 3.74E-10 | 8.90E-01 | 0.9848 |
| bt__ldl_cholesterol | bt__non_hdl_cholesterol | 0.996 | 0.9577 | 1.62E-09 | 0.884 | 0.985 |
| bmi | total_scan_vat_volume | 0.9858 | 0.9574 | 1.70E-09 | 0.8832 | 0.9849 |
| body_trunk_area | standing_three_min_blood_pressure_d | 0.976 | 0.9566 | 1.97E-09 | 8.81E-01 | 0.9845 |
| bt__lymphocytes_abs | total_scan_sat_mass | 0.9746 | 0.9553 | 6.94E-10 | 0.8817 | 0.9835 |
| bt__lymphocytes_abs | bt__wbc | 0.9901 | 0.9549 | 7.51E-10 | 0.8806 | 0.9834 |
| t_mv_aVL_ECGTextLoader | bt__hemoglobin | 0.9797 | 0.9546 | 2.73E-09 | 8.76E-01 | 0.9838 |
| bmi | body_comp_total_tissue_mass | 0.9969 | 0.9534 | 3.28E-09 | 0.8728 | 0.9834 |
| height | body_comp_arms_total_mass | 0.9787 | 0.9533 | 3.34E-09 | 0.8724 | 0.9834 |
| body_head_bmd | body_comp_android_bone_mass | 0.9885 | 0.9528 | 3.64E-09 | 8.71E-01 | 0.9832 |
| femur_left_wards_area | body_weight | 0.9745 | 0.9525 | 3.83E-09 | 8.70E-01 | 0.9831 |
| femur_right_shaft_area | body_weight | 0.9772 | 0.9519 | 4.16E-09 | 0.8687 | 0.9829 |
| body_comp_trunk_right_bone_mass | body_comp_total_tissue_mass | 0.9916 | 0.9515 | 4.42E-09 | 8.68E-01 | 0.9827 |
| l_brachial_pressure_ABILoader | lying_blood_pressure_diastolic | 0.9883 | 0.9515 | 4.43E-09 | 0.8677 | 0.9827 |
| body_comp_gynoid_bone_mass | body_comp_total_tissue_mass | 0.9879 | 0.9499 | 5.64E-09 | 0.8634 | 0.9821 |
| body_comp_arms_total_mass | bt__triglycerides | 0.9753 | 0.9496 | 5.89E-09 | 0.8627 | 0.982 |

Table S12: Survival C-Index (All Endpoints)

| Endpoint | Method | C-index | SD | 95% CI lower | 95% CI upper |
|---|---|---|---|---|---|
| All-Cause Mortality | Age + Sex + BMI | 0.7154 | 0.0016 | 0.7123 | 0.7186 |
| All-Cause Mortality | Foundation Model | 0.7845 | 0.0021 | 0.7804 | 0.7886 |
| All-Cause Mortality | All Model Covariates | 0.7621 | 0.0026 | 0.7569 | 0.7672 |
| All-Cause Mortality | Ensemble | 0.7827 | 0.0024 | 0.7779 | 0.7875 |
| Alzheimers Disease | Age + Sex + BMI | 0.8182 | 0.0037 | 0.8109 | 0.8255 |
| Alzheimers Disease | Foundation Model | 0.8272 | 0.0060 | 0.8154 | 0.8391 |
| Alzheimers Disease | All Model Covariates | 0.8239 | 0.0046 | 0.8149 | 0.8330 |
| Alzheimers Disease | Ensemble | 0.8366 | 0.0045 | 0.8279 | 0.8454 |
| Angina Pectoris | Age + Sex + BMI | 0.6852 | 0.0024 | 0.6805 | 0.6900 |
| Angina Pectoris | Foundation Model | 0.7073 | 0.0034 | 0.7006 | 0.7140 |
| Angina Pectoris | All Model Covariates | 0.7135 | 0.0049 | 0.7038 | 0.7232 |
| Angina Pectoris | Ensemble | 0.7160 | 0.0041 | 0.7079 | 0.7241 |
| Atrial Fibrillation | Age + Sex + BMI | 0.7428 | 0.0032 | 0.7366 | 0.7490 |
| Atrial Fibrillation | Foundation Model | 0.7668 | 0.0019 | 0.7631 | 0.7704 |
| Atrial Fibrillation | All Model Covariates | 0.7647 | 0.0020 | 0.7607 | 0.7687 |
| Atrial Fibrillation | Ensemble | 0.7700 | 0.0021 | 0.7659 | 0.7741 |
| COPD | Age + Sex + BMI | 0.6852 | 0.0030 | 0.6794 | 0.6910 |
| COPD | Foundation Model | 0.8233 | 0.0025 | 0.8185 | 0.8282 |
| COPD | All Model Covariates | 0.8239 | 0.0027 | 0.8185 | 0.8292 |
| COPD | Ensemble | 0.8297 | 0.0025 | 0.8247 | 0.8347 |
| COVID-19 Mortality | Age + Sex + BMI | 0.7771 | 0.0111 | 0.7553 | 0.7989 |
| COVID-19 Mortality | Foundation Model | 0.7741 | 0.0078 | 0.7589 | 0.7894 |
| COVID-19 Mortality | All Model Covariates | 0.8038 | 0.0079 | 0.7883 | 0.8193 |
| COVID-19 Mortality | Ensemble | 0.8076 | 0.0081 | 0.7917 | 0.8235 |
| Cancer | Age + Sex + BMI | 0.6467 | 0.0013 | 0.6442 | 0.6492 |
| Cancer | Foundation Model | 0.6563 | 0.0008 | 0.6548 | 0.6578 |
| Cancer | All Model Covariates | 0.6542 | 0.0011 | 0.6520 | 0.6563 |
| Cancer | Ensemble | 0.6585 | 0.0010 | 0.6566 | 0.6604 |
| Cancer Mortality | Age + Sex + BMI | 0.6903 | 0.0012 | 0.6880 | 0.6927 |
| Cancer Mortality | Foundation Model | 0.7444 | 0.0017 | 0.7410 | 0.7478 |
| Cancer Mortality | All Model Covariates | 0.7246 | 0.0056 | 0.7136 | 0.7356 |
| Cancer Mortality | Ensemble | 0.7452 | 0.0021 | 0.7412 | 0.7493 |
| Cardiovascular Mortality | Age + Sex + BMI | 0.7603 | 0.0030 | 0.7545 | 0.7661 |
| Cardiovascular Mortality | Foundation Model | 0.8265 | 0.0050 | 0.8167 | 0.8363 |
| Cardiovascular Mortality | All Model Covariates | 0.8047 | 0.0056 | 0.7938 | 0.8157 |
| Cardiovascular Mortality | Ensemble | 0.8278 | 0.0045 | 0.8189 | 0.8367 |
| Chronic Ischaemic Heart Disease | Age + Sex + BMI | 0.7167 | 0.0028 | 0.7113 | 0.7221 |
| Chronic Ischaemic Heart Disease | Foundation Model | 0.7461 | 0.0013 | 0.7436 | 0.7486 |
| Chronic Ischaemic Heart Disease | All Model Covariates | 0.7437 | 0.0024 | 0.7391 | 0.7483 |
| Chronic Ischaemic Heart Disease | Ensemble | 0.7495 | 0.0019 | 0.7459 | 0.7532 |
| Chronic Kidney Disease | Age + Sex + BMI | 0.7436 | 0.0022 | 0.7393 | 0.7479 |
| Chronic Kidney Disease | Foundation Model | 0.8341 | 0.0021 | 0.8299 | 0.8383 |
| Chronic Kidney Disease | All Model Covariates | 0.8328 | 0.0026 | 0.8277 | 0.8379 |
| Chronic Kidney Disease | Ensemble | 0.8412 | 0.0024 | 0.8366 | 0.8458 |
| Crohns Disease | Age + Sex + BMI | 0.5358 | 0.0125 | 0.5112 | 0.5604 |
| Crohns Disease | Foundation Model | 0.6195 | 0.0106 | 0.5987 | 0.6403 |
| Crohns Disease | All Model Covariates | 0.6747 | 0.0128 | 0.6497 | 0.6997 |
| Crohns Disease | Ensemble | 0.6632 | 0.0106 | 0.6425 | 0.6840 |
| Dementia | Age + Sex + BMI | 0.8022 | 0.0013 | 0.7996 | 0.8048 |
| Dementia | Foundation Model | 0.8260 | 0.0024 | 0.8213 | 0.8307 |
| Dementia | All Model Covariates | 0.8187 | 0.0016 | 0.8155 | 0.8219 |
| Dementia | Ensemble | 0.8309 | 0.0015 | 0.8280 | 0.8339 |
| Depressive Episode | Age + Sex + BMI | 0.5755 | 0.0028 | 0.5701 | 0.5809 |
| Depressive Episode | Foundation Model | 0.6301 | 0.0048 | 0.6207 | 0.6395 |
| Depressive Episode | All Model Covariates | 0.6305 | 0.0049 | 0.6209 | 0.6402 |
| Depressive Episode | Ensemble | 0.6387 | 0.0052 | 0.6285 | 0.6489 |
| Digestive Liver Mortality | Age + Sex + BMI | 0.7011 | 0.0096 | 0.6823 | 0.7198 |
| Digestive Liver Mortality | Foundation Model | 0.8181 | 0.0060 | 0.8064 | 0.8299 |
| Digestive Liver Mortality | All Model Covariates | 0.8375 | 0.0175 | 0.8032 | 0.8718 |
| Digestive Liver Mortality | Ensemble | 0.8468 | 0.0103 | 0.8266 | 0.8670 |
| Epilepsy | Age + Sex + BMI | 0.6050 | 0.0101 | 0.5852 | 0.6247 |
| Epilepsy | Foundation Model | 0.6281 | 0.0060 | 0.6163 | 0.6398 |
| Epilepsy | All Model Covariates | 0.6384 | 0.0080 | 0.6227 | 0.6540 |
| Epilepsy | Ensemble | 0.6491 | 0.0074 | 0.6345 | 0.6637 |
| Heart Failure | Age + Sex + BMI | 0.7610 | 0.0020 | 0.7570 | 0.7650 |
| Heart Failure | Foundation Model | 0.8080 | 0.0031 | 0.8020 | 0.8141 |
| Heart Failure | All Model Covariates | 0.7982 | 0.0025 | 0.7933 | 0.8031 |
| Heart Failure | Ensemble | 0.8102 | 0.0023 | 0.8057 | 0.8148 |
| Hypertension | Age + Sex + BMI | 0.6822 | 0.0031 | 0.6761 | 0.6883 |
| Hypertension | Foundation Model | 0.7545 | 0.0029 | 0.7489 | 0.7601 |
| Hypertension | All Model Covariates | 0.7549 | 0.0031 | 0.7489 | 0.7610 |
| Hypertension | Ensemble | 0.7600 | 0.0030 | 0.7542 | 0.7659 |
| Ischaemic Stroke | Age + Sex + BMI | 0.7160 | 0.0046 | 0.7070 | 0.7250 |
| Ischaemic Stroke | Foundation Model | 0.7426 | 0.0042 | 0.7344 | 0.7508 |
| Ischaemic Stroke | All Model Covariates | 0.7456 | 0.0040 | 0.7378 | 0.7533 |
| Ischaemic Stroke | Ensemble | 0.7515 | 0.0041 | 0.7435 | 0.7596 |
| Metabolic Mortality | Age + Sex + BMI | 0.7335 | 0.0126 | 0.7088 | 0.7583 |
| Metabolic Mortality | Foundation Model | 0.8237 | 0.0085 | 0.8070 | 0.8403 |
| Metabolic Mortality | All Model Covariates | 0.8388 | 0.0119 | 0.8154 | 0.8621 |
| Metabolic Mortality | Ensemble | 0.8482 | 0.0092 | 0.8301 | 0.8662 |
| Multiple Sclerosis | Age + Sex + BMI | 0.6104 | 0.0330 | 0.5457 | 0.6750 |
| Multiple Sclerosis | Foundation Model | 0.5743 | 0.0273 | 0.5207 | 0.6278 |
| Multiple Sclerosis | All Model Covariates | 0.6366 | 0.0318 | 0.5743 | 0.6988 |
| Multiple Sclerosis | Ensemble | 0.6263 | 0.0314 | 0.5648 | 0.6878 |
| Myocardial Infarction | Age + Sex + BMI | 0.7075 | 0.0039 | 0.6998 | 0.7152 |
| Myocardial Infarction | Foundation Model | 0.7452 | 0.0037 | 0.7380 | 0.7525 |
| Myocardial Infarction | All Model Covariates | 0.7458 | 0.0038 | 0.7385 | 0.7532 |
| Myocardial Infarction | Ensemble | 0.7518 | 0.0034 | 0.7451 | 0.7585 |
| Neurodegenerative Mortality | Age + Sex + BMI | 0.7808 | 0.0060 | 0.7691 | 0.7925 |
| Neurodegenerative Mortality | Foundation Model | 0.8036 | 0.0070 | 0.7898 | 0.8173 |
| Neurodegenerative Mortality | All Model Covariates | 0.7837 | 0.0162 | 0.7520 | 0.8154 |
| Neurodegenerative Mortality | Ensemble | 0.8086 | 0.0084 | 0.7922 | 0.8251 |
| Osteoporosis | Age + Sex + BMI | 0.7556 | 0.0040 | 0.7477 | 0.7634 |
| Osteoporosis | Foundation Model | 0.7820 | 0.0040 | 0.7742 | 0.7898 |
| Osteoporosis | All Model Covariates | 0.7786 | 0.0022 | 0.7743 | 0.7830 |
| Osteoporosis | Ensemble | 0.7864 | 0.0040 | 0.7786 | 0.7941 |
| Parkinsons Disease | Age + Sex + BMI | 0.7634 | 0.0056 | 0.7524 | 0.7744 |
| Parkinsons Disease | Foundation Model | 0.7522 | 0.0076 | 0.7372 | 0.7671 |
| Parkinsons Disease | All Model Covariates | 0.7700 | 0.0078 | 0.7547 | 0.7853 |
| Parkinsons Disease | Ensemble | 0.7723 | 0.0079 | 0.7568 | 0.7878 |
| Pneumonia | Age + Sex + BMI | 0.6893 | 0.0042 | 0.6811 | 0.6974 |
| Pneumonia | Foundation Model | 0.7513 | 0.0031 | 0.7453 | 0.7573 |
| Pneumonia | All Model Covariates | 0.7422 | 0.0035 | 0.7353 | 0.7491 |
| Pneumonia | Ensemble | 0.7536 | 0.0034 | 0.7470 | 0.7602 |
| Pulmonary Embolism | Age + Sex + BMI | 0.6713 | 0.0075 | 0.6567 | 0.6860 |
| Pulmonary Embolism | Foundation Model | 0.6878 | 0.0051 | 0.6778 | 0.6978 |
| Pulmonary Embolism | All Model Covariates | 0.7008 | 0.0063 | 0.6886 | 0.7131 |
| Pulmonary Embolism | Ensemble | 0.7017 | 0.0053 | 0.6914 | 0.7121 |
| Respiratory Mortality | Age + Sex + BMI | 0.7613 | 0.0109 | 0.7400 | 0.7826 |
| Respiratory Mortality | Foundation Model | 0.8607 | 0.0026 | 0.8556 | 0.8658 |
| Respiratory Mortality | All Model Covariates | 0.8719 | 0.0140 | 0.8444 | 0.8993 |
| Respiratory Mortality | Ensemble | 0.8807 | 0.0063 | 0.8684 | 0.8931 |
| Rheumatoid Arthritis | Age + Sex + BMI | 0.6486 | 0.0169 | 0.6155 | 0.6816 |
| Rheumatoid Arthritis | Foundation Model | 0.6820 | 0.0000 | 0.6820 | 0.6820 |
| Rheumatoid Arthritis | All Model Covariates | 0.7660 | 0.0142 | 0.7382 | 0.7937 |
| Rheumatoid Arthritis | Ensemble | 0.7429 | 0.0255 | 0.6930 | 0.7928 |
| Stroke | Age + Sex + BMI | 0.6979 | 0.0030 | 0.6920 | 0.7038 |
| Stroke | Foundation Model | 0.7235 | 0.0017 | 0.7202 | 0.7268 |
| Stroke | All Model Covariates | 0.7247 | 0.0016 | 0.7214 | 0.7279 |
| Stroke | Ensemble | 0.7317 | 0.0013 | 0.7291 | 0.7343 |

S.d 0.000 is due to instability issues - only one fold succeeded.

Table S13: Binning Strategy Summary

| Category | N Modalities | Mean Bins | Median Bins | Min Bins | Max Bins |
|---|---|---|---|---|---|
| DEXA | 93 | 16.9785 | 17 | 16 | 17 |
| Sleep | 74 | 16.1892 | 18 | 1 | 25 |
| Blood Tests | 69 | 16.0725 | 17 | 2 | 27 |
| ECG | 61 | 16.5574 | 17 | 3 | 30 |
| Ultrasound | 37 | 16.7838 | 17 | 15 | 17 |
| Diet | 35 | 18.8857 | 19 | 17 | 19 |
| Questionnaires | 31 | 14 | 14 | 6 | 27 |
| Body Measures | 29 | 15.2759 | 17 | 0 | 17 |
| Nightingale | 20 | 16.9 | 17 | 16 | 17 |
| Retina | 18 | 16.0556 | 17 | 0 | 17 |
| Genetics | 16 | 16.3125 | 16 | 15 | 17 |
| ABI / Vascular | 12 | 17 | 17 | 17 | 17 |
| Wearable | 12 | 20.25 | 19.5 | 10 | 33 |
| Location | 5 | 13.4 | 16 | 3 | 16 |
| MicroStats | 3 | 17 | 17 | 17 | 17 |
| CGM | 1 | 129 | 129 | 129 | 129 |
| Other | 1 | 38 | 38 | 38 | 38 |

S14 Complete modality list

| Modality | Category | Type | Bins | N Patients | % Coverage | Total Tokens |
|---|---|---|---|---|---|---|
| ABI / Vascular (12 modalities) | | | | | | |
| L Ankle Pressure | ABI / Vascular | continuous | 17 | 12857 | 83.9 | 23015 |
| R Ankle Pressure | ABI / Vascular | continuous | 17 | 12856 | 83.9 | 23014 |
| L Abi | ABI / Vascular | continuous | 17 | 12852 | 83.9 | 22987 |
| R Abi | ABI / Vascular | continuous | 17 | 12842 | 83.8 | 22971 |
| R Brachial Pressure | ABI / Vascular | continuous | 17 | 12815 | 83.6 | 22917 |
| L Brachial Pressure | ABI / Vascular | continuous | 17 | 12751 | 83.2 | 22724 |
| From L Thigh To L Ankle Distance | ABI / Vascular | continuous | 17 | 11220 | 73.2 | 14398 |
| From R Thigh To R Ankle Distance | ABI / Vascular | continuous | 17 | 11219 | 73.2 | 14394 |
| From R Thigh To R Ankle Transit | ABI / Vascular | continuous | 17 | 11216 | 73.2 | 14389 |
| From L Thigh To L Ankle Transit | ABI / Vascular | continuous | 17 | 11206 | 73.2 | 14377 |
| From L Thigh To L Ankle Pwv | ABI / Vascular | continuous | 17 | 10629 | 69.4 | 13470 |
| From R Thigh To R Ankle Pwv | ABI / Vascular | continuous | 17 | 10626 | 69.4 | 13468 |
| Blood Tests (69 modalities) | | | | | | |
| Lymphocytes % | Blood Tests | continuous | 18 | 13022 | 85 | 36782 |
| Mchc | Blood Tests | continuous | 18 | 13018 | 85 | 36740 |
| Neutrophils % | Blood Tests | continuous | 18 | 13014 | 85 | 36721 |
| Monocytes Abs | Blood Tests | continuous | 18 | 13011 | 84.9 | 36577 |
| Lymphocytes Abs | Blood Tests | continuous | 18 | 13004 | 84.9 | 36557 |
| Basophils % | Blood Tests | continuous | 16 | 13004 | 84.9 | 36516 |
| Hemoglobin | Blood Tests | continuous | 18 | 13001 | 84.9 | 36532 |
| Monocytes % | Blood Tests | continuous | 18 | 12992 | 84.8 | 36449 |
| Hct | Blood Tests | continuous | 18 | 12989 | 84.8 | 36082 |
| Eosinophils % | Blood Tests | continuous | 18 | 12968 | 84.6 | 36366 |
| Mcv | Blood Tests | continuous | 18 | 12958 | 84.6 | 36523 |
| Neutrophils Abs | Blood Tests | continuous | 18 | 12955 | 84.6 | 36025 |
| Mch | Blood Tests | continuous | 18 | 12945 | 84.5 | 36269 |
| Rdw | Blood Tests | continuous | 18 | 12939 | 84.5 | 36110 |
| Wbc | Blood Tests | continuous | 18 | 12922 | 84.4 | 35914 |
| Eosinophils Abs | Blood Tests | continuous | 17 | 12895 | 84.2 | 35449 |
| Mean Platelet Volume | Blood Tests | continuous | 18 | 12889 | 84.1 | 35311 |
| Rbc | Blood Tests | continuous | 18 | 12867 | 84 | 35660 |
| Platelets | Blood Tests | continuous | 18 | 12838 | 83.8 | 35536 |
| Basophils Abs | Blood Tests | continuous | 12 | 12508 | 81.6 | 31416 |
| Glucose | Blood Tests | continuous | 17 | 10859 | 70.9 | 23758 |
| Creatinine | Blood Tests | continuous | 17 | 10850 | 70.8 | 24027 |
| Alt Gpt | Blood Tests | continuous | 17 | 10339 | 67.5 | 22141 |
| Potassium | Blood Tests | continuous | 14 | 10233 | 66.8 | 21506 |
| Sodium | Blood Tests | continuous | 13 | 10217 | 66.7 | 21406 |
| Triglycerides | Blood Tests | continuous | 17 | 10085 | 65.8 | 20955 |
| Total Cholesterol | Blood Tests | continuous | 17 | 10057 | 65.6 | 20970 |
| Hdl Cholesterol | Blood Tests | continuous | 17 | 9970 | 65.1 | 20767 |
| Non Hdl Cholesterol | Blood Tests | continuous | 17 | 9639 | 62.9 | 19317 |
| Alkaline Phosphatase | Blood Tests | continuous | 17 | 9459 | 61.8 | 19107 |
| Urea | Blood Tests | continuous | 17 | 9413 | 61.4 | 19926 |
| Ast Got | Blood Tests | continuous | 16 | 9191 | 60 | 18751 |
| Tsh | Blood Tests | continuous | 17 | 9116 | 59.5 | 17280 |
| Calcium | Blood Tests | continuous | 17 | 8505 | 55.5 | 16101 |
| Albumin | Blood Tests | continuous | 16 | 8149 | 53.2 | 15199 |
| Protein Total | Blood Tests | continuous | 17 | 7841 | 51.2 | 14483 |
| Vitamin B12 | Blood Tests | continuous | 17 | 7445 | 48.6 | 12190 |
| Bilirubin Total | Blood Tests | continuous | 15 | 7183 | 46.9 | 12675 |
| Ferritin | Blood Tests | continuous | 17 | 7035 | 45.9 | 11574 |
| Uric Acid | Blood Tests | continuous | 17 | 6944 | 45.3 | 12352 |
| Hba1C | Blood Tests | continuous | 15 | 6677 | 43.6 | 10386 |
| Iron | Blood Tests | continuous | 17 | 5962 | 38.9 | 9815 |
| Vitamin D | Blood Tests | continuous | 17 | 5721 | 37.4 | 8339 |
| Ldl Cholesterol | Blood Tests | continuous | 17 | 5714 | 37.3 | 10571 |
| Urine Ph | Blood Tests | continuous | 10 | 5631 | 36.8 | 8976 |
| Folic Acid | Blood Tests | continuous | 17 | 5032 | 32.8 | 7011 |
| Phosphorus | Blood Tests | continuous | 17 | 4820 | 31.5 | 8242 |
| Egfr | Blood Tests | continuous | 16 | 4433 | 28.9 | 8553 |
| Ldh | Blood Tests | continuous | 17 | 4388 | 28.6 | 7510 |
| Crp Wide Range | Blood Tests | continuous | 13 | 3927 | 25.6 | 5226 |
| Luc % | Blood Tests | continuous | 17 | 3546 | 23.2 | 7285 |
| Luc Abs | Blood Tests | continuous | 9 | 3541 | 23.1 | 7261 |
| Rbc Hypo % | Blood Tests | continuous | 17 | 3434 | 22.4 | 7095 |
| Rbc Micro% | Blood Tests | continuous | 12 | 3407 | 22.2 | 7063 |
| Rbc Macro% | Blood Tests | continuous | 17 | 3275 | 21.4 | 6695 |
| Transferrin | Blood Tests | continuous | 17 | 3264 | 21.3 | 4680 |
| Rbc Hyper % | Blood Tests | continuous | 13 | 3145 | 20.5 | 6320 |
| Rbc Micro Hypo Ratio | Blood Tests | continuous | 13 | 3020 | 19.7 | 6436 |
| Bilirubin Direct | Blood Tests | continuous | 17 | 2619 | 17.1 | 4725 |
| T4 Free | Blood Tests | continuous | 17 | 2126 | 13.9 | 3384 |
| Pt | Blood Tests | continuous | 16 | 1816 | 11.8 | 2164 |
| Urine Sg | Blood Tests | continuous | 17 | 1771 | 11.6 | 2615 |
| Magnesium | Blood Tests | continuous | 12 | 1735 | 11.3 | 2367 |
| mmature Granulocytes % | Blood Tests | continuous | 6 | 883 | 5.8 | 1521 |
| mmature Granulocytes | Blood Tests | continuous | 27 | 878 | 5.7 | 1513 |
| Pt % | Blood Tests | continuous | 15 | 761 | 5 | 921 |
| Estradiol E2 | Blood Tests | continuous | 16 | 742 | 4.8 | 896 |
| Creatine Kinase | Blood Tests | continuous | 13 | 100 | 0.6 | 173 |
| Ebv Igg | Blood Tests | continuous | 2 | 100 | 0.6 | 101 |
| Body Measures (26 modalities) | | | | | | |
| Height | Body Measures | continuous | 17 | 13293 | 86.8 | 25067 |
| Body Weight | Body Measures | continuous | 17 | 13292 | 86.8 | 25062 |
| Waist | Body Measures | continuous | 17 | 13291 | 86.8 | 25098 |
| Sitting Blood Pressure Pulse Rate | Body Measures | continuous | 17 | 13290 | 86.8 | 25082 |
| Bmi | Body Measures | continuous | 17 | 13289 | 86.8 | 25054 |
| Sitting Blood Pressure Systolic | Body Measures | continuous | 17 | 13287 | 86.7 | 25070 |
| Sitting Blood Pressure Diastolic | Body Measures | continuous | 17 | 13286 | 86.7 | 25078 |
| Hips | Body Measures | continuous | 17 | 13277 | 86.7 | 25059 |
| Hand Grip Right | Body Measures | continuous | 17 | 13273 | 86.6 | 25022 |
| Hand Grip Left | Body Measures | continuous | 17 | 13270 | 86.6 | 25013 |
| Lying Blood Pressure Pulse Rate | Body Measures | continuous | 17 | 11794 | 77 | 15924 |
| Lying Blood Pressure Diastolic | Body Measures | continuous | 17 | 11791 | 77 | 15922 |
| Lying Blood Pressure Systolic | Body Measures | continuous | 17 | 11789 | 77 | 15916 |
| Standing One Min Blood Pressure Systolic | Body Measures | continuous | 17 | 11761 | 76.8 | 15879 |
| Standing One Min Blood Pressure Diastolic | Body Measures | continuous | 17 | 11757 | 76.8 | 15873 |
| Standing Three Min Blood Pressure Systolic | Body Measures | continuous | 17 | 11755 | 76.7 | 15856 |
| Standing One Min Blood Pressure Pulse | Body Measures | continuous | 17 | 11754 | 76.7 | 15870 |
| Standing Three Min Blood Pressure Diastolic | Body Measures | continuous | 17 | 11749 | 76.7 | 15848 |
| Standing Three Min Blood Pressure Pulse | Body Measures | continuous | 17 | 11749 | 76.7 | 15843 |
| Trunk Fat | Body Measures | continuous | 16 | 3746 | 24.4 | 3908 |
| Body Fat | Body Measures | continuous | 16 | 3745 | 24.4 | 3907 |
| Bmr | Body Measures | continuous | 17 | 3744 | 24.4 | 3905 |
| Fingers Grip Right | Body Measures | continuous | 17 | 1312 | 8.6 | 1312 |
| Fingers Grip Left | Body Measures | continuous | 17 | 1309 | 8.5 | 1309 |
| Number Of Days From Last Period | Body Measures | continuous | 15 | 630 | 4.1 | 656 |
| Number Of Days In Cycle | Body Measures | continuous | 17 | 546 | 3.6 | 570 |
| CGM (1 modalities) | | | | | | |
| Glucosevalue | CGM | continuous | 129 | 7349 | 48 | 8122068 |
| DEXA (95 modalities) | | | | | | |
| Body Total Bmc | DEXA | continuous | 17 | 11902 | 77.7 | 20364 |
| Body Trunk Area | DEXA | continuous | 17 | 11902 | 77.7 | 20363 |
| Body Total Area | DEXA | continuous | 17 | 11902 | 77.7 | 20363 |
| Body Pelvis Area | DEXA | continuous | 17 | 11902 | 77.7 | 20362 |
| Body Arm Right Bmd | DEXA | continuous | 17 | 11901 | 77.7 | 20360 |
| Body Head Bmd | DEXA | continuous | 17 | 11901 | 77.7 | 20359 |
| Body Arms Area | DEXA | continuous | 17 | 11900 | 77.7 | 20358 |
| Body Head Bmc | DEXA | continuous | 17 | 11900 | 77.7 | 20357 |
| Body Ribs Bmc | DEXA | continuous | 17 | 11900 | 77.7 | 20355 |
| Body Spine Area | DEXA | continuous | 17 | 11900 | 77.7 | 20354 |
| Body Head Area | DEXA | continuous | 17 | 11899 | 77.7 | 20344 |
| Body Comp Arm Left Tissue Percent | DEXA | continuous | 16 | 11876 | 77.5 | 20275 |
| Body Comp Android Bone Mass | DEXA | continuous | 17 | 11875 | 77.5 | 20275 |
| Body Comp Trunk Right Bone Mass | DEXA | continuous | 17 | 11875 | 77.5 | 20273 |
| Body Comp Leg Right Bone Mass | DEXA | continuous | 17 | 11874 | 77.5 | 20274 |
| Body Comp Total Tissue Mass | DEXA | continuous | 17 | 11874 | 77.5 | 20273 |
| Body Comp Trunk Left Bone Mass | DEXA | continuous | 17 | 11874 | 77.5 | 20271 |
| Body Comp Arms Total Mass | DEXA | continuous | 17 | 11874 | 77.5 | 20271 |
| Body Comp Arms Bone Mass | DEXA | continuous | 17 | 11874 | 77.5 | 20266 |
| Body Comp Gynoid Bone Mass | DEXA | continuous | 17 | 11873 | 77.5 | 20267 |
| Body Comp Trunk Diff Bone Mass | DEXA | continuous | 16 | 11872 | 77.5 | 20201 |
| Body Comp Arms Fat Mass | DEXA | continuous | 17 | 11855 | 77.4 | 20224 |
| Body Comp Legs Fat Mass | DEXA | continuous | 17 | 11854 | 77.4 | 20229 |
| Body Comp Arm Left Fat Mass | DEXA | continuous | 17 | 11851 | 77.4 | 20218 |
| Body Comp Arm Right Fat Mass | DEXA | continuous | 17 | 11849 | 77.4 | 20221 |
| Body Comp Trunk Diff Fat Free Mass | DEXA | continuous | 17 | 11820 | 77.2 | 20051 |
| Body Comp Trunk Diff Lean Mass | DEXA | continuous | 17 | 11815 | 77.1 | 20070 |
| Body Comp Total Diff Tissue Mass | DEXA | continuous | 17 | 11815 | 77.1 | 20065 |
| Body Comp Total Diff Fat Free Mass | DEXA | continuous | 17 | 11815 | 77.1 | 20039 |
| Body Comp Trunk Diff Tissue Mass | DEXA | continuous | 17 | 11811 | 77.1 | 20063 |
| Total Scan Sat Volume | DEXA | continuous | 17 | 11808 | 77.1 | 20089 |
| ody Comp Total Diff Lean Mass | DEXA | continuous | 17 | 11808 | 77.1 | 20017 |
| ody Comp Total Diff Total Mass | DEXA | continuous | 17 | 11807 | 77.1 | 19900 |
| ody Comp Trunk Diff Region Percent | DEXA | continuous | 17 | 11806 | 77.1 | 19611 |
| ody Comp Trunk Diff Total Mass | DEXA | continuous | 17 | 11799 | 77 | 19869 |
| pine L4 Average Height | DEXA | continuous | 0 | 11777 | 76.9 | 19880 |
| pine L4 Am Percent | DEXA | continuous | 17 | 11775 | 76.9 | 19864 |
| ody Comp Legs Diff Bone Mass | DEXA | continuous | 17 | 11763 | 76.8 | 19796 |
| emur Right Troch Area | DEXA | continuous | 17 | 11759 | 76.8 | 19775 |
| ody Comp Total Diff Region Percent | DEXA | continuous | 17 | 11759 | 76.8 | 19342 |
| emur Right Shaft Area | DEXA | continuous | 17 | 11757 | 76.8 | 19774 |
| otal Scan Sat Mass | DEXA | continuous | 17 | 11755 | 76.7 | 19924 |
| pine L3 L4 Area | DEXA | continuous | 17 | 11752 | 76.7 | 19810 |
| pine L3 L4 Bmc | DEXA | continuous | 17 | 11751 | 76.7 | 19800 |
| otal Scan Vat Volume | DEXA | continuous | 17 | 11742 | 76.6 | 19954 |
| emur Left Wards Area | DEXA | continuous | 17 | 11736 | 76.6 | 19728 |
| otal Scan Vat Area | DEXA | continuous | 17 | 11735 | 76.6 | 19923 |
| ody Comp Total Diff Fat Mass | DEXA | continuous | 17 | 11733 | 76.6 | 19813 |
| ody Comp Legs Diff Tissue Percent | DEXA | continuous | 17 | 11707 | 76.4 | 19355 |
| ody Comp Legs Diff Region Percent | DEXA | continuous | 17 | 11707 | 76.4 | 19330 |
| ody Comp Trunk Diff Fat Mass | DEXA | continuous | 17 | 11699 | 76.4 | 19663 |
| pine L1 L4 Average Height | DEXA | continuous | 0 | 11697 | 76.4 | 19645 |
| Body Comp Arm Right Bone Mass | DEXA | continuous | 17 | 11668 | 76.2 | 19735 |
| Femur Shaft Diff Bmd | DEXA | continuous | 17 | 11668 | 76.2 | 19559 |
| Femur Upper Neck Diff Bmd | DEXA | continuous | 17 | 11668 | 76.2 | 19515 |
| Femur Troch Diff Bmd | DEXA | continuous | 17 | 11665 | 76.2 | 19528 |
| Femur Shaft Diff Area | DEXA | continuous | 17 | 11661 | 76.1 | 19509 |
| Femur Shaft Diff Bmc | DEXA | continuous | 17 | 11658 | 76.1 | 19537 |
| Femur Total Diff Bmd | DEXA | continuous | 17 | 11658 | 76.1 | 19519 |
| Femur Neck Diff Bmd | DEXA | continuous | 17 | 11658 | 76.1 | 19503 |
| Femur Lower Neck Diff Bmd | DEXA | continuous | 17 | 11656 | 76.1 | 19499 |
| Body Comp Legs Diff Tissue Mass | DEXA | continuous | 17 | 11651 | 76.1 | 19538 |
| Femur Wards Diff Bmd | DEXA | continuous | 17 | 11650 | 76 | 19505 |
| Femur Troch Diff Bmc | DEXA | continuous | 17 | 11643 | 76 | 19458 |
| Femur Troch Diff Area | DEXA | continuous | 17 | 11641 | 76 | 19451 |
| Femur Total Diff Bmc | DEXA | continuous | 17 | 11638 | 76 | 19491 |
| Body Comp Legs Diff Total Mass | DEXA | continuous | 17 | 11628 | 75.9 | 19132 |
| Femur Total Diff Area | DEXA | continuous | 17 | 11617 | 75.8 | 19408 |
| Body Comp Legs Diff Lean Mass | DEXA | continuous | 17 | 11615 | 75.8 | 19454 |
| Body Comp Legs Diff Fat Free Mass | DEXA | continuous | 17 | 11613 | 75.8 | 19484 |
| Femur Upper Neck Diff Bmc | DEXA | continuous | 17 | 11592 | 75.7 | 19253 |
| Femur Neck Diff Bmc | DEXA | continuous | 17 | 11588 | 75.6 | 19315 |
| Femur Lower Neck Diff Bmc | DEXA | continuous | 17 | 11565 | 75.5 | 19190 |
| Femur Wards Diff Bmc | DEXA | continuous | 17 | 11558 | 75.4 | 19249 |
| Body Comp Arm Left Bone Mass | DEXA | continuous | 17 | 11557 | 75.4 | 19473 |
| Femur Wards Diff Area | DEXA | continuous | 17 | 11538 | 75.3 | 19198 |
| Body Comp Legs Diff Fat Mass | DEXA | continuous | 17 | 11534 | 75.3 | 19137 |
| Body Comp Total Diff Bone Mass | DEXA | continuous | 17 | 11512 | 75.2 | 18949 |
| Femur Lower Neck Diff Area | DEXA | continuous | 17 | 11494 | 75 | 18967 |
| Femur Upper Neck Diff Area | DEXA | continuous | 17 | 11491 | 75 | 18952 |
| Spine L1 L4 Area | DEXA | continuous | 17 | 11430 | 74.6 | 17231 |
| Spine L1 L4 Bmc | DEXA | continuous | 17 | 11426 | 74.6 | 17226 |
| Total Scan Vat Mass | DEXA | continuous | 17 | 11356 | 74.1 | 19185 |
| Body Comp Arms Diff Bone Mass | DEXA | continuous | 17 | 10635 | 69.4 | 16767 |
| Body Comp Arms Diff Tissue Percent | DEXA | continuous | 17 | 10581 | 69.1 | 16643 |
| Body Comp Arms Diff Region Percent | DEXA | continuous | 17 | 10572 | 69 | 16614 |
| Body Comp Arms Diff Tissue Mass | DEXA | continuous | 17 | 10375 | 67.7 | 16328 |
| Body Comp Arms Diff Lean Mass | DEXA | continuous | 17 | 10369 | 67.7 | 16285 |
| Body Comp Arms Diff Fat Free Mass | DEXA | continuous | 17 | 10341 | 67.5 | 16288 |
| Body Comp Arms Diff Total Mass | DEXA | continuous | 17 | 10270 | 67 | 15795 |
| Body Comp Arms Diff Fat Mass | DEXA | continuous | 17 | 10115 | 66 | 15558 |
| Femur Neck Mean Z Score | DEXA | continuous | 17 | 6535 | 42.7 | 9665 |
| Femur Left Total T Score | DEXA | continuous | 17 | 6248 | 40.8 | 9077 |
| Spine L2 T Score | DEXA | continuous | 17 | 6189 | 40.4 | 10138 |
| Femur Left Upper Neck T Score | DEXA | continuous | 17 | 6184 | 40.4 | 8618 |
| Diet (35 modalities) | | | | | | |
| Weight | Diet | continuous | 19 | 7349 | 48 | 471827 |
| Water | Diet | continuous | 19 | 7349 | 48 | 471421 |
| Energy | Diet | continuous | 19 | 7349 | 48 | 471044 |
| Calcium, Ca | Diet | continuous | 19 | 7349 | 48 | 467968 |
| Carbohydrate, By Difference | Diet | continuous | 19 | 7349 | 48 | 467437 |
| Iron, Fe | Diet | continuous | 19 | 7349 | 48 | 466286 |
| Sodium, Na | Diet | continuous | 19 | 7349 | 48 | 461439 |
| Protein | Diet | continuous | 19 | 7349 | 48 | 458453 |
| Fatty Acids, Total Polyunsaturated | Diet | continuous | 19 | 7349 | 48 | 455243 |
| Total Lipid (Fat) | Diet | continuous | 19 | 7349 | 48 | 454135 |
| Fatty Acids, Total Monounsaturated | Diet | continuous | 19 | 7349 | 48 | 448689 |
| Sugars, Total | Diet | continuous | 19 | 7349 | 48 | 448665 |
| Fatty Acids, Total Saturated | Diet | continuous | 19 | 7349 | 48 | 448541 |
| Vitamin B-6 | Diet | continuous | 19 | 7349 | 48 | 447268 |
| Vitamin E (Alpha-Tocopherol) | Diet | continuous | 19 | 7349 | 48 | 439189 |
| Vitamin K | Diet | continuous | 19 | 7349 | 48 | 436039 |
| Starch | Diet | continuous | 19 | 7349 | 48 | 400118 |
| Vitamin A, Rae | Diet | continuous | 19 | 7349 | 48 | 389728 |
| Fatty Acids, Total Trans-Polyenoic | Diet | continuous | 19 | 7348 | 48 | 404639 |
| Vitamin A, Iu | Diet | continuous | 19 | 7348 | 48 | 388898 |
| Fatty Acids, Total Trans-Monoenoic | Diet | continuous | 19 | 7348 | 48 | 381732 |
| Glucose | Diet | continuous | 19 | 7347 | 48 | 436712 |
| Fiber, Total Dietary | Diet | continuous | 19 | 7347 | 48 | 402637 |
| Sucrose | Diet | continuous | 19 | 7345 | 48 | 426283 |
| Vitamin B-12 | Diet | continuous | 19 | 7345 | 48 | 314837 |
| Vitamin C, Total Ascorbic Acid | Diet | continuous | 19 | 7344 | 47.9 | 344351 |
| Fructose | Diet | continuous | 19 | 7344 | 47.9 | 342969 |
| Cholesterol | Diet | continuous | 19 | 7341 | 47.9 | 297062 |
| Maltose | Diet | continuous | 19 | 7341 | 47.9 | 264694 |
| Fatty Acids, Total Trans | Diet | continuous | 19 | 7338 | 47.9 | 219809 |
| Vitamin D (D2 + D3) | Diet | continuous | 19 | 7335 | 47.9 | 247595 |
| Lactose | Diet | continuous | 18 | 7319 | 47.8 | 217067 |
| Galactose | Diet | continuous | 19 | 7315 | 47.8 | 134228 |
| Caffeine | Diet | continuous | 17 | 7234 | 47.2 | 170660 |
| Alcohol, Ethyl | Diet | continuous | 18 | 5422 | 35.4 | 24500 |
| ECG (61 modalities) | | | | | | |
| R R Ms | ECG | continuous | 17 | 13085 | 85.4 | 23929 |
| Qt Ms | ECG | continuous | 17 | 13079 | 85.4 | 23896 |
| R Ms Ii | ECG | continuous | 16 | 13077 | 85.4 | 23912 |
| Qtc Ms | ECG | continuous | 17 | 13074 | 85.3 | 23897 |
| R Mv V4 | ECG | continuous | 17 | 13073 | 85.3 | 23889 |
| R Ms I | ECG | continuous | 14 | 13060 | 85.2 | 23860 |
| R Mv V3 | ECG | continuous | 17 | 13055 | 85.2 | 23844 |
| R Ms V6 | ECG | continuous | 14 | 13047 | 85.2 | 23798 |
| R Ms V5 | ECG | continuous | 13 | 13035 | 85.1 | 23728 |
| R Ms V3 | ECG | continuous | 14 | 13028 | 85 | 23758 |
| Qrs Ms | ECG | continuous | 16 | 13028 | 85 | 23757 |
| R Ms V4 | ECG | continuous | 13 | 13021 | 85 | 23659 |
| Pr Ms | ECG | continuous | 17 | 12999 | 84.9 | 23603 |
| P Ms | ECG | continuous | 16 | 12993 | 84.8 | 23541 |
| R Mv V2 | ECG | continuous | 17 | 12992 | 84.8 | 23617 |
| R Ms Avf | ECG | continuous | 17 | 12989 | 84.8 | 23631 |
| T Mv I | ECG | continuous | 17 | 12974 | 84.7 | 23586 |
| R Ms V2 | ECG | continuous | 15 | 12962 | 84.6 | 23540 |
| T Mv V3 | ECG | continuous | 17 | 12915 | 84.3 | 23366 |
| T Mv V4 | ECG | continuous | 17 | 12871 | 84 | 23225 |
| T Mv V2 | ECG | continuous | 17 | 12789 | 83.5 | 23027 |
| R Ms Iii | ECG | continuous | 17 | 12777 | 83.4 | 22963 |
| R Mv Iii | ECG | continuous | 17 | 12747 | 83.2 | 22888 |
| P Axis | ECG | continuous | 17 | 12693 | 82.9 | 22753 |
| R Ms V1 | ECG | continuous | 14 | 12665 | 82.7 | 22679 |
| R Mv V1 | ECG | continuous | 17 | 12636 | 82.5 | 22604 |
| R Ms Avl | ECG | continuous | 17 | 12604 | 82.3 | 22740 |
| P Mv Ii | ECG | continuous | 15 | 12595 | 82.2 | 22438 |
| St Mv V3 | ECG | continuous | 17 | 12530 | 81.8 | 22118 |
| St Mv V1 | ECG | continuous | 15 | 12518 | 81.7 | 21947 |
| T Axis | ECG | continuous | 17 | 12485 | 81.5 | 22127 |
| P Mv V3 | ECG | continuous | 13 | 12342 | 80.6 | 21480 |
| P Mv V4 | ECG | continuous | 12 | 12299 | 80.3 | 21199 |
| P Mv I | ECG | continuous | 27 | 12011 | 78.4 | 20657 |
| P Mv V2 | ECG | continuous | 15 | 11937 | 77.9 | 20304 |
| P Mv V6 | ECG | continuous | 11 | 11888 | 77.6 | 20206 |
| P Mv Avf | ECG | continuous | 30 | 11697 | 76.4 | 20129 |
| Qrs Axis | ECG | continuous | 17 | 11299 | 73.8 | 19967 |
| St Mv V5 | ECG | continuous | 14 | 10931 | 71.4 | 17854 |
| T Mv Avl | ECG | continuous | 17 | 10831 | 70.7 | 18657 |
| R Ms Avr | ECG | continuous | 25 | 10765 | 70.3 | 18422 |
| J Mv V2 | ECG | continuous | 16 | 10716 | 70 | 17370 |
| P Mv V1 | ECG | continuous | 15 | 10608 | 69.2 | 16649 |
| P Mv Iii | ECG | continuous | 13 | 9966 | 65.1 | 16266 |
| J Mv V3 | ECG | continuous | 15 | 9391 | 61.3 | 14622 |
| Q Ms Avr | ECG | continuous | 14 | 8640 | 56.4 | 14223 |
| J Mv V1 | ECG | continuous | 15 | 8166 | 53.3 | 11998 |
| J Mv Avr | ECG | continuous | 20 | 7199 | 47 | 11081 |
| J Mv Ii | ECG | continuous | 24 | 7059 | 46.1 | 10951 |
| J Mv I | ECG | continuous | 21 | 6648 | 43.4 | 10192 |
| Q Ms Avl | ECG | continuous | 22 | 6581 | 43 | 10647 |
| J Mv V6 | ECG | continuous | 22 | 6056 | 39.5 | 8398 |
| T Mv V1 | ECG | continuous | 17 | 5668 | 37 | 8416 |
| Q Ms I | ECG | continuous | 17 | 5192 | 33.9 | 8212 |
| Q Ms Iii | ECG | continuous | 30 | 4662 | 30.4 | 7125 |
| Q Ms Ii | ECG | continuous | 12 | 3754 | 24.5 | 5591 |
| St Mv Avr | ECG | continuous | 12 | 3608 | 23.6 | 4841 |
| Q Ms Avf | ECG | continuous | 16 | 3371 | 22 | 4955 |
| Q Ms V4 | ECG | continuous | 11 | 1319 | 8.6 | 1702 |
| Q Ms V1 | ECG | continuous | 16 | 1025 | 6.7 | 1329 |
| T Mv Avr | ECG | continuous | 3 | 78 | 0.5 | 93 |
| Genetics (16 modalities) | | | | | | |
| Pc5 Prs | Genetics | continuous | 17 | 5660 | 37 | 5676 |
| Pc15 Prs | Genetics | continuous | 17 | 5352 | 34.9 | 5365 |
| Pc6 Prs | Genetics | continuous | 17 | 4730 | 30.9 | 4742 |
| Pc16 Prs | Genetics | continuous | 16 | 4632 | 30.2 | 4646 |
| Pc9 Prs | Genetics | continuous | 16 | 4477 | 29.2 | 4490 |
| Pc8 Prs | Genetics | continuous | 17 | 4429 | 28.9 | 4444 |
| Pc2 Prs | Genetics | continuous | 16 | 4420 | 28.8 | 4429 |
| Pc10 Prs | Genetics | continuous | 16 | 4404 | 28.8 | 4412 |
| Pc14 Prs | Genetics | continuous | 17 | 4206 | 27.5 | 4214 |
| Pc7 Prs | Genetics | continuous | 16 | 4201 | 27.4 | 4212 |
| Pc4 Prs | Genetics | continuous | 17 | 4181 | 27.3 | 4191 |
| Pc11 Prs | Genetics | continuous | 16 | 3678 | 24 | 3691 |
| Pc1 Prs | Genetics | continuous | 16 | 3295 | 21.5 | 3303 |
| Pc3 Prs | Genetics | continuous | 16 | 2969 | 19.4 | 2977 |
| Pc13 Prs | Genetics | continuous | 16 | 1329 | 8.7 | 1336 |
| Pc12 Prs | Genetics | continuous | 15 | 450 | 2.9 | 452 |
| Location (5 modalities) | | | | | | |
| Average Itm X | Location | continuous | 16 | 11117 | 72.6 | 15407 |
| Area Id | Location | continuous | 16 | 11117 | 72.6 | 15407 |
| Average Itm Y | Location | continuous | 16 | 11104 | 72.5 | 15389 |
| Cluster | Location | continuous | 3 | 7234 | 47.2 | 9839 |
| Total Population Of Area | Location | continuous | 16 | 5026 | 32.8 | 6870 |
| Medications (1 modalities) | | | | | | |
| Medication | Medications | categorical | 0 | 4390 | 28.7 | 61668 |
| MicroStats (3 modalities) | | | | | | |
| Simpson Diversity | MicroStats | continuous | 17 | 12610 | 82.3 | 21831 |
| Shannon Diversity | MicroStats | continuous | 17 | 12610 | 82.3 | 21831 |
| Richness | MicroStats | continuous | 17 | 12610 | 82.3 | 21831 |
| Microbiome (1 modalities) | | | | | | |
| Microbiome Species | Microbiome | categorical | 0 | 12627 | 82.4 | 6147075 |
| Nightingale (20 modalities) | | | | | | |
| Ala | Nightingale | continuous | 17 | 8512 | 55.6 | 10243 |
| S Hdl Ce Pct | Nightingale | continuous | 17 | 8502 | 55.5 | 10228 |
| His | Nightingale | continuous | 17 | 8497 | 55.5 | 10223 |
| S Hdl Fc Pct | Nightingale | continuous | 17 | 8496 | 55.5 | 10225 |
| Phe | Nightingale | continuous | 17 | 8495 | 55.4 | 10222 |
| Creatinine | Nightingale | continuous | 16 | 8494 | 55.4 | 10219 |
| Gly | Nightingale | continuous | 17 | 8494 | 55.4 | 10214 |
| Acetate | Nightingale | continuous | 17 | 8489 | 55.4 | 10202 |
| Pyruvate | Nightingale | continuous | 17 | 8486 | 55.4 | 10202 |
| Glucose | Nightingale | continuous | 17 | 8456 | 55.2 | 10156 |
| .actate | Nightingale | continuous | 17 | 8448 | 55.2 | 10125 |
| . Hdl Fc Pct | Nightingale | continuous | 16 | 8407 | 54.9 | 10076 |
| Xl Vldl Fc Pct | Nightingale | continuous | 17 | 8103 | 52.9 | 9629 |
| 3ohbutyrate | Nightingale | continuous | 17 | 7851 | 51.2 | 9269 |
| Xxl Vldl Pl Pct | Nightingale | continuous | 17 | 7681 | 50.1 | 9018 |
| Gln | Nightingale | continuous | 17 | 7539 | 49.2 | 8878 |
| Xxl Vldl Ce Pct | Nightingale | continuous | 17 | 7444 | 48.6 | 8684 |
| Total Bcaa | Nightingale | continuous | 17 | 6272 | 40.9 | 7148 |
| Glycerol | Nightingale | continuous | 17 | 2796 | 18.2 | 2970 |
| Other (7 modalities) | | | | | | |
| Medical Condition Code | Other | categorical | 0 | 14511 | 94.7 | 69298 |
| Birth Event | Other | categorical | 0 | 13339 | 87.1 | 13340 |
| Gender At Birth | Other | categorical | 0 | 13339 | 87.1 | 13340 |
| Age | Other | continuous | 0 | 13306 | 86.9 | 25141 |
| Gender | Other | categorical | 0 | 13303 | 86.8 | 25138 |
| Sleep Event | Other | continuous | 0 | 10729 | 70 | 16246013 |
| Exercise Type | Other | categorical | 0 | 5553 | 36.2 | 125993 |
| RNA (1 modalities) | | | | | | |
| Rna Gene | RNA | categorical | 0 | 3871 | 25.3 | 1908870 |
| Retina (17 modalities) | | | | | | |
| Artery Tortuosity Density R Eye | Retina | continuous | 17 | 9552 | 62.4 | 12508 |
| Vessel Density R Eye | Retina | continuous | 17 | 9552 | 62.4 | 12506 |
| Distance Tortuosity R Eye | Retina | continuous | 17 | 9546 | 62.3 | 12486 |
| Artery Vessel Density R Eye | Retina | continuous | 17 | 9544 | 62.3 | 12498 |
| Vein Vessel Density R Eye | Retina | continuous | 17 | 9542 | 62.3 | 12496 |
| Vein Distance Tortuosity R Eye | Retina | continuous | 17 | 9527 | 62.2 | 12466 |
| Artery Distance Tortuosity R Eye | Retina | continuous | 17 | 9524 | 62.2 | 12457 |
| Artery Tortuosity Density L Eye | Retina | continuous | 17 | 9469 | 61.8 | 12337 |
| Vessel Density L Eye | Retina | continuous | 17 | 9468 | 61.8 | 12334 |
| Artery Vessel Density L Eye | Retina | continuous | 17 | 9465 | 61.8 | 12332 |
| Distance Tortuosity L Eye | Retina | continuous | 17 | 9456 | 61.7 | 12308 |
| Vein Vessel Density L Eye | Retina | continuous | 17 | 9454 | 61.7 | 12321 |
| Vein Distance Tortuosity L Eye | Retina | continuous | 17 | 9445 | 61.7 | 12297 |
| Artery Distance Tortuosity L Eye | Retina | continuous | 17 | 9436 | 61.6 | 12280 |
| Cdr Horizontal R Eye | Retina | continuous | 17 | 3188 | 20.8 | 3297 |
| Cdr Vertical R Eye | Retina | continuous | 17 | 3188 | 20.8 | 3297 |
| Cdr Vertical L Eye | Retina | continuous | 17 | 1924 | 12.6 | 1983 |

4. SUPPLEMENTARY FIGURES

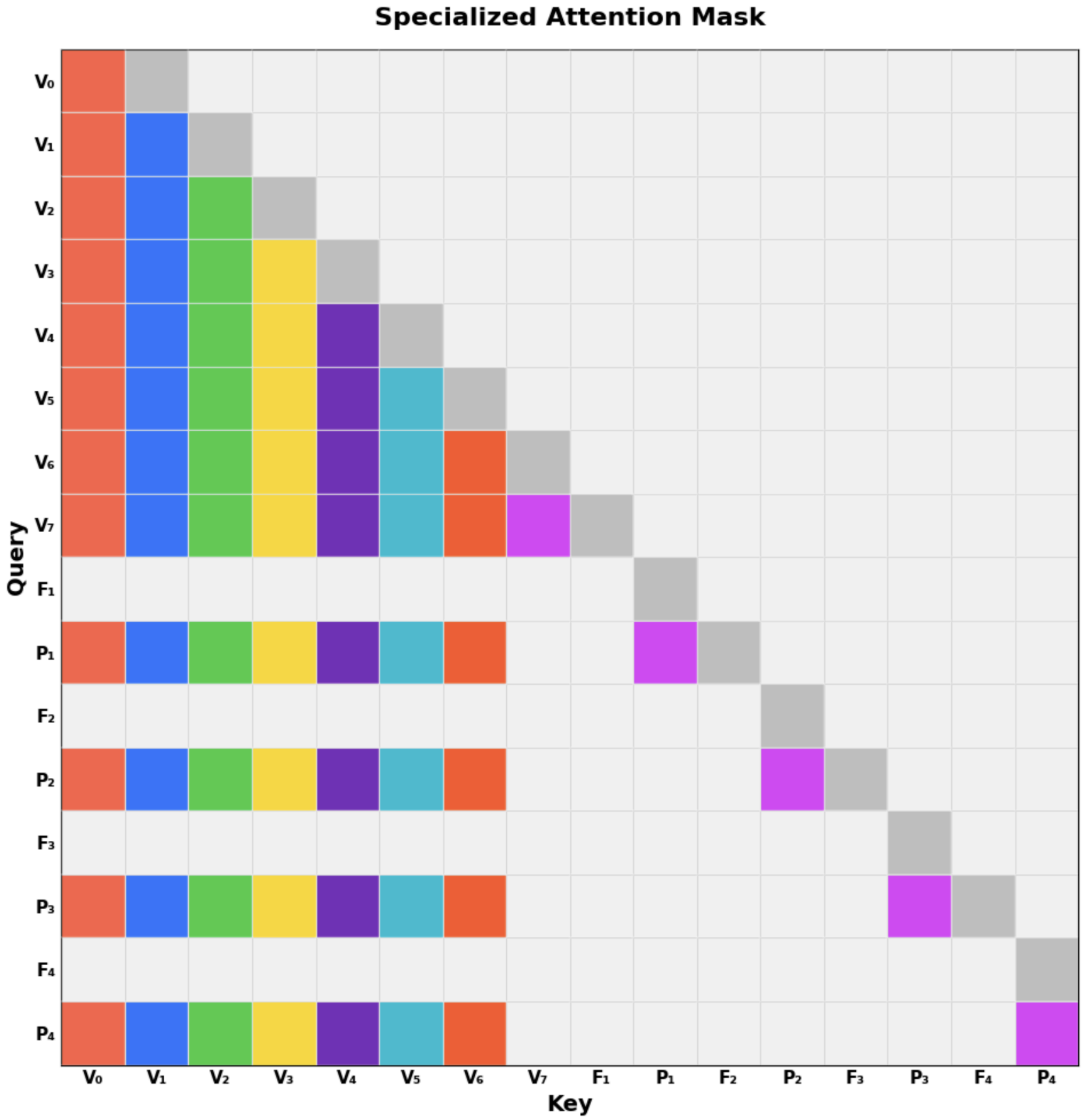

**Fig. S1 | Specialised attention mask for parallel visit-2 prediction.**
Schematic of the attention pattern used to predict every visit-2 parameter in a single forward pass (Methods; cf. main Fig. 2c). Rows are queries, columns are keys; coloured cells indicate attended positions. Visit-1 history tokens $V_0$–$V_7$ attend causally to themselves and to prior history tokens. For each predicted parameter i, an "F_i" probe token (which does not attend to anything except itself) is paired with a "P_i" prediction token; P_i attends to the full visit-1 history $V_0$–$V_7$ and to its own F_i, but is blocked from seeing other F_j or P_j (j≠i). This block-diagonal structure across (F_i, P_i) pairs makes the visit-2 predictions conditionally independent given the shared visit-1 context, so all parameters can be decoded simultaneously without information leakage between target variables. Dark grey cells are the ones being predicted by the colorful sequence to the left of it.

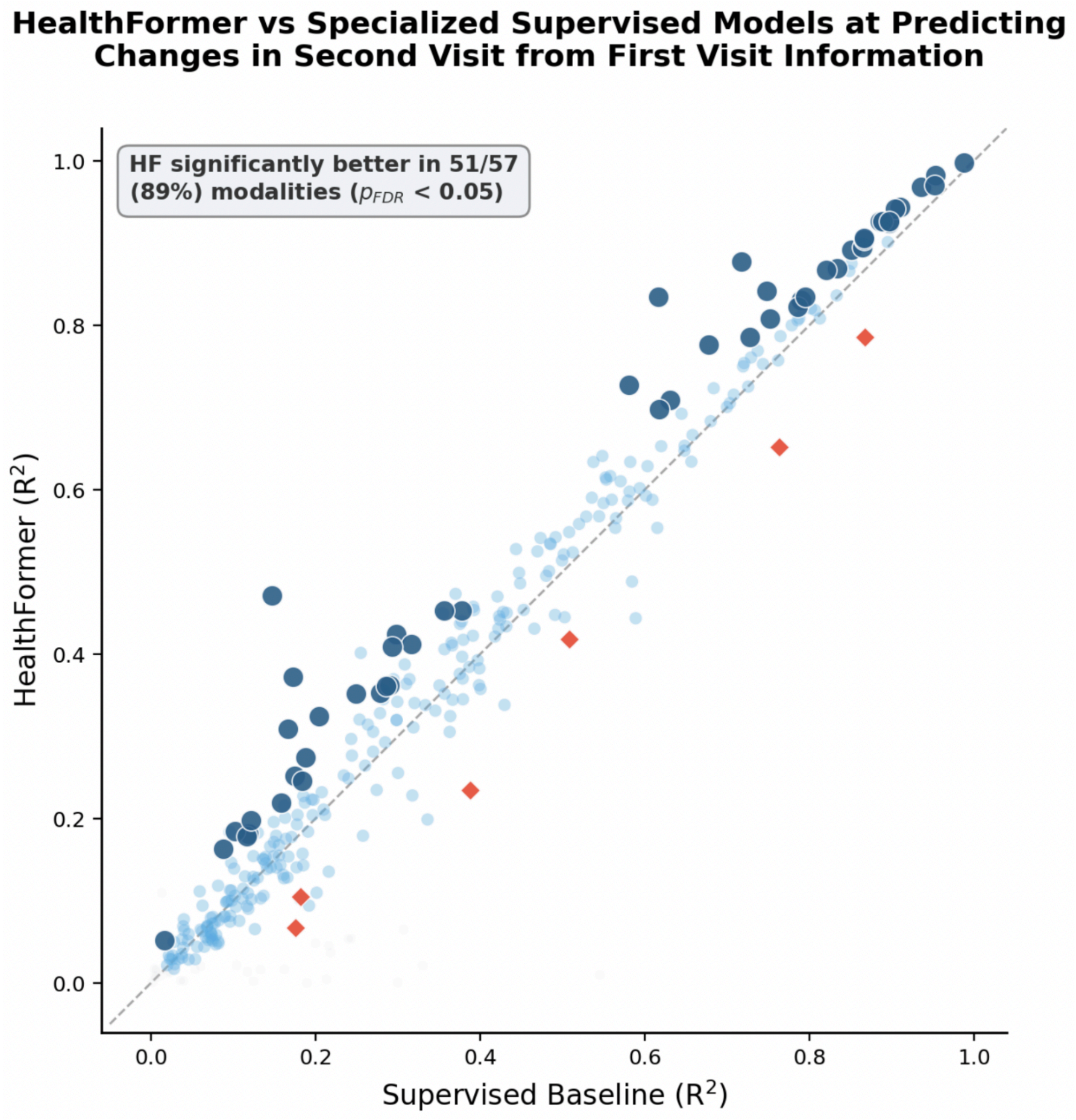

**Fig. S2 | HealthFormer outperforms specialised per-modality supervised baselines on visit-2 prediction.**

Held-out test set (n = 3,217 HPP participants). Each point is one modality; axes show R^2 for predicting the visit-2 value of that modality from visit-1 information. The y-axis is HealthFormer; the x-axis is a modality-specific supervised regression model trained directly on the same input/target pair. The dashed line is the y = x identity. HealthFormer is significantly better than the matched supervised baseline (p_FDR < 0.05) for 51 of 57 modalities with significant differences (89%; dark blue points). Light blue points are modalities with no significant difference; red diamonds are the few cases where the supervised baseline outperforms HealthFormer. The result indicates that the unified generative representation captures shared physiological structure beyond what modality-specific optimisation recovers. Specifically, the six modalities where the supervised baseline significantly outperformed HealthFormer were DEXA Body Head Bmd, DEXA Femur Left Total T Score, DEXA Femur

Wards Diff Bmd, ECG R Ms V1, ECG P Ms, and RetinaScan Automorph Artery Tortuosity Density L Eye.

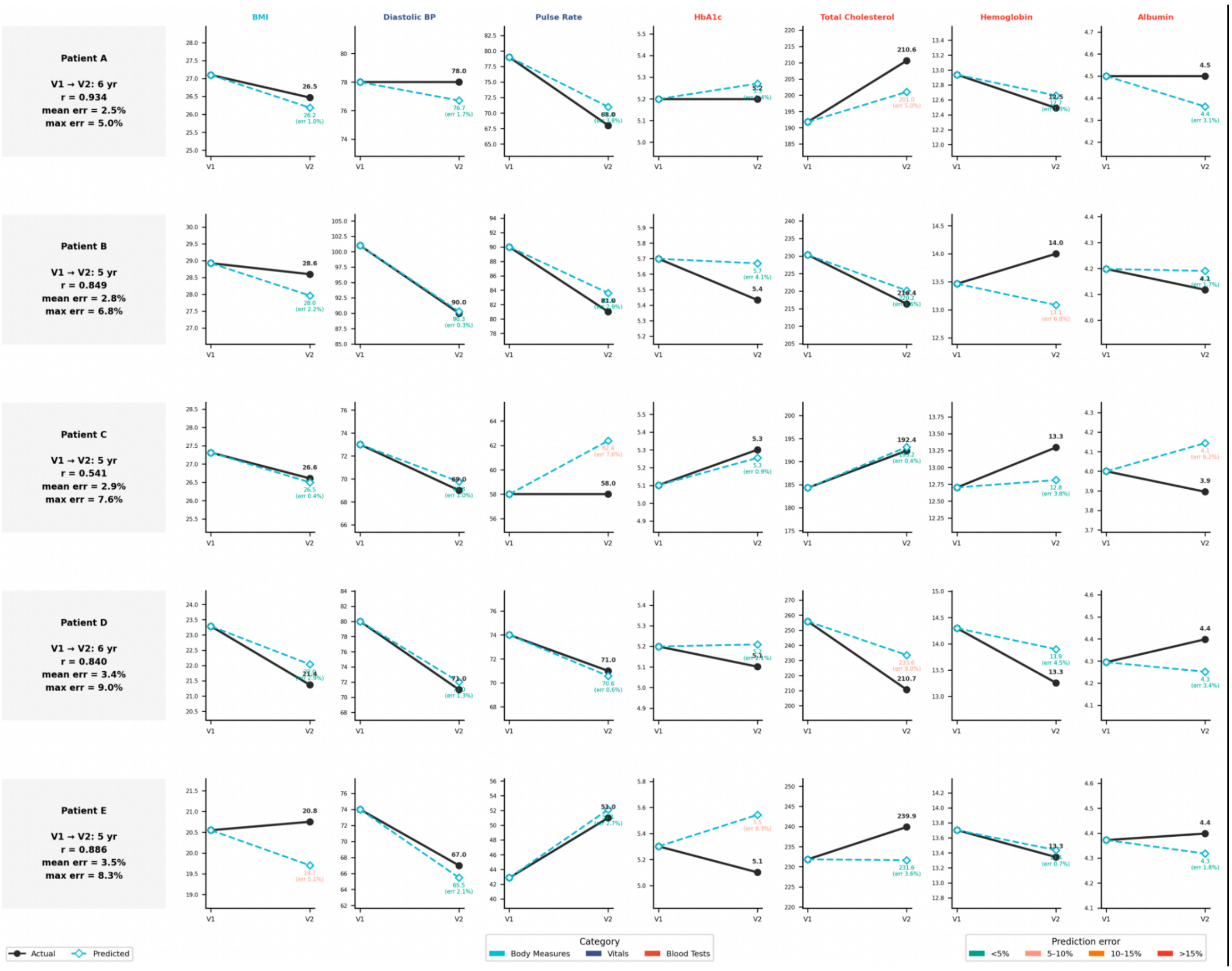

**Fig. S3 | Per-participant longitudinal trajectories across seven biomarkers.**
Five UKBB participants (A–E; rows) shown across seven biomarkers spanning body, vascular, glycaemic, lipid, haematological, and hepatic-renal domains (BMI, diastolic blood pressure, pulse rate, HbA1c, total cholesterol, haemoglobin, albumin; columns). Black filled circles connected by solid lines are observed values at visit 1 (V1) and the follow-up visit (V2); cyan open circles connected by dashed lines are HealthFormer predictions for V2 conditioned only on V1 inputs. The header for each row reports the V1->V2 interval, the participant-level Pearson r between predicted and observed values across the seven biomarkers, and the mean and maximum percentage error. Per-cell colour coding (top legend) groups biomarkers into body measures, vital signs, and blood tests; the bottom colour bar indicates per-cell prediction error magnitude.

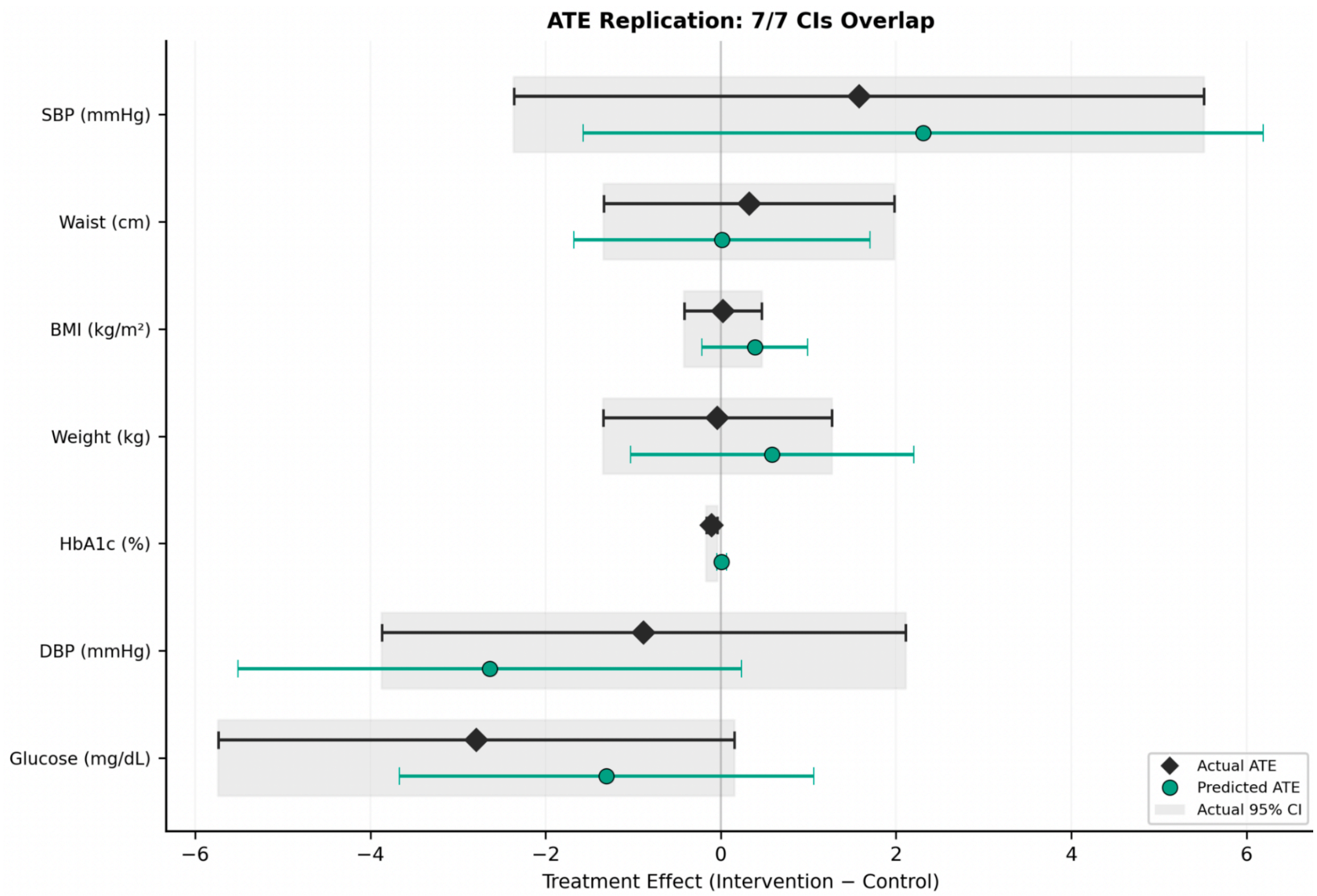

**Fig. S4 | Replication of average treatment effects in PNP3.**

Forest plot comparing the trial-reported average treatment effect (black diamonds, with 95% CI shown as grey shaded bands) against HealthFormer-predicted ATE (teal circles, with 95% CI as teal whiskers) for seven biomarkers in the personalised-nutrition arm of PNP3: SBP, waist circumference, BMI, body weight, HbA1c, DBP, and fasting glucose. Effects are expressed as treatment minus control. The predicted CI overlaps the trial-reported CI in 7 of 7 endpoints, including the two endpoints where the trial reports a directional effect (DBP and glucose).

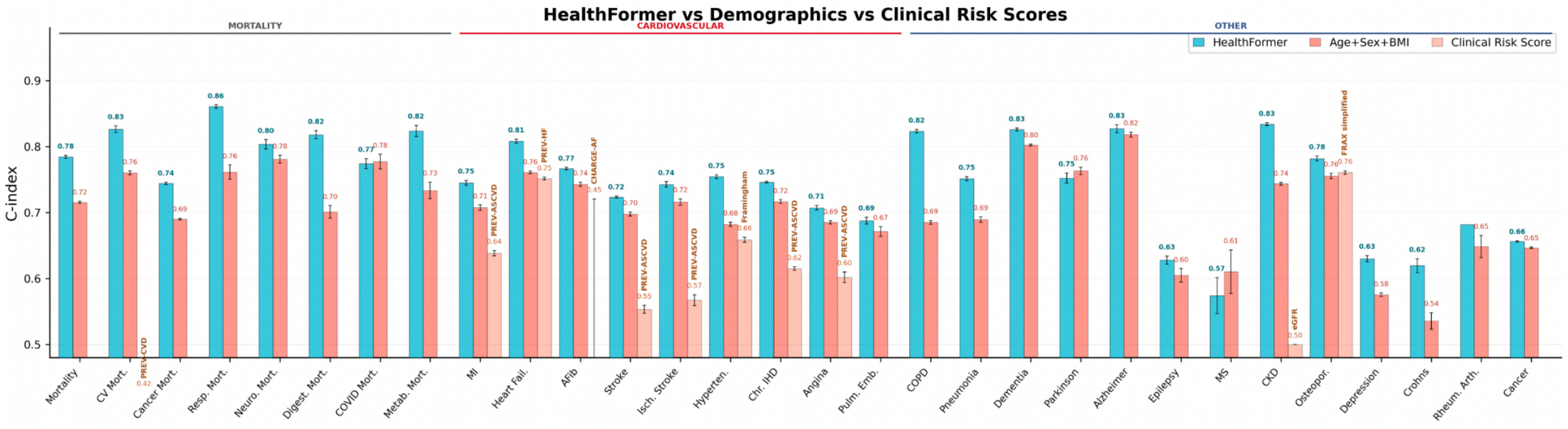

**Fig. S5 | HealthFormer embeddings vs demographics and established clinical risk scores across UK Biobank endpoints.**

Per-endpoint C-index for incident-disease and mortality models built on UK Biobank baseline data. Bars compare HealthFormer embeddings (blue), an age + sex + BMI demographic baseline (red), and the strongest disease-specific clinical risk score where one is established (orange; e.g, PREVENT-ASCVD for ischaemic stroke, SCORE2 for cardiovascular mortality). Endpoints are grouped into mortality outcomes (left block) and incident-disease outcomes (right block). HealthFormer exceeds the demographic baseline for 27 of 30 endpoints, and exceeds the matched clinical score for every endpoint where one is available. Error bars are mean +/- s.d. across five cross-validation folds.

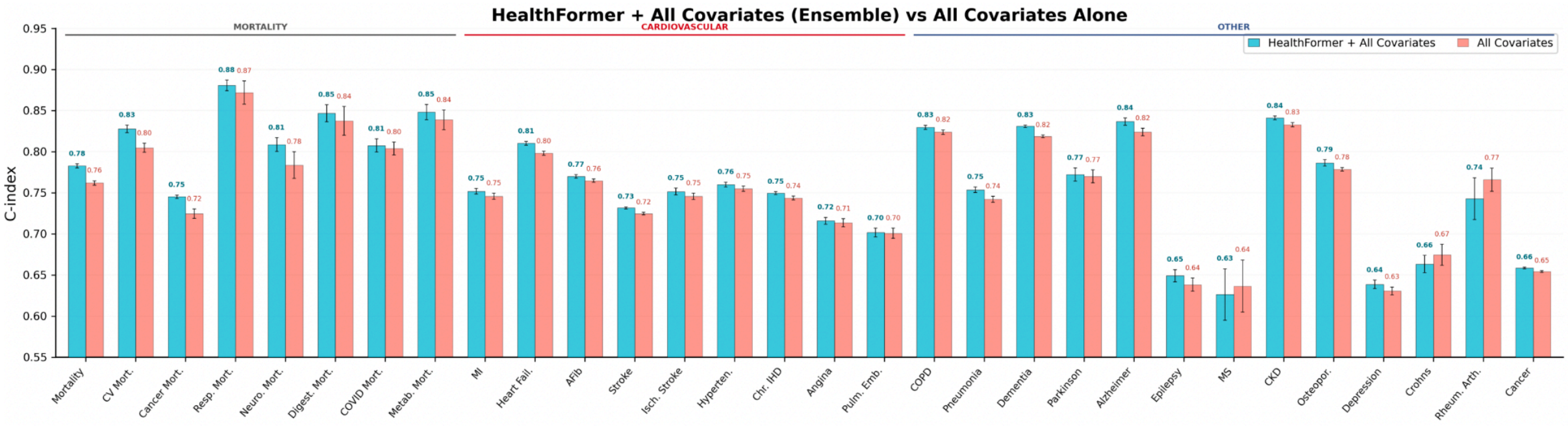

**Fig. S6 | Ensemble of HealthFormer embeddings plus standard covariates improves over a covariate-only model across UK Biobank endpoints.**

Per-endpoint C-index for a penalised Cox model trained on (i) all available covariates including demographics, anthropometrics, blood biomarkers, lifestyle, and medication exposure (red), versus (ii) the same covariates concatenated with HealthFormer embeddings (blue). The ensemble improves over the covariate-only model in 27 of 30 endpoints, with the largest gains for neurodegenerative mortality (+0.025), cardiovascular mortality (+0.023), and all-cause mortality (+0.021). Error bars are mean +/- s.d. across five cross-validation folds.

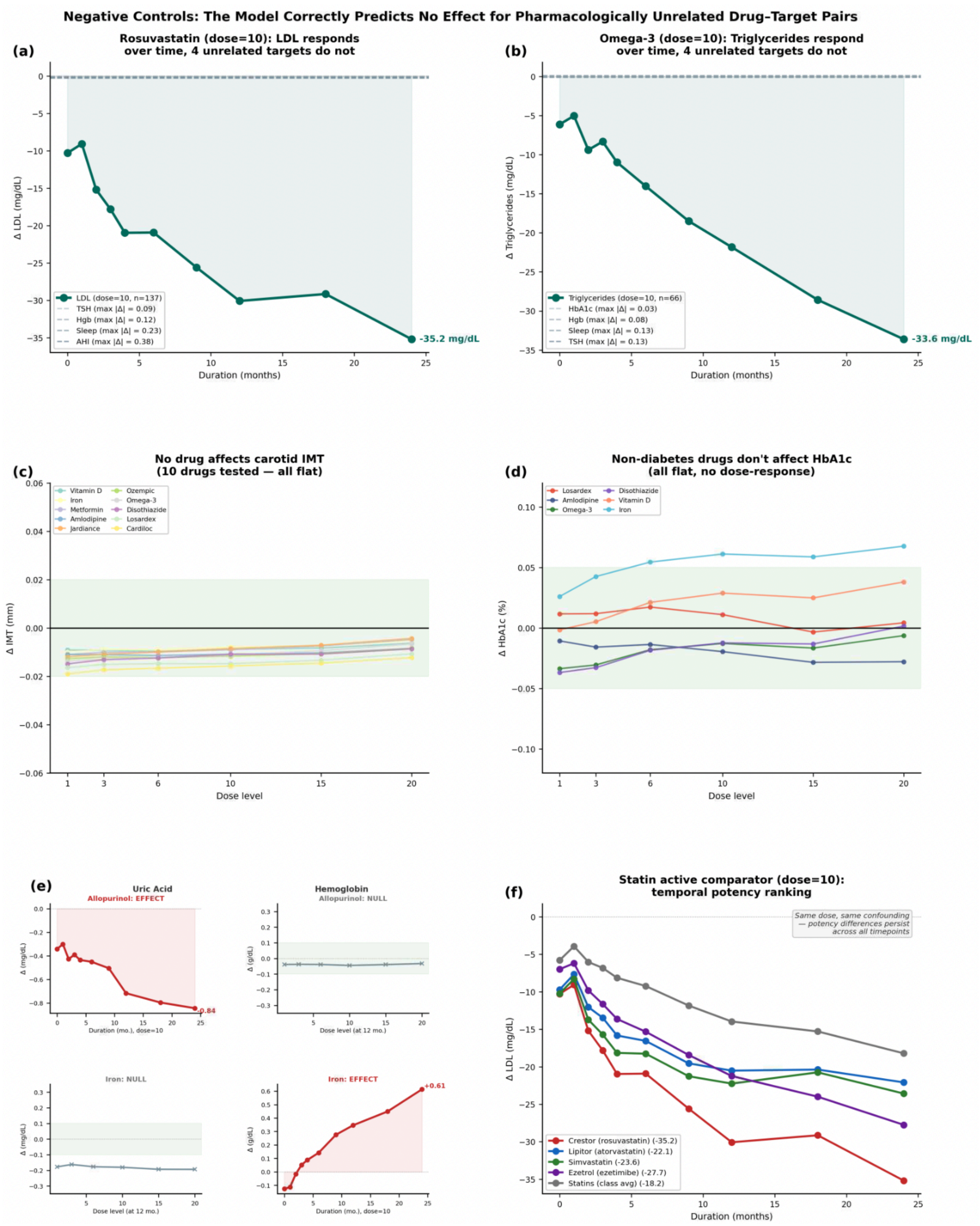

**Fig. S7 | Negative-control simulations: HealthFormer correctly predicts no effect for pharmacologically unrelated drug-target pairs.**

Each panel conditions a synthetic indication-matched population on a candidate intervention and tracks twelve-month predicted trajectories of pharmacologically related and unrelated targets.

a, Rosuvastatin (dose = 10 tokens) specifically reduces LDL cholesterol over time (green curve), while four pharmacologically unrelated targets (TSH, haemoglobin, sleep efficiency, AHI) remain flat (dashed lines), confirming target specificity.

b, Omega-3 supplementation (dose = 10 tokens) specifically reduces triglycerides, while four unrelated targets (HbA1c, haemoglobin, sleep efficiency, TSH) show no response, mirroring the statin pattern in panel a.
c, Carotid intima-media thickness (IMT) is unaffected by any of 10 tested drugs across 6 dose levels (all curves remain within the green null zone), consistent with the known absence of acute pharmacological effects on vascular wall thickness.
d, HbA1c is unaffected by non-diabetes drugs (statins, antihypertensives, omega-3, iron, vitamin D) across dose levels, confirming that only diabetes-targeted interventions shift glycated haemoglobin.
e, Double dissociation between allopurinol and iron. Allopurinol reduces uric acid (top-left, red curve) but does not affect haemoglobin (top-right, flat). Conversely, iron supplementation increases haemoglobin (bottom-right, red curve) but does not affect uric acid (bottom-left, flat). This 2 x 2 pattern rules out generic drug-token confounding and demonstrates mechanism-specific predictions.
f, Active-comparator analysis: five statin-class drugs at the same dose (10 tokens) over time. The model recovers the published potency ranking (rosuvastatin > atorvastatin > simvastatin > ezetimibe + simvastatin > ezetimibe > generic statin token). Because all statins share identical confounding by indication, the observed potency differences reflect learned pharmacological properties rather than observational bias.

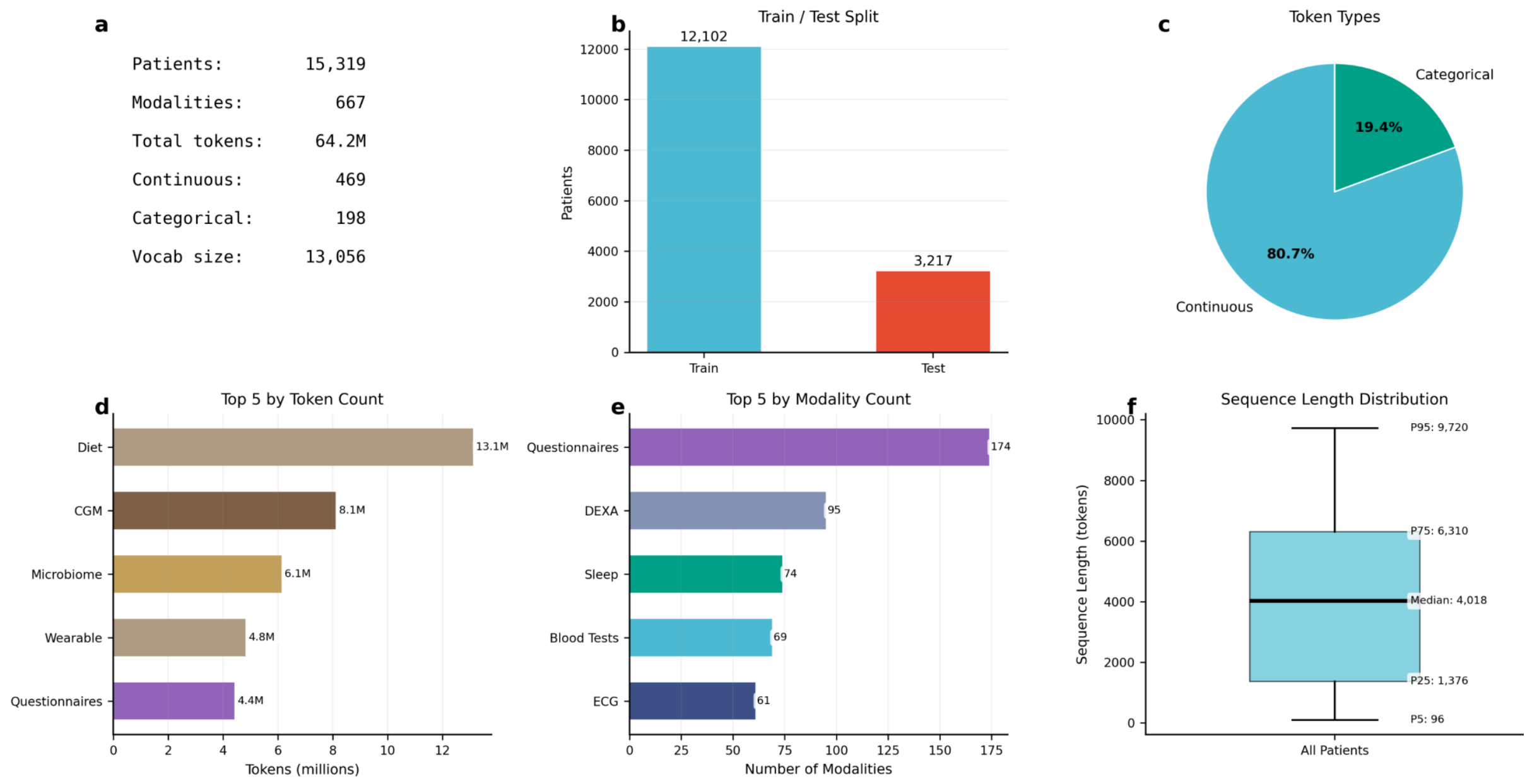

**Fig. S8 | HPP training-corpus overview.**
a, Summary statistics: 15,319 participants, 667 modalities, 64.2 M tokens (469 continuous + 198 categorical modalities; vocabulary size 13,056).
b, Train/test split: 12,102 participants in training, 3,217 held out for testing.
c, Token-type distribution: 80.7% continuous, 19.4% categorical.

d, Top five modality categories by total token count: diet (13.1 M), CGM (8.1 M), microbiome (6.1 M), wearable (4.8 M), questionnaires (4.4 M).
e, Top five modality categories by number of modalities: questionnaires (174), DEXA (95), sleep (74), blood tests (69), ECG (61).
f, Per-participant sequence-length distribution: median 4,018 tokens (P5 = 96, P25 = 1,376, P75 = 6,310, P95 = 9,720).

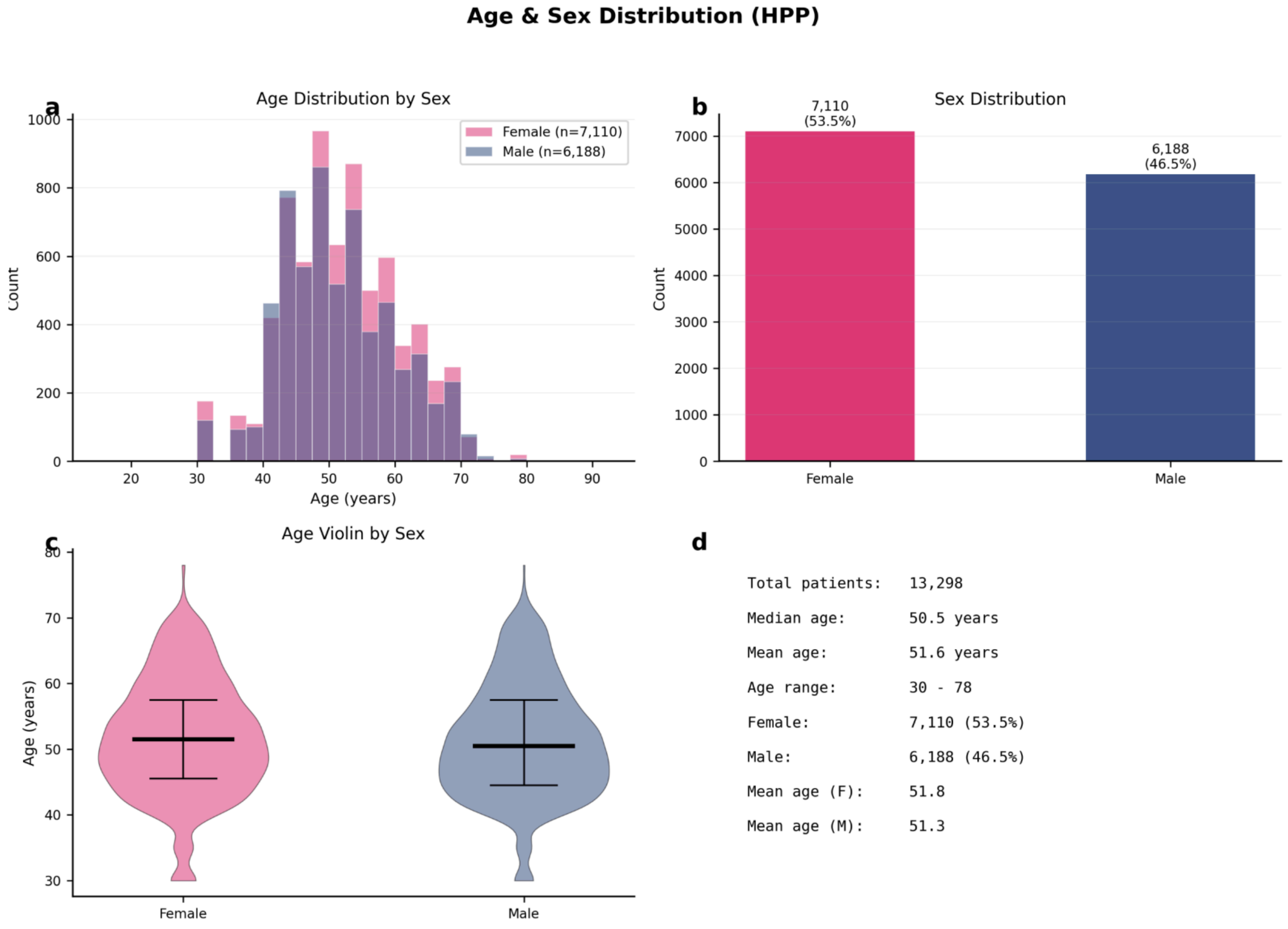

**Fig. S9 | Age and sex distribution of the HPP cohort.**
a, Stacked age histogram by sex (female n = 7,110, male n = 6,188).
b, Sex composition: 53.5% female, 46.5% male.
c, Age distribution by sex shown as violin plots; the male and female distributions are nearly identical (mean female 51.8 yr, mean male 51.3 yr).
d, Summary statistics: 13,298 participants with complete demographic data, median age 50.5 yr, mean 51.6 yr, range 30–78 yr.

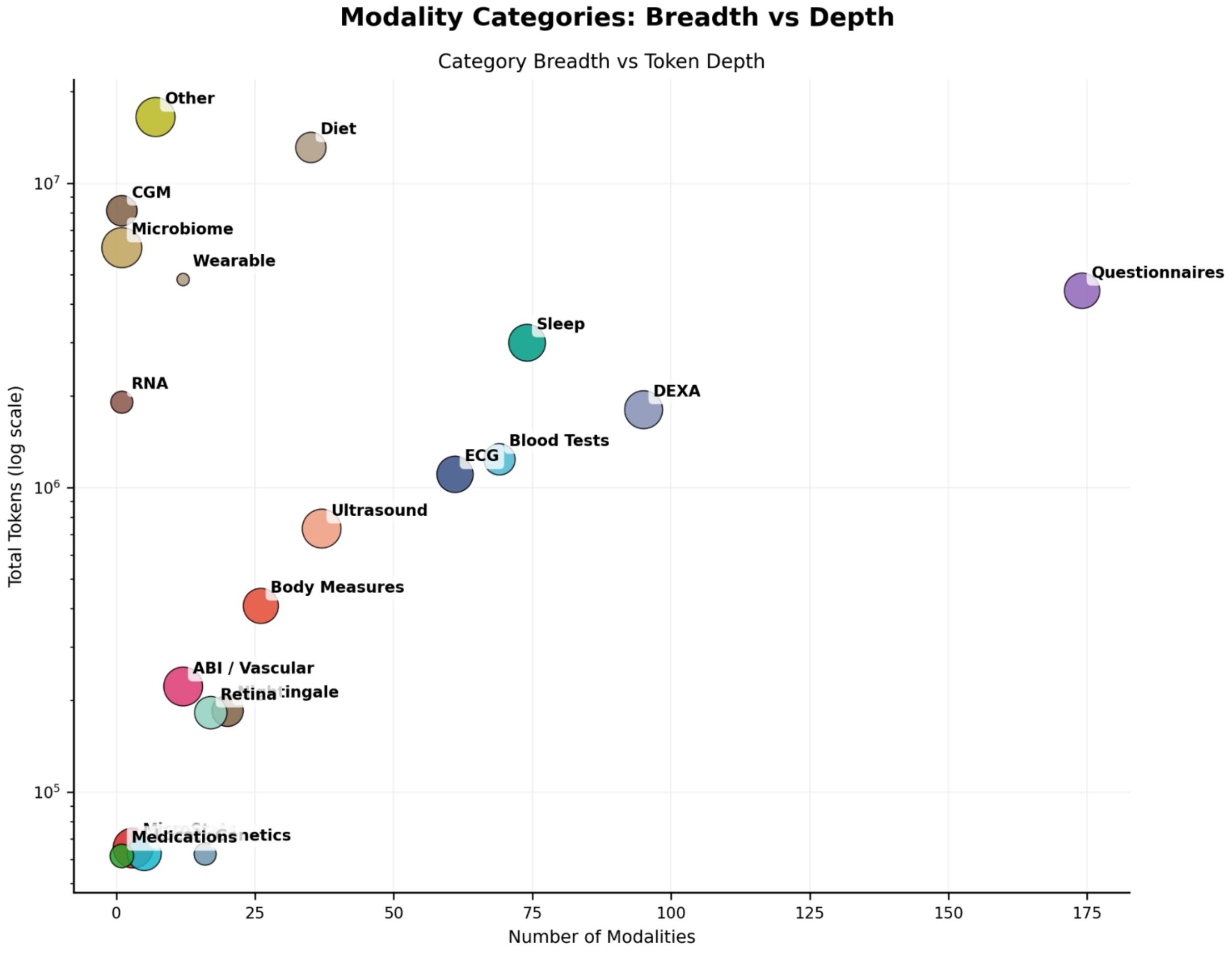

**Fig. S10 | Modality categories: breadth vs depth in the HPP training corpus.**
Each point is a modality category. The x-axis shows category breadth (number of distinct modalities within the category) and the y-axis shows category depth (total tokens, log scale). Bubble area scales with median per-participant coverage. Categories cluster into three regimes: high-breadth, high-depth (questionnaires, DEXA, sleep, blood tests, ECG); low-breadth, high-depth (CGM, microbiome, diet, wearable), where a small number of modalities each contribute many tokens per participant; and low-breadth, low-depth (medications, genetics, RNA), which are sparse on both axes. This shape illustrates the heterogeneity of the data the unified tokenizer must absorb.

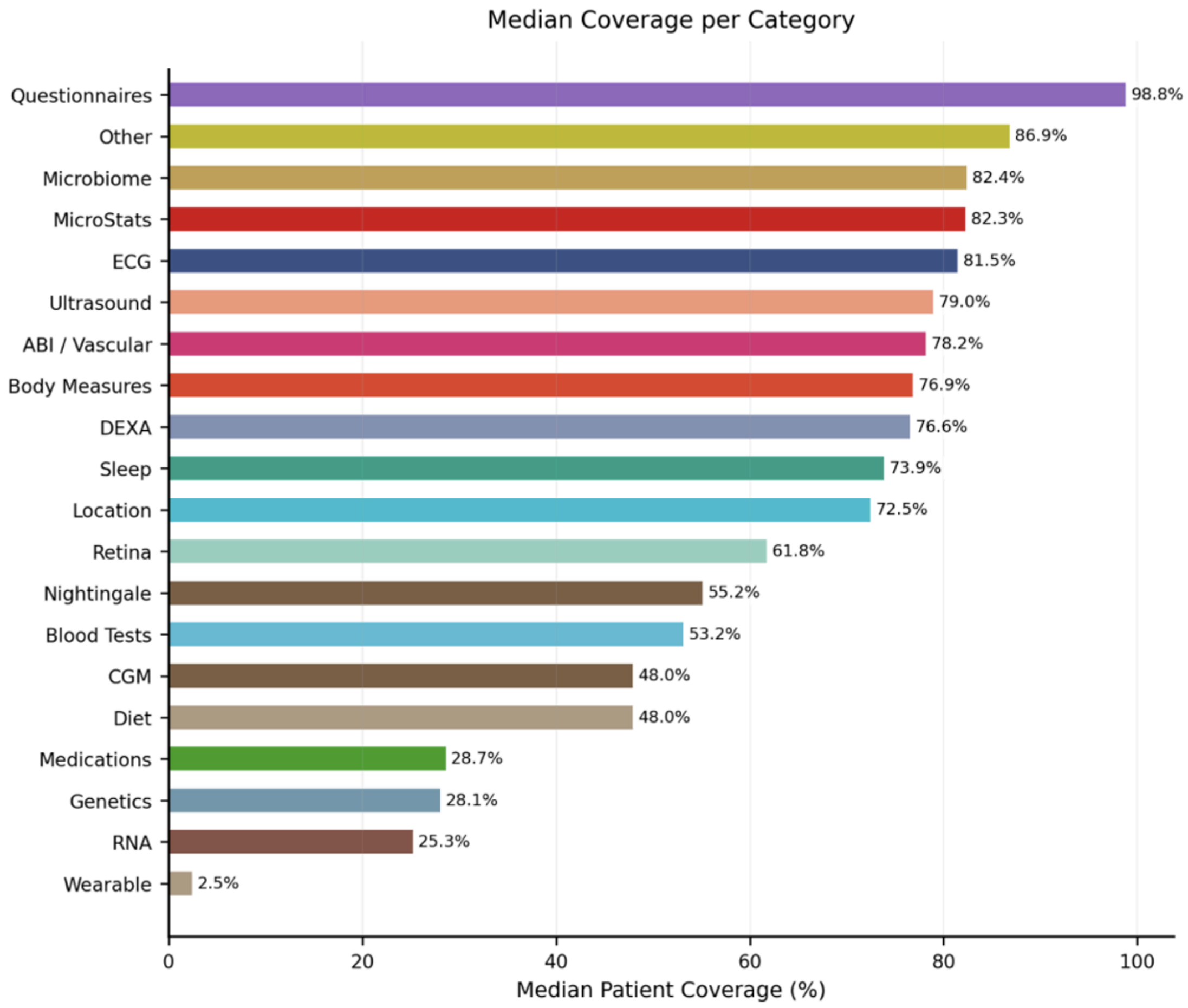

**Fig. S11 | Median per-participant coverage by modality category.**
Bar chart showing the fraction of HPP participants with at least one measurement available in each modality category at baseline. Coverage is highest for questionnaires (98.8%), other (86.9%), microbiome (82.4%), MicroStats (82.3%), and ECG (81.5%); intermediate for DEXA, body measures, ABI/vascular, ultrasound, sleep, and location (73–79%); and lowest for medications (28.7%), genetics (28.1%), RNA (25.3%), and wearable (2.5%, reflecting the smartwatch sub-cohort).

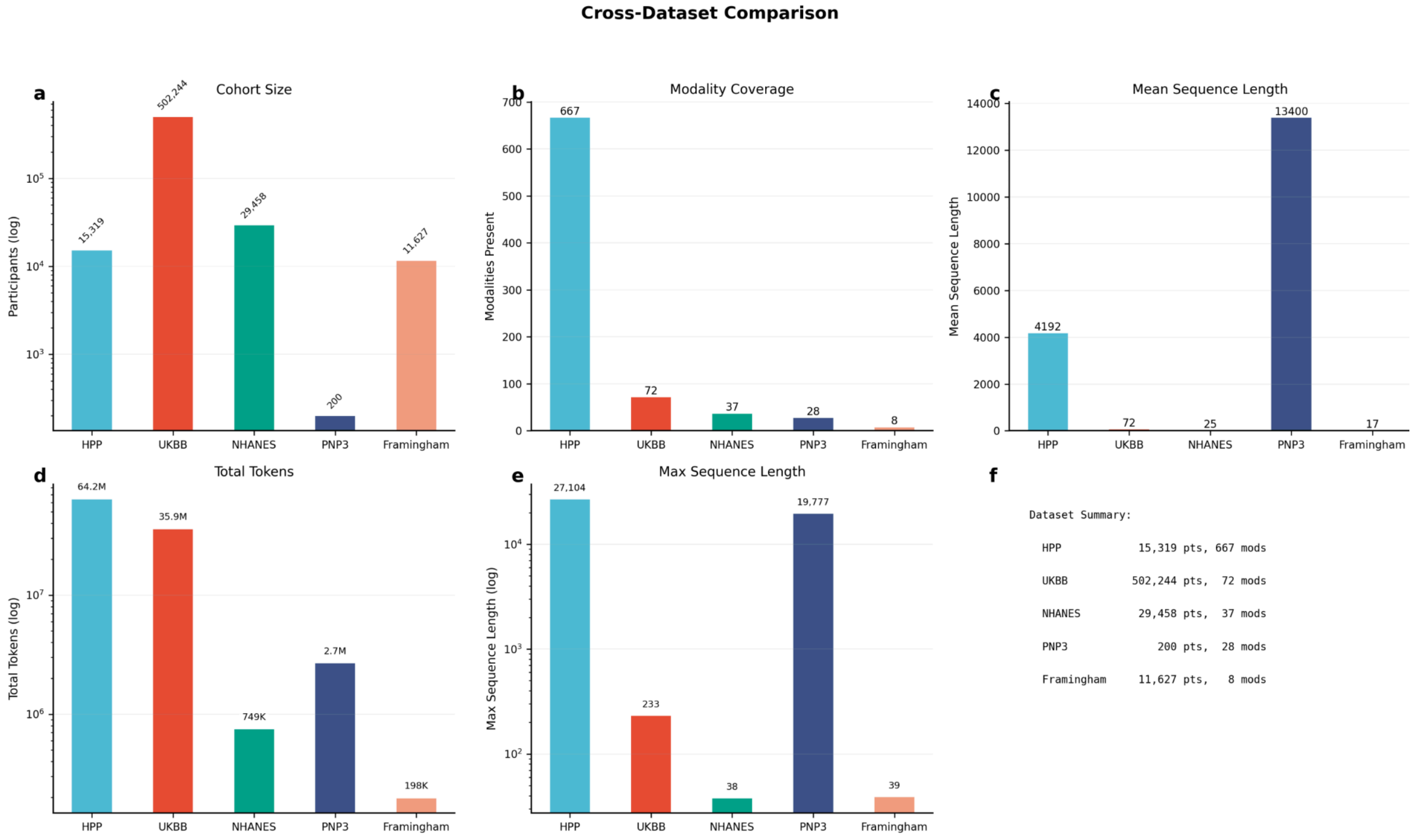

**Fig. S12 | Cross-dataset comparison across the five evaluation cohorts.**
a, Cohort size (log scale).
b, Number of HPP-vocabulary modalities present in each cohort.
c, Mean per-participant sequence length.
d, Total tokens per cohort (log scale).
e, Maximum per-participant sequence length.
f, Dataset summary table.
The combination of HPP's modality breadth and PNP3's per-participant depth motivates using HPP for pre-training and PNP3 for prospective intervention evaluation.

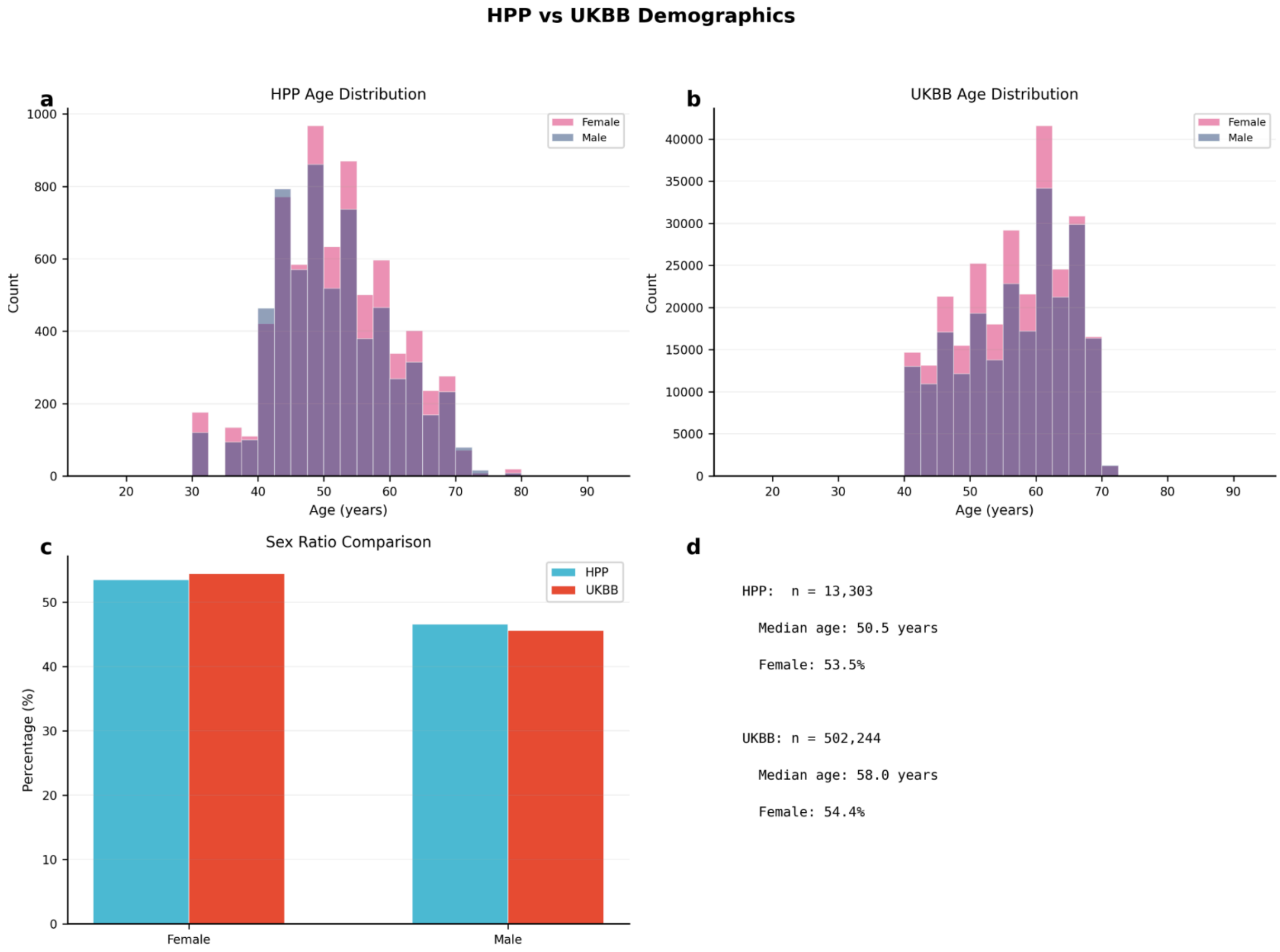

**Fig. S13 | Demographic comparison between HPP and UK Biobank.**
a, HPP age distribution by sex.
b, UK Biobank age distribution by sex .
c, Sex ratio.
d, Summary statistics.

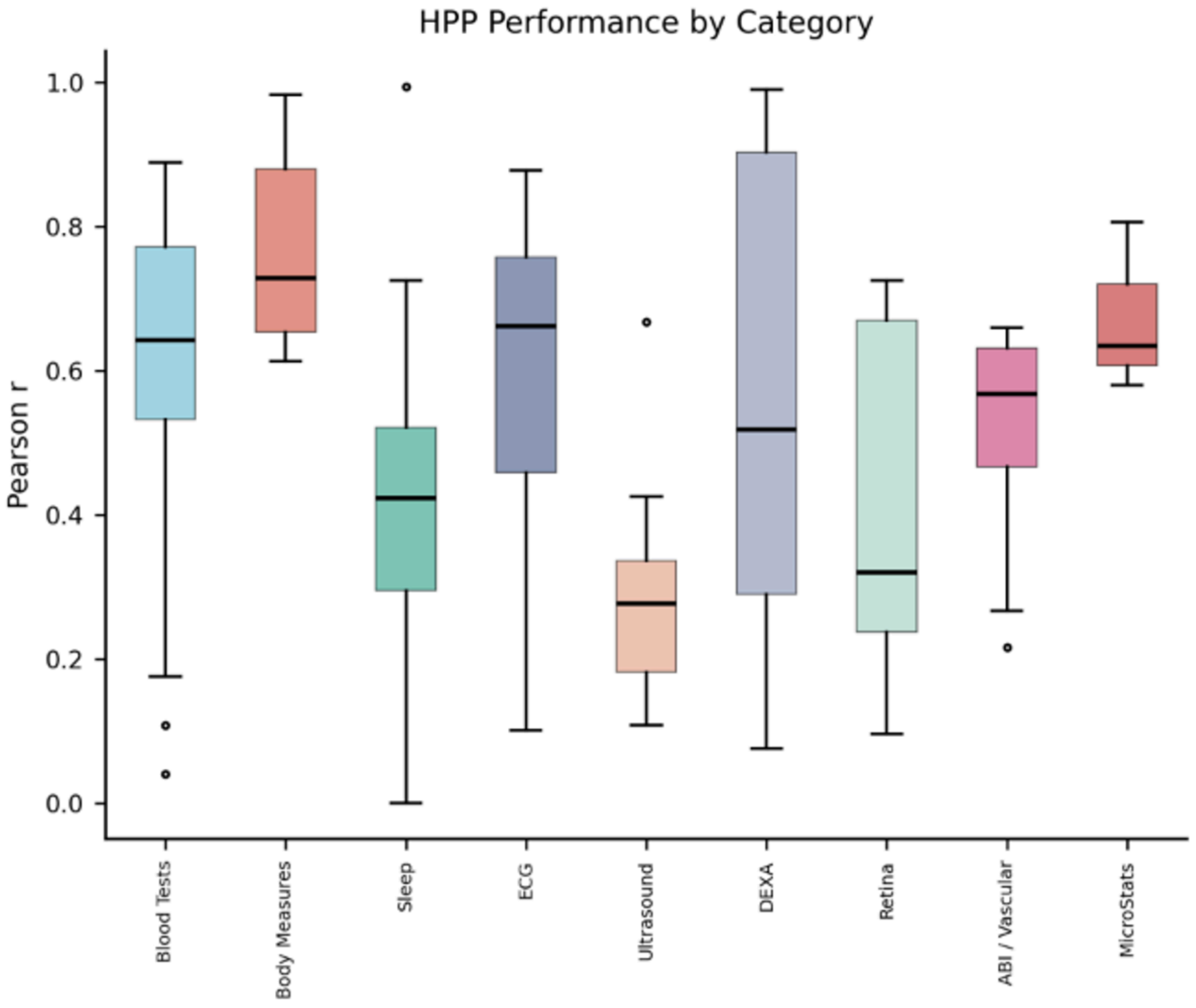

**Fig. S14 | Prediction performance by modality category in HPP.**
Box plots of per-modality Pearson r on the held-out HPP test set, grouped into nine categories. Boxes show the interquartile range, whiskers extend to 1.5 x IQR, and points are outlier modalities. Median performance is highest for body measures, MicroStats, ECG, and blood tests, and lowest for ultrasound and retina . The DEXA category has the widest spread, reflecting the mix of stable (bone area, lean mass) and noisy (regional fat fractions) measurements within it.

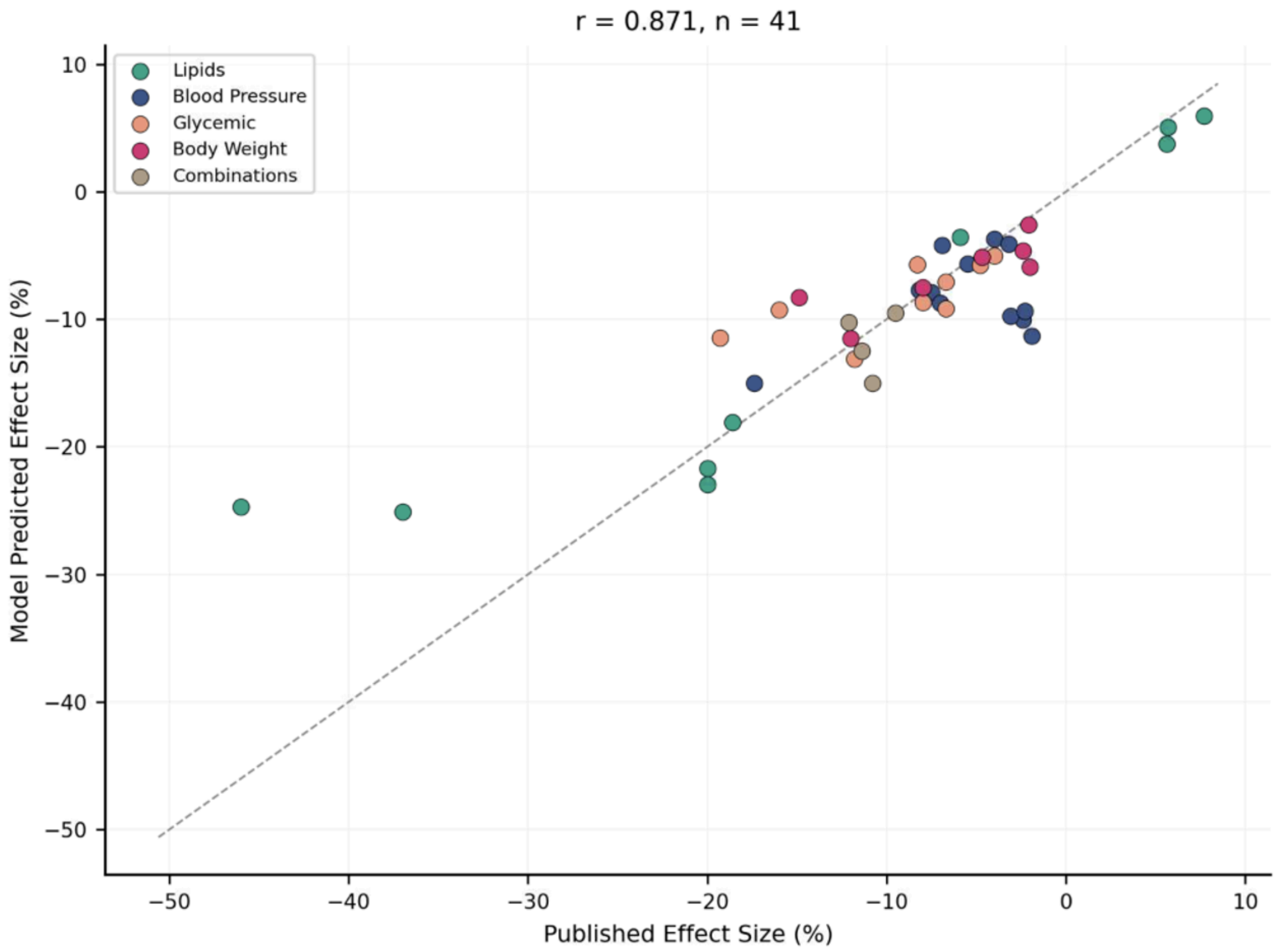

**Fig. S15 | Predicted vs published effect sizes across 41 randomised-trial comparisons.** Scatter of HealthFormer-predicted absolute percentage change from baseline (y-axis) against the published trial point estimate (x-axis) for each of the 41 intervention-outcome pairs in main Fig. 5. Points are coloured by outcome domain: lipids (teal), blood pressure (dark blue), glycaemic (orange), body weight (pink), and combinations (tan). The dashed line is y = x. Across all 41 pairs the predicted direction matches the published estimate in every case, and the overall agreement is r = 0.871. The two outlying lipid points in the lower-left correspond to high-potency statin endpoints from STELLAR (rosuvastatin and atorvastatin on LDL), where the model under-predicts magnitude; this is the same pattern discussed in the main text for the highest-effect-size endpoints.

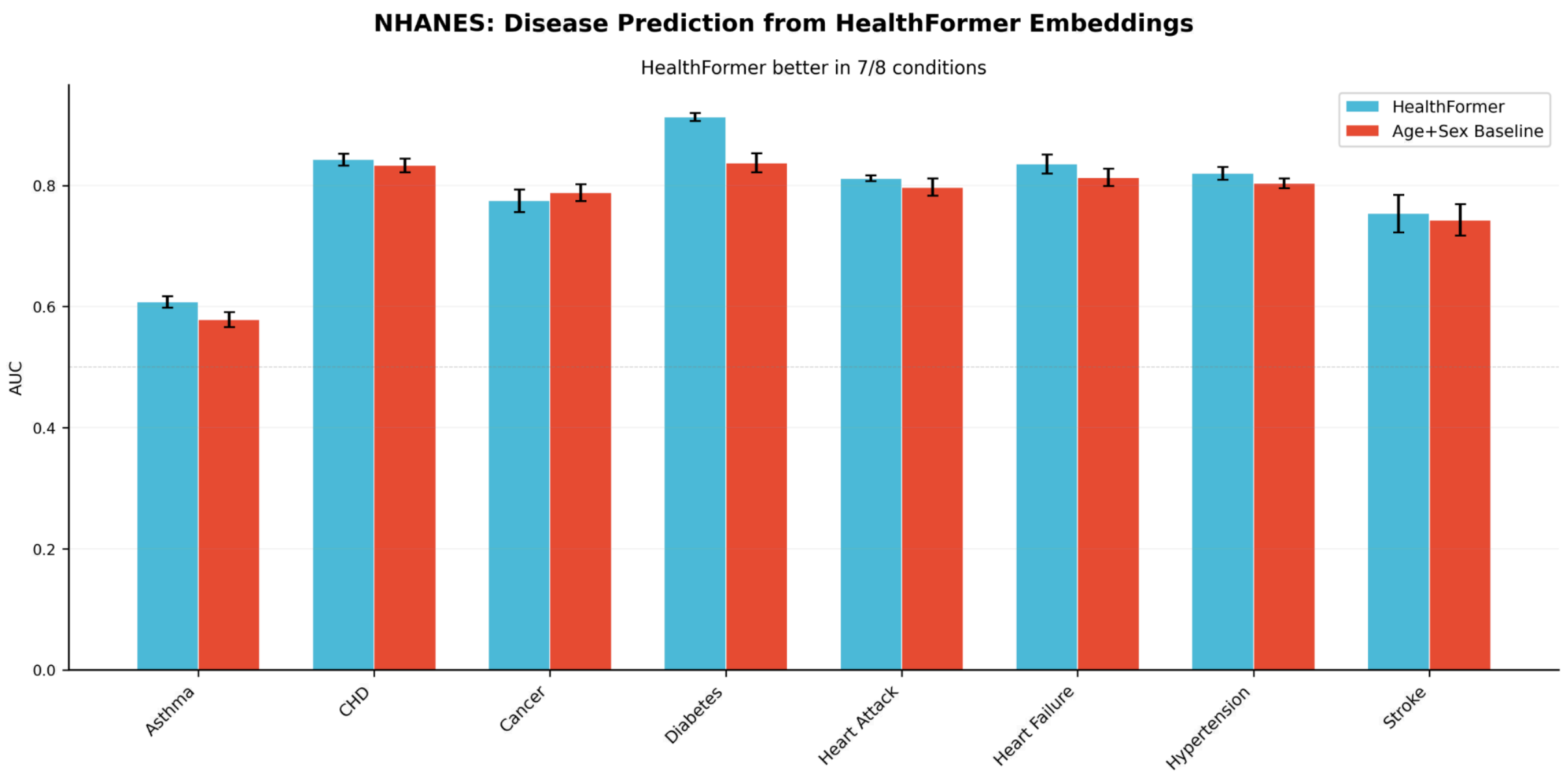

**Fig. S16 | Disease prediction from HealthFormer embeddings in NHANES.**
AUC for binary classification of eight prevalent conditions in NHANES, comparing a logistic model on HealthFormer embeddings (blue) against an age + sex baseline (red). Endpoints: asthma, coronary heart disease (CHD), cancer, diabetes, heart attack, heart failure, hypertension, and stroke. HealthFormer embeddings improve over the demographic baseline in 7 of 8 conditions, with the largest gain for diabetes (about +0.08 AUC). Error bars are mean +/- s.d. across cross-validation folds. The result extends the UK Biobank survival analysis (main Fig. 3d) to a second independent cohort with a much smaller modality overlap with HPP (37 modalities), supporting the generality of the embeddings.

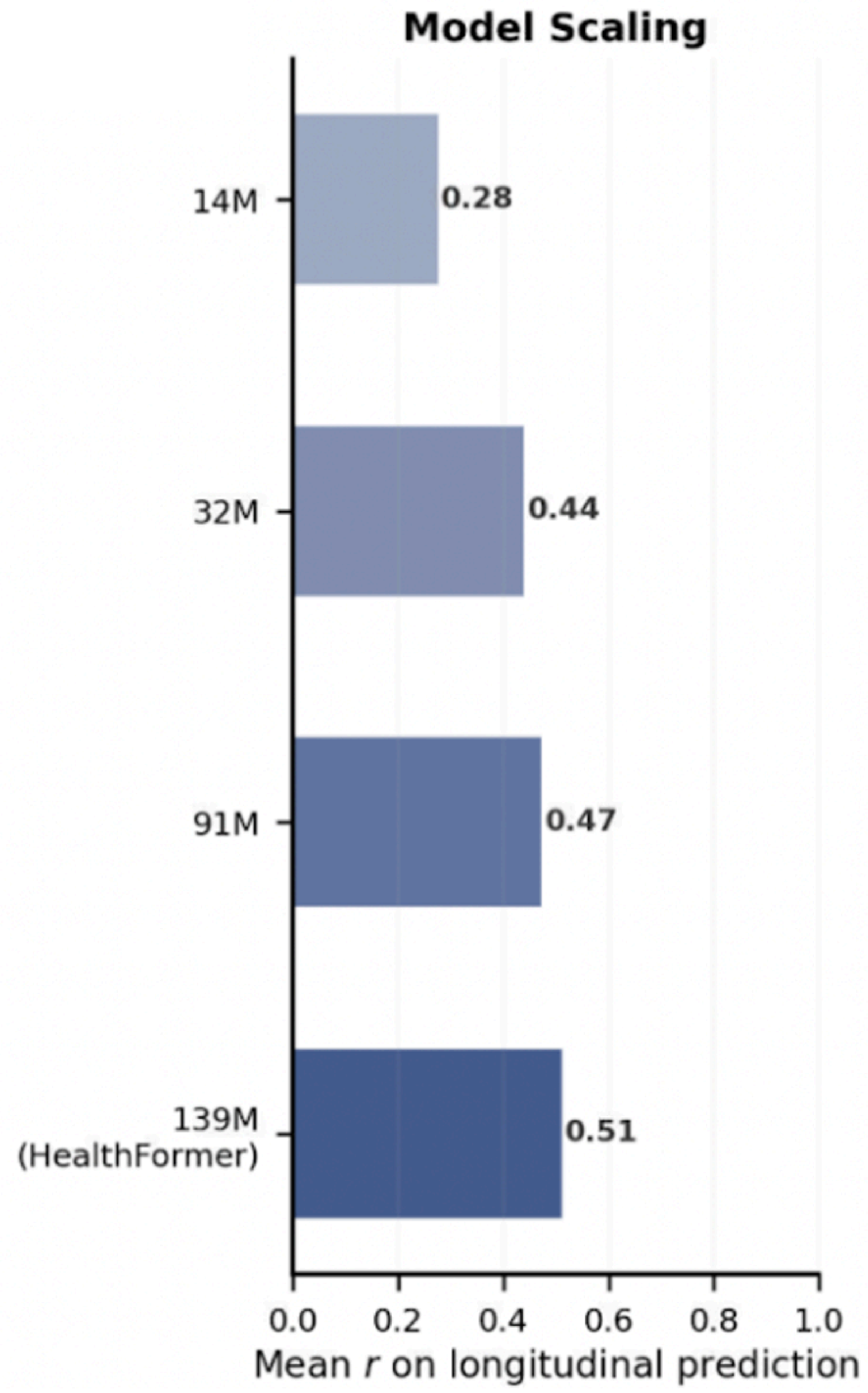

**Fig. S17 | Model-size scaling for longitudinal prediction.**
Mean Pearson r across modalities for predicting visit-2 measurements from visit-1 data, as a function of HealthFormer parameter count. Increasing model size from 14 M to 32 M to 91 M to 139 M parameters yields monotonic improvements (mean r = 0.28, 0.44, 0.47, 0.51). All subsequent results in the paper use the largest (139 M) configuration. The scaling shape is consistent with patterns reported for transformer language models and supports further gains from larger physiological-trajectory models.

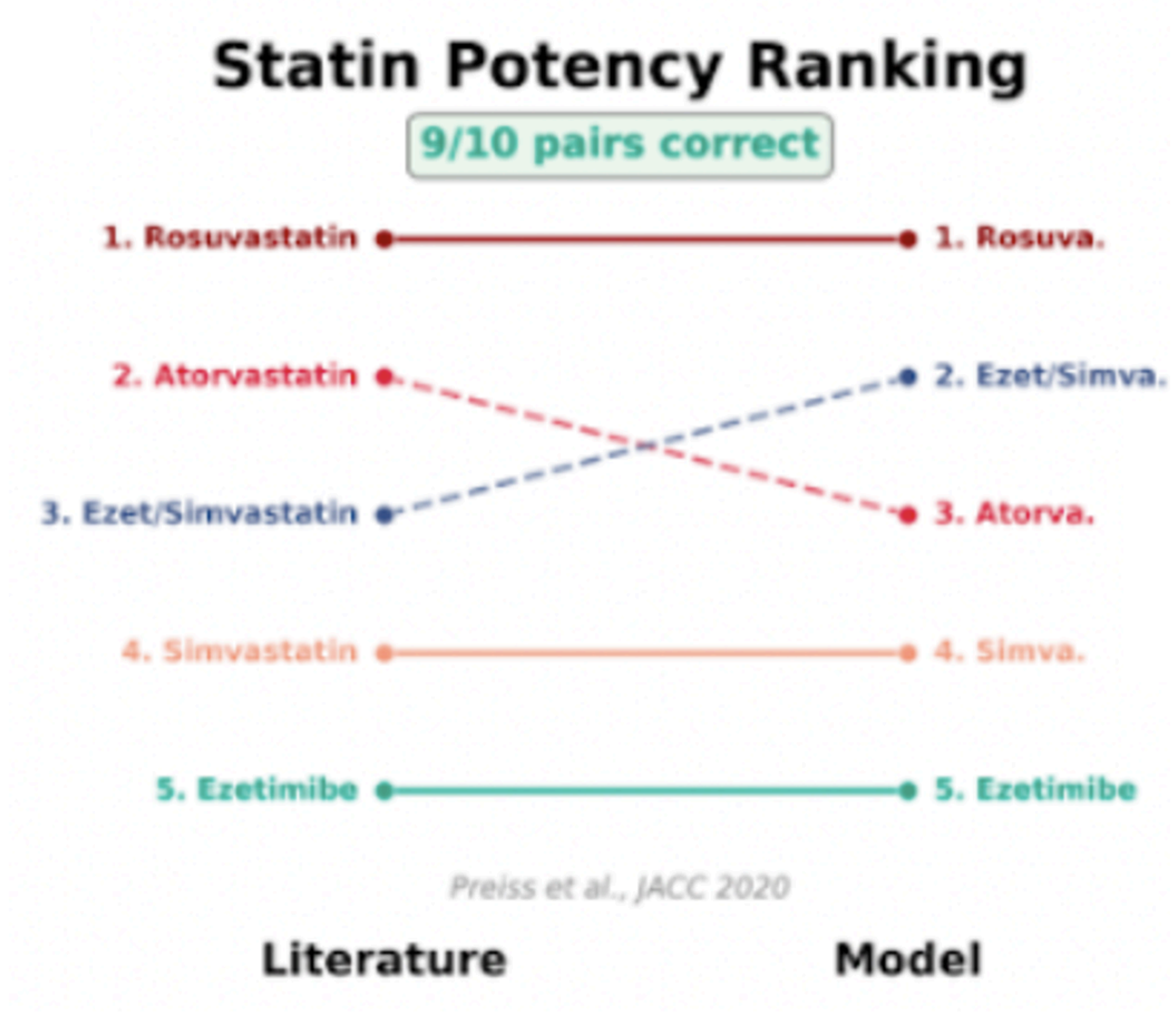

**Fig. S18 | Statin potency ranking: literature vs HealthFormer.**
Slope chart comparing the published clinical potency ranking of five lipid-lowering regimens (Preiss et al., JACC 2020) on the left against the HealthFormer-predicted ranking on the right. Coloured solid lines connect agents that the model places at the same position as the literature (rosuvastatin = 1, simvastatin = 4, ezetimibe = 5); dashed crossed lines connect the two adjacent positions that the model swaps (atorvastatin and ezetimibe + simvastatin, ranks 2 and 3). The model recovers 9 of 10 pairwise potency comparisons, indicating that drug-class structure beyond mere indication confounding is encoded in the learned representation.